\documentclass{article}
\usepackage{amsmath}     
\usepackage{amsthm}     
\usepackage{adjustbox}   
\newtheorem{theorem}{Theorem}
\newtheorem{definition}[theorem]{Definition}
\newtheorem{proposition}[theorem]{Proposition}
\usepackage{amsthm}
\usepackage[normalem]{ulem}
\usepackage{soul}
\usepackage{xcolor}
\newtheorem{analysis}{Analysis}
\newtheorem{conclusion}{Conclusion}
\usepackage{lmodern} 
\usepackage[T1]{fontenc}
\usepackage{lmodern}
\usepackage[unicode]{hyperref}

\usepackage[final]{neurips_2025}
\usepackage{bm}
\usepackage[utf8]{inputenc} 
\usepackage[T1]{fontenc}    
\usepackage{hyperref}       
\usepackage{cleveref}
\usepackage{url}            
\usepackage{booktabs}       
\usepackage{amsfonts}       
\usepackage{nicefrac}       
\usepackage{microtype}      
\usepackage{xcolor}         
\usepackage{amssymb}
\usepackage{amsmath}
\usepackage{wrapfig}
\usepackage{subcaption} 
\usepackage{tcolorbox}
\usepackage{caption}
\usepackage{adjustbox} 
\usepackage{float} 
\usepackage{soul}
\usepackage{geometry}
\usepackage{booktabs}
\usepackage{siunitx}
\usepackage{colortbl}
\usepackage{enumitem}
\usepackage{siunitx}
\usepackage{enumitem}
\usepackage{multirow} 
\usepackage{array} 
\usepackage{amsfonts} 
\usepackage[noend]{algpseudocode} 
\usepackage{algorithm}
\usepackage{times} 

\usepackage{amstext}

\algnewcommand\algorithmicinput{\textbf{Input:}}
\algnewcommand\Input{\item[\algorithmicinput]}
\algnewcommand\algorithmicoutput{\textbf{Output:}}
\algnewcommand\Output{\item[\algorithmicoutput]}
\algnewcommand\algorithmicparameters{\textbf{Parameters:}}
\algnewcommand\Parameters{\item[\algorithmicparameters]}

\usepackage[dvipsnames]{xcolor}

\definecolor{mygreen}{RGB}{112,173,71}
\definecolor{myblue}{RGB}{91,155,213}
\definecolor{myorange}{RGB}{237,125,49}
\definecolor{myred}{RGB}{255,22,67}
\definecolor{honey}{RGB}{201,89,95}
\definecolor{amber}{RGB}{180,148,41}
\definecolor{darkblue}{RGB}{84,96,126}

\newif\ifdraft
\drafttrue
\ifdraft
\newcommand{\hwdc}[1]{{\color{blue}[\textbf{Howard:} #1]}}
\newcommand{\kac}[1]{{\color{purple}[\textbf{Kfir:} #1]}}
\newcommand{\yac}[1]{{\color{red}[\textbf{Yuval:} #1]}}
\newcommand{\jwc}[1]{{\color{red}[\textbf{Jian} #1]}}

\newcommand{\todo}[1]{{\color{blue}[TODO: #1]}}

\else
\newcommand{\hwdc}[1]{}
\newcommand{\kac}[1]{}
\newcommand{\yac}[1]{}
\newcommand{\jwc}[1]{}
\newcommand{\todo}[1]{}

\fi

\setul{0.5ex}{0.3ex}
\newcommand{\colorul}[2][orange]{%
  \bgroup
  \setulcolor{#1}%
  \ul{#2}%
  \egroup
}

\title{GAM-Agent: Game-Theoretic and Uncertainty-Aware Collaboration for Complex Visual Reasoning}

\author{
Jusheng Zhang\textsuperscript{1,*},
Yijia Fan\textsuperscript{1,*},
Wenjun Lin\textsuperscript{1},
Ruiqi Chen\textsuperscript{1},
\\[1.5ex]
Haoyi Jiang\textsuperscript{1},
Wenhao Chai\textsuperscript{2},
Jian Wang\textsuperscript{3},
Keze Wang\textsuperscript{1,†}
\\[1ex]
\textsuperscript{1}Sun Yat-sen University\\
\textsuperscript{2}Princeton University\\
\textsuperscript{3}Snap Inc.\\[0.5ex]
\textsuperscript{}
\textsuperscript{†}Corresponding author: \texttt{kezewang@gmail.com}
}

\begin{document}
\maketitle
\begin{abstract}
    {We propose GAM-Agent, a game-theoretic multi-agent framework for enhancing vision-language reasoning. Unlike prior single-agent or monolithic models, GAM-Agent formulates the reasoning process as a non-zero-sum game between base agents, each specializing in visual perception subtasks, and a critical agent that verifies logical consistency and factual correctness. Agents communicate through structured claims, evidence, and uncertainty estimates. An uncertainty-aware controller dynamically adjusts collaboration and triggers multi-round debates when disagreement or ambiguity is detected, yielding more robust and interpretable predictions. Experiments on four challenging benchmarks—MMMU, MMBench, MVBench, and V*Bench—show that GAM-Agent consistently improves performance across diverse VLM backbones. In particular, it improves small-to-mid-scale models such as Qwen2.5-VL-7B and InternVL3-14B by 5–6\%, while still improving strong models such as GPT-4o by up to 2–3\%. GAM-Agent is modular, scalable, and generalizable, providing a practical path toward reliable and explainable multi-agent multimodal reasoning. Code is publicly available at https://github.com/jushengzhang/Gam-Agent. }
\end{abstract}
\vspace{-15pt}
\section{Introduction}
Recently, large language models (LLMs) \cite{GPT2,liu2024deepseek,achiam2023gpt,guo2025deepseek, llama2,Attention,MAT} has made significant progress in tasks such as complex reasoning, code generation, and knowledge integration, overcoming the limitations of single models through multi-agent collaboration \cite{Z1,LLM-Agent1,LLM-Agent2,LLM-Agent3,LLM-Agent4,Z8}. In contrast, although there have been advances in integrating visual perception with language understanding in the field of vision language models (VLMs) \cite{CLIP,VLMSB1,VLMZS,VLMZS2,VIT,Z3,Z7,zhang2025mmcotabenchmarkprobingvisual,zhang2026doubtsyourselftradingvisual}, the potential of multi-agent collaboration for VLMs remains largely untapped. Most existing VLM approaches rely on single models or simple ensemble strategies, which struggle to address challenges like multi-step reasoning and visual ambiguity in complex visual reasoning tasks \cite{danyi,VLMZS2}.
Benefiting from the development of multi-agent systems \cite{agentKS,AgentZS,ZHINENGTI}, some recent studies have proposed multi-agent debate frameworks to compensate for the shortcomings of VLMs and have obtained promising results\cite{wang11,LLM-Agent1,wang202,DMPL}. However, these methods typically adopt simple mechanisms such as averaging or voting mechanisms and lack visual reasoning-based strategic interactions between agents \cite{AgentZS,AgentZS2}. Hence, they perform poorly in high-complexity visual reasoning scenarios (e.g., MMMU \cite{MMMU}). Just as human experts reach consensus through strategic game-like interactions when they disagree \cite{renlei,renlei2,renlei3,pmlr-v80-barrett18a}, introducing a deep collaborative game-theoretic mechanism into complex visual reasoning is a valuable endeavor\cite{9694243}, which has already seen preliminary applications in some LLM-based works \cite{LLM-Agent3,bo2}.

However, current collaborative game-theoretic methods are often highly complex and heavily rely on reasoning paths, clues, and the fusion of multiple information, making them difficult to apply directly to visual reasoning \cite{zanshi,zanshi2,zanshi3,Z6,Z4}. To address this issue, this paper explores the underlying architecture of existing VLMs, effectively utilizes important intermediate results in the reasoning process, and extracts representations of uncertainty in reasoning outcomes \cite{heinemann2009measuring}. Specifically, we propose a collaborative game-theoretic framework \cite{Nash1950,laurière2024learningmeanfieldgames,2333,55555}, named GAM-Agent, based on game-theoretic and uncertainty-aware inference. In this way, the complex visual reasoning process can be modeled as a non-zero-sum game involving multiple agents collaborating to reach a consensus\cite{yu2022learningefficientmultiagentcooperative,laurière2024learningmeanfieldgames}. Specifically, we encourage agents to share their respective assessments of reasoning uncertainty and engage in a progressive interactive game to guide GAM-Agent to evolve step-by-step and ultimately reach a consensus. 

To address these challenges, we introduce \textbf{GAM-Agent}, a novel agent collaboration framework centered around a strategic interplay between two specialized agent cohorts: \textit{Base Agents} and \textit{Critical Agents}.
The Base Agents are tasked with initial visual interpretation and evidence generation from distinct perspectives, such as object recognition, scene description, and textual analysis from images.
Concurrently, Critical Agents, acting as reasoning critique experts, scrutinize the outputs from Base Agents and evaluate factual accuracy, logical coherence, and overall completeness.
The core of our GAM-Agent lies in modeling the interaction between these Base and Critical Agents as a non-zero-sum game, fundamentally arbitrated by quantified uncertainty.
In this game, agents iteratively share and refine their uncertainty assessments regarding their claims and evidence, engaging in a structured debate process aimed at progressively reducing ambiguity and converging toward a consensus. This uncertainty-driven, game-theoretic collaboration allows for dynamic and strategic integration of diverse insights, leading to more robust and reliable visual reasoning outcomes.
Specifically, \textit{Base Agents} first generate diverse preliminary analyses and identify supporting evidence for their claims.
These outputs are then processed by a \textit{Claim Parser} module, which deconstructs the unstructured responses into structured information tuples, and an \textit{Evidence Mapping} module, which links these textual claims to specific visual regions in the input image, thereby grounding the reasoning process.
Besides, an \textit{Uncertainty Quantification} mechanism continuously assesses the confidence of each agent's contribution.
The entire process is orchestrated by a \textit{Debate Controller \& Integrator}.
This component first evaluates the initial consensus and system uncertainty. If significant discrepancies or high uncertainty are detected, it initiates an iterative debate.
During this debate, \textit{Base Agents} refine their arguments, while \textit{Critical Agents} provide evaluations, with the \textit{Uncertainty Quantification} guiding the dynamic weighting and integration of information.
This iterative loop continues, progressively refining the collective understanding and reducing uncertainty until a robust consensus is achieved or termination criteria are met. Extensive and comprehensive evaluations on large-scale benchmarks demonstrate the superiority of our GAM-Agent. Experimental results show that GAM-Agent achieves significant performance improvements on multiple complex visual reasoning benchmarks such as MMMU, MMBench, MVBench, and V*Bench. For example, it boosts the accuracy of small-to-mid scale VLMs by 5--6\% and enhances the accuracy of top-tier models like GPT-4o by up to 2--3\%.

\section{Methodology}
Our GAM-Agent is defined as a six-tuple $S = (E, A, \Phi, M, P, D)$: $E = {1, 2, \ldots, N}$ represents the set of $N$ expert agents, each possessing distinct cognitive perspectives. $A = {A_1, A_2, \ldots, A_N}$ defines the set of analysis capability mapping functions, where $A_i: \mathcal{X} \times \mathcal{P}r \rightarrow \mathcal{R}$ maps an input image $X$ and a question $P_r$ to agent $i$'s response $R_i$. $\Phi = {\Phi_1, \Phi_2, \ldots, \Phi_N}$ is the set of uncertainty assessment functions, where $\Phi_i: \mathcal{R} \rightarrow [0,1]$ quantifies the uncertainty $U_i$ associated with response $R_i$. $M = {M_1, M_2, \ldots, M_N}$ represents the set of evidence localization mapping functions, where $M_i: \mathcal{C}i \times \mathcal{X} \rightarrow \mathcal{V} \times [0,1]$ associates a claim $c$ with a visual region $r$ and a confidence score $\sigma$. $\mathcal{C}i$ denotes the set of claims made by agent $i$, and $\mathcal{V}$ is the space of visual regions. $P: \mathcal{R} \rightarrow {(c_j, \sigma_j, e_j, r_j)}{j=1}^m$ is the claim and evidence parser, converting an unstructured response $R$ into a set of structured information tuples, each containing a claim $c_j$, confidence $\sigma_j$, textual evidence $e_j$, and visual region reference $r_j$. $D: \mathcal{X} \times \mathcal{P}r \times {R_i, U_i}{i=1}^N \rightarrow \mathcal{R}_{final}$
the debate module aligns conflict and consensus for $R_{final}$.
\begin{figure}[t]
    \centering
    \includegraphics[width=0.8 \linewidth]{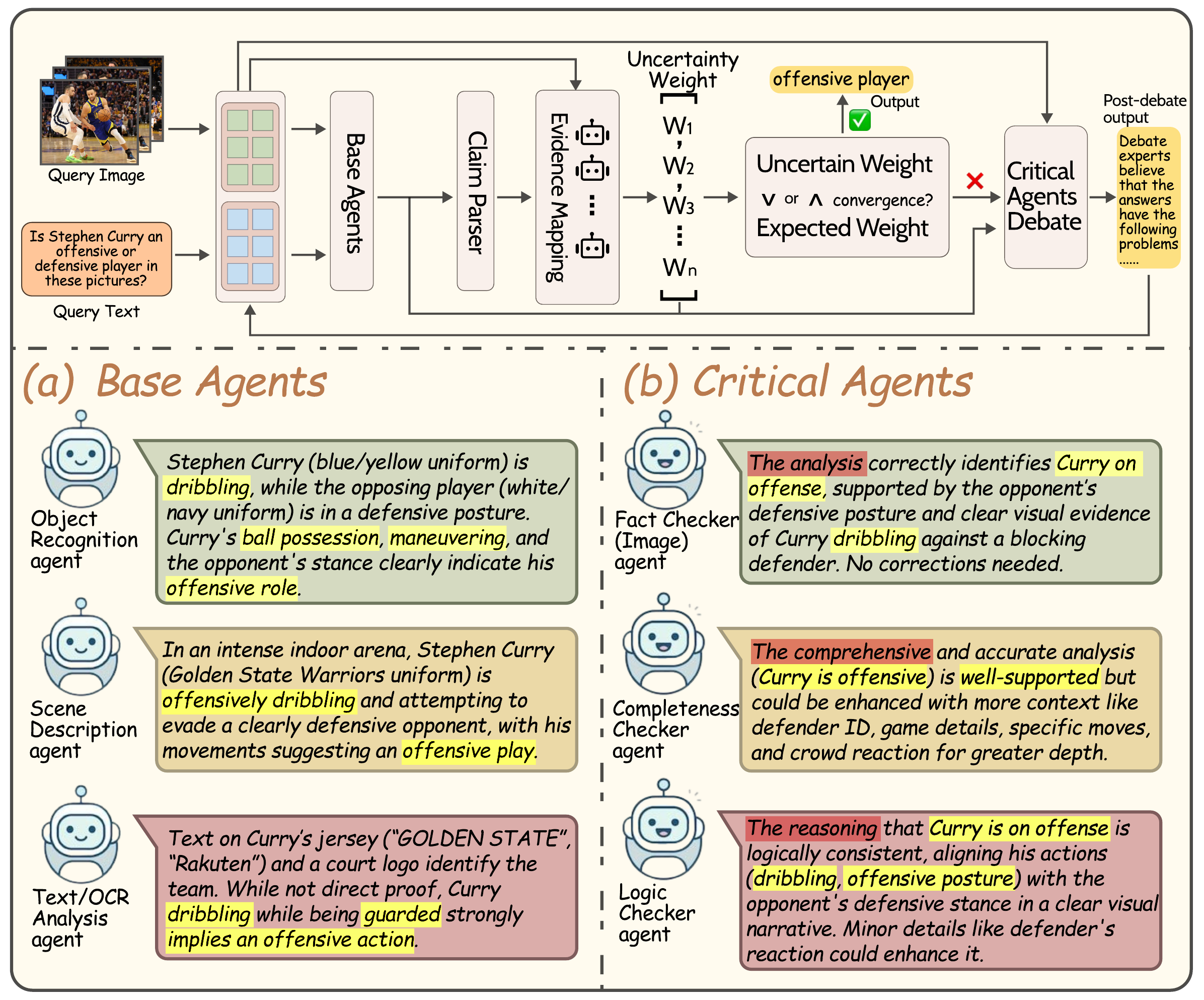}
    \vspace{-8pt}
    \caption{Our GAM-Agent pipeline processing a visual query: Initial responses from Base Agents($R_\text{i}$) undergo Claim Parsing and Evidence Mapping before an Uncertainty Convergence Judgment assesses consensus. If required, the Debate Controller \& Integrator orchestrate an iterative loop, engaging both Base Agents for argumentation and Critical Agents for verification, refining the response until convergence or termination to produce the final output.}
    \vspace{-12pt}
    \label{fig:pipeline}
\end{figure}
\vspace{1mm}
\subsection{Base and Critical Agents}
Our GAM-Agent employs two distinct categories of agents: Base Agents and Critical Agents, each with specialized roles in the visual reasoning process.
\textbf{Base Agents\cite{LLM-Agent1}} are conceived as \textit{specialized visual reasoning experts}. Their primary function is to analyze the visual input and generate initial interpretations or findings related to specific aspects of the scene or query. Examples of Base Agents include: an \textbf{Object Recognition Agent}, tasked with identifying and localizing objects within the visual input; a \textbf{Scene Description Agent}, focused on generating a textual description of the overall scene, its context, and the relationships between entities; and an \textbf{OCR Agent}, specialized in extracting textual information from images. Critical Agents function as \textit{specialized reasoning critique experts}. Their role is to scrutinize the outputs generated by the Base Agents (and potentially other Critical Agents), identify inconsistencies, and assess factual accuracy, logical coherence, and overall completeness. Examples of Critical Agents include: a \textbf{Fact Checker Agent}, which verifies the factual claims made by other agents against known information or contextual understanding; a \textbf{Completeness Checker Agent}, which assesses whether the information provided is sufficient and comprehensive in addressing the query or task; and a \textbf{Logic Checker Agent}, which evaluates the logical consistency and validity of the reasoning steps and conclusions presented.
With these two tiers of expert agents, our GAM-Agent introduces an iterative game-theoretic interplay and information refinement process via uncertainty quantification to perform robust and accurate visual reasoning.

\vspace{-10pt}
\subsection{Uncertainty Quantification Function ($\Phi$)}
\label{sec:uncertainty}

We propose a dual-level $\Phi_i$ that models uncertainty from both response generation and overall semantics. \textbf{Generation-Process Based Uncertainty ($\Phi_{igen+}$)}
When the probability distribution $P_{i,t}$ of the underlying VLM during the generation of the token sequence $Y_i = (y_{i,1}, \dots, y_{i,T_i})$ is accessible, we propose an integrated multi-feature uncertainty metric $\Phi_{igen+}(R_i)$
\begin{equation}
{\small
\Phi_{igen+}(R_i) = \frac{1}{T_i} \sum_{t=1}^{T_i} \left[ \alpha \cdot \underbrace{H(P_{i,t})}_{\substack{\text{Information Entropy}\\\text{(Hesitation)}}} + \beta \cdot \underbrace{\max(0, 1 - \Delta_{top}(P_{i,t}))}_{\substack{\text{Probability Difference Inverse}\\\text{(Lack of Conviction)}}} \right]
}
\end{equation}
where $H(P_{i,t})$ is the information entropy quantifying the dispersion of the token distribution, reflecting the model's hesitation, and $\Delta_{top}(P_{i,t}) = p_{i,t}(1) - p_{i,t}(2)$ is the difference between the top two token probabilities, capturing the model's conviction in its top choice. Parameters $\alpha$ and $\beta$ balance these complementary signals. This method captures microscopic uncertainty patterns during token generation, shifting from output evaluation to generation process evaluation.

\subsubsection{Semantic-Marker Based Uncertainty ($\Phi_{isem}$)}
When generation probabilities are unavailable, our GAM-Agent adopts a semantic marker-based strategy, $\Phi_{isem}(R_i)$, to assess uncertainty. This strategy systematically identifies and quantifies uncertainty markers within the text:
$
\Phi_{isem}(R_i) = \sigma_{\text{sigmoid}} \left( k \cdot \left( \rho(R_i) - \text{offset} \right) \right),
$
where $W$ represents a multi-level lexicon of uncertainty markers. For each marker $w \in W$, $\text{weight}(w)$ reflects its uncertainty intensity, $\text{count}(w, R_i)$ denotes its frequency in response $R_i$, and $|R_i|$ is the length of the response. The parameters $k$ and $\text{offset}$ control scaling and bias, respectively, while $\sigma_{\text{sigmoid}}(\cdot)$ normalizes the score to the $(0,1)$ interval. By statistically analyzing semantic features in the text, this strategy provides a reliable estimate of uncertainty. Moreover, the uncertainty $U_i$ for agent $i$ is determined using a priority strategy: if generation probabilities are available, $U_i = \Phi_{igen+}(R_i)$, capturing uncertainty inherent to the generation process. Otherwise, $U_i = \Phi_{isem}(R_i)$ is used, 
reflecting uncertainty inferred from semantic markers. This enables reliable uncertainty quantification for $U_i$ in any setting, supporting response weighting, conflict detection, and iterative debate.
\vspace{-8pt}
\subsection{Initial Integration and Conflict Detection}
\label{sec:integration_conflict}
After obtaining initial responses from expert agents, GAM-Agent introduces an adaptive weight allocation and conflict assessment mechanism. This initial integration process efficiently aggregates diverse viewpoints and precisely identifies cognitive disagreements requiring debate.

\textbf{Initial Weight Allocation\cite{Z1} and Response Integration.}
GAM-Agent dynamically assigns initial weights $w_i^{(0)}$ based on each agent's uncertainty assessment $U_i$ and generates an integrated response $R^{(0)}$:
$w_i^{(0)} = \frac{e^{-\beta U_i}}{\sum_{j=1}^{N}e^{-\beta U_j}}$ → $R^{(0)} = \text{IntegrateJudge}(X,P_r,{(R_i,w_i^{(0)})}_{i=1}^N)$. The parameter $\beta$ controls the sensitivity of weights to uncertainty. The $\text{IntegrateJudge}$ function performs quality assessment, mutual support analysis, and information merging, balancing high-confidence opinions with diverse perspectives. 

\textbf{Conflict Detection and Debate Triggering.}
GAM-Agent calculates the weighted average uncertainty $U_{sys}^{(0)}$ and an inter-expert conflict score ($ConflictScore$) to decide whether to initiate the debate process using a dual-criterion strategy:

\begin{equation}
\small
\resizebox{0.75\textwidth}{!}{$ 
  U_{sys}^{(0)} = \underbrace{\sum_{i=1}^{N} w_i^{(0)} U_i}_{\text{Weighted average uncertainty}} \quad \rightarrow \quad \text{TriggerDebate} = \underbrace{(U_{sys}^{(0)} > \theta_U)}_{\substack{\text{System uncertainty}\\\text{exceeds threshold}}} \lor \underbrace{(ConflictScore > \theta_C)}_{\substack{\text{Significant inter-expert}\\\text{conflict detected}}}
$}
\end{equation}
where $ConflictScore$ is computed from the consistency of key claims, evidence interpretation, and agent logic. If $U_{sys}^{(0)} > \theta_U$ or $ConflictScore > \theta_C$, the system triggers an iterative debate. This ensures quick consensus on easy cases while invoking deeper collaboration when needed.
\vspace{-10pt}
\subsection{Evidence Mapping (M) and Claim Parsing (P)}
\label{sec:evidence_parsing}
GAM-Agent incorporates an Evidence Mapping module ($M$) and a Claim Parsing module ($P$) as core components bridging visual and language reasoning. These modules construct a traceable network linking textual claims to specific visual evidence, ensuring precise alignment between reasoning and visual information.
The collaborative mechanism of these modules can be formalized as follows: the evidence mapping module $M_i$ establishes explicit links between an agent's claim $c \in \mathcal{C}_i$ and a visual region $r$ in image $X$ (specified by bounding box, mask, or description), with $\sigma \in [0, 1]$ quantifying the confidence of this association, that is, $M_i: \mathcal{C}_i \times X \to (r,\, \sigma)$. This creates a direct “claim-visual evidence” link. The claim parsing module $P$ transforms an unstructured response $R_i$ into a set of structured tuples $P(R_i) \to \left\{ (c_j,\, \sigma_j,\, e_j,\, r_j) \right\}_{j=1}^{m_i}$, where each tuple consists of $c_j$ (the specific textual claim), $\sigma_j$ (the associated confidence), $e_j$ (a description of the supporting textual evidence), and $r_j$ (a reference to the relevant visual region).

\textbf{Working Mechanism of Evidence Mapping and Parsing.} The evidence mapping process can be represented as a conditional mapping function from the claim space to the visual region space: $p(r|c,X) = M_i(c,X) \to (r,\sigma)$
where $p(r|c,X)$ represents the probability distribution over visual regions $r$ being relevant evidence given claim $c$ and image $X$, and $\sigma$ is the confidence of this mapping.
The claim parsing process can be formalized as text decomposition and structural reorganization:
\begin{equation}
  \resizebox{0.9\textwidth}{!}{$
    \displaystyle
    P(R_i)
    =
    \underbrace{M_i(\{c_j\},X)\to\{r_j\}_{j=1}^{m_i}}_{\text{Visual Region Mapping}}
    \;\circ\;
    \underbrace{\mathrm{Associate}(\{c_j\},R_i)\to\{e_j\}_{j=1}^{m_i}}_{\text{Evidence Association}}
    \;\circ\;
    \underbrace{\mathrm{Assess}(\{c_j\})\to\{\sigma_j\}_{j=1}^{m_i}}_{\text{Confidence Assessment}}
    \;\circ\;
    \underbrace{\mathrm{Extract}(R_i)\to\{c_j\}_{j=1}^{m_i}}_{\text{Claim Extraction}}
  $}
\end{equation}

This decomposition highlights four steps: claim extraction, confidence assessment, evidence association, and visual region mapping, collectively transforming $R_i$ into structured tuples. This can enhance transparency and evidence traceability in complex visual reasoning by grounding the visual evidence, enabling precise conflict localization, and performing fine-grained debate focusing.

\subsection{Iterative Debate Process: Dynamics, Termination, and Consensus}
\label{sec:debate}
When initial integration reveals high uncertainty or critical disagreements, GAM-Agent initiates an iterative debate process (``Debating Uncertainty''). This structured interaction drives the agent collective towards a high-quality consensus through a feedback loop focusing on key disputes, evidence sharing, and dynamic weight adjustments. The iterative debate process is formalized as a sequence of state transitions. Each round of debate k involves five key steps forming a complete state evolution cycle: (a-b) Disput focus and expert argumentation: Model identifies key points of contention and guides experts to generate targeted arguments: $C_{debate}^{(k)} = \text{IdentifyDisputes}((c_j, \sigma_j, e_j, r_j)j = 1^m, R^{(k-1)}, U_{sys}^{(k-1)})$ $Arg_i^{(k)} = A_i(X, P_r, R^{(k-1)}, C_{debate}^{(k)})$ The IdentifyDisputes function analyzes structured information from the previous round to select key claims $C_{debate}^{(k)}$ with low confidence $\sigma_j$ or significant inter-expert conflict. Each expert then generates an argumentation package including argumentative text $w_i^{(k)} = \frac{\exp(-\beta U_i^{(k)}) \text{ or } \exp(\gamma C_i^{(k)})}{\sum_{j=1}^{N} \exp(-\beta U_j^{(k)}) \text{ or } \sum_{j=1}^{N} \exp(\gamma C_j^{(k)})}$ Parameters $\beta$ and $\gamma$ control sensitivity to uncertainty or confidence changes. This adaptive weighting incentivizes effective argumentation, aligning with game-theoretic principles. \textbf{(d-e) Iterative Response Generation and Uncertainty Evaluation:} we integrate updated expert views to generate the response of the current round and assess the consensus 
\begin{equation}
\small
  \scalebox{0.75}{$
    R^{(k)}
    = \mathrm{IntegrateJudge}\Bigl(
      \underbrace{X, P_r}_{\text{Original Input}},
      \underbrace{R^{(k-1)}}_{\text{Prev.\ Round Base}},
      \underbrace{\{R_j\}_{j=1}^N}_{\text{Initial Resp.}},
      \underbrace{\{Arg_j^{(k)},E_j^{(k)},r_j^{(k)},C_j^{(k)}\}_{j=1}^N}_{\text{Current Expert Updates}},
      \underbrace{\{w_j^{(k)}\}_{j=1}^N}_{\text{Dynamic Weights}},
      \underbrace{C_{\mathrm{debate}}^{(k)}}_{\text{Dispute Focus}}
    \Bigr)
  $}
\end{equation}
In this process, the $\mathrm{IntegrateJudge}$ function fuses the original input $(X, P_r)$, the previous round’s answer $R^{(k-1)}$, each expert’s initial response ${R_j}*{j=1}^N$, and their structured debate updates ${Arg_j^{(k)}, E_j^{(k)}, r_j^{(k)}, C_j^{(k)}}*{j=1}^N$, along with dynamic weights ${w_j^{(k)}}*{j=1}^N$ and the current set of disputed points $C*{\mathrm{debate}}^{(k)}$, to produce the integrated decision $R^{(k)}$. This progressive integration focuses on resolving the present disputes $C_{\mathrm{debate}}^{(k)}$ while preserving consensus from $R^{(k-1)}$. The overall system uncertainty $U_{sys}^{(k)} = {\sum_{i=1}^{N} w_i^{(k)} U_i^{(k)}}
$quantifies the level of consensus. Through multiple iterations, GAM-Agent aggregates agreement and resolves conflict, steadily converging to a high-confidence, low-uncertainty final answer.
\textbf{Debate Termination Decision Mechanism}  monitors multiple criteria to decide whether to continue the debate. Termination occurs as follows:
\begin{equation}
\small
\text{Terminate Debate}
=
\underbrace{\bigl(U_{\mathrm{sys}}^{(k)} < \theta_U\bigr)}_{\text{Uncertainty threshold condition}}
\;\lor\;
\underbrace{\bigl(k \ge K_{\max}\bigr)}_{\text{Max iteration limit}}
\;\lor\;
\underbrace{\bigl(\Delta U_{\mathrm{sys}}^{(k)} < \epsilon\bigr)}_{\text{Convergence speed condition}}
\label{eq:terminate-debate}
\end{equation}
where $\Delta U_{sys}^{(k)} = |U_{sys}^{(k)} - U_{sys}^{(k-1)}|$ represents the change in system uncertainty between consecutive rounds, and $\epsilon$ is a convergence threshold. These conditions correspond to reaching satisfactory certainty, resource limits, or debate stagnation. Upon termination at round $K$, the final response is $R_{final} = R^{(K)}$. This multi-step, multi-round debate enables knowledge sharing and complementarity among agents and forms a self-improving cognitive loop for complex visual scenarios.
\vspace{-10pt}
\section{Experiments}

\subsection{Image and Video Understanding Benchmarks}
\label{sec:image_video_benchmarks_revised_v2}
\begin{table}[t]
\centering
\caption{The evaluation of image and video understanding. Values in parentheses denote the improvement in accuracy over the respective base model (Ori). The best performance for each base model on each benchmark is \textbf{bolded}. Best is in \protect\colorbox{red!25}{red}, and Base (Ori) data is in \protect\colorbox{blue!25}{blue}.}
\label{tab:combined_benchmark_results_revised_v3} 
\resizebox{0.75\textwidth}{!}{%
\scriptsize
\begin{tabular}{ll|ccc}
\toprule
\textbf{Base Model} & \textbf{Framework} & \textbf{MMMU} & \textbf{MMBench\_V11\_Test} & \textbf{MVBench\_Test} \\
\midrule
\multirow{7}{*}{Qwen2.5VL-7B (Small)} & Base (Ori) & \cellcolor{blue!25}{53.82} & \cellcolor{blue!25}{82.61} & \cellcolor{blue!25}{69.62} \\
& DMAD & 55.53 (+1.71) & 85.62 (+3.01) & 72.78 (+3.16) \\
& DMPL & 56.27 (+2.45) & 86.24 (+3.63) & 73.49 (+3.87) \\
& ChatEval & 56.98 (+3.16) & 86.93 (+4.32) & 74.25 (+4.63) \\
& MAD & 55.01 (+1.19) & 85.11 (+2.50) & 72.17 (+2.55) \\
& DebUnc & 57.59 (+3.77) & 87.65 (+5.04) & 74.89 (+5.27) \\
& \textbf{GAM-Agent (Ours)} & \cellcolor{red!25}\textbf{58.93 (+5.11)} & \cellcolor{red!25}\textbf{89.02 (+6.41)} & \cellcolor{red!25}\textbf{76.15 (+6.53)} \\
\midrule
\multirow{7}{*}{Qwen2.5VL-72B (Large)} & Base (Ori) & \cellcolor{blue!25}{68.24} & \cellcolor{blue!25}{88.39} & \cellcolor{blue!25}{70.38} \\
& DMAD & 68.98 (+0.74) & 89.88 (+1.49) & 71.85 (+1.47) \\
& DMPL & 69.43 (+1.19) & 90.31 (+1.92) & 72.36 (+1.98) \\
& ChatEval & 69.81 (+1.57) & 90.75 (+2.36) & 72.81 (+2.43) \\
& MAD & 68.73 (+0.49) & 89.47 (+1.08) & 71.42 (+1.04) \\
& DebUnc & 70.15 (+1.91) & 91.12 (+2.73) & 73.19 (+2.81) \\
& \textbf{GAM-Agent (Ours)} & \cellcolor{red!25}\textbf{70.98 (+2.74)} & \cellcolor{red!25}\textbf{91.82 (+3.43)} & \cellcolor{red!25}\textbf{74.12 (+3.74)} \\
\midrule
\multirow{7}{*}{InternVL3-14B (Small)} & Base (Ori) & \cellcolor{blue!25}{67.09} & \cellcolor{blue!25}{83.54} & \cellcolor{blue!25}{76.59} \\
& DMAD & 68.83 (+1.74) & 86.63 (+3.09) & 79.78 (+3.19) \\
& DMPL & 69.52 (+2.43) & 87.35 (+3.81) & 80.57 (+3.98) \\
& ChatEval & 70.27 (+3.18) & 88.14 (+4.60) & 81.39 (+4.80) \\
& MAD & 68.25 (+1.16) & 86.01 (+2.47) & 79.13 (+2.54) \\
& DebUnc & 70.94 (+3.85) & 88.86 (+5.32) & 82.15 (+5.56) \\
& \textbf{GAM-Agent (Ours)} & \cellcolor{red!25}\textbf{72.26 (+5.17)} & \cellcolor{red!25}\textbf{90.15 (+6.61)} & \cellcolor{red!25}\textbf{83.23 (+6.64)} \\
\midrule
\multirow{7}{*}{InternVL3-78B (Large)} & Base (Ori) & \cellcolor{blue!25}{72.18} & \cellcolor{blue!25}{87.65} & \cellcolor{blue!25}{78.81} \\
& DMAD & 72.99 (+0.81) & 89.21 (+1.56) & 80.43 (+1.62) \\
& DMPL & 73.47 (+1.29) & 89.68 (+2.03) & 80.95 (+2.14) \\
& ChatEval & 73.88 (+1.70) & 90.15 (+2.50) & 81.41 (+2.60) \\
& MAD & 72.71 (+0.53) & 88.83 (+1.18) & 80.01 (+1.20) \\
& DebUnc & 74.24 (+2.06) & 90.57 (+2.92) & 81.83 (+3.02) \\
& \textbf{GAM-Agent (Ours)} & \cellcolor{red!25}\textbf{75.01 (+2.83)} & \cellcolor{red!25}\textbf{91.53 (+3.88)} & \cellcolor{red!25}\textbf{82.62 (+3.81)} \\
\midrule
\multirow{7}{*}{GPT-4o-0513 (Large)} & Base (Ori) & \cellcolor{blue!25}{68.97} & \cellcolor{blue!25}{83.13} & \cellcolor{blue!25}{75.48} \\
& DMAD & 69.78 (+0.81) & 84.77 (+1.64) & 77.15 (+1.67) \\
& DMPL & 70.26 (+1.29) & 85.29 (+2.16) & 77.71 (+2.23) \\
& ChatEval & 70.65 (+1.68) & 85.73 (+2.60) & 78.19 (+2.71) \\
& MAD & 69.42 (+0.45) & 84.31 (+1.18) & 76.73 (+1.25) \\
& DebUnc & 71.03 (+2.06) & 86.15 (+3.02) & 78.57 (+3.09) \\
& \textbf{GAM-Agent (Ours)} & \cellcolor{red!25}\textbf{71.91 (+2.94)} & \cellcolor{red!25}\textbf{87.04 (+3.91)} & \cellcolor{red!25}\textbf{79.52 (+4.04)} \\
\bottomrule
\end{tabular}%
}
\end{table}

\paragraph{Experiment Setup}
\label{sec:exp_setup_combined_revised_v2}
To comprehensively evaluate our GAM-Agent on challenging image and video benchmarks, we compare it with recent multi-agent methods using various state-of-the-art vision-language models.
We select five prominent VLMs: Qwen2.5VL\cite{qwen25} (7B and 72B parameter versions), InternVL3\cite{InternVL3} (14B and 78B parameter versions), and GPT-4o-0513\cite{achiam2023gpt}. On top of these base models, we implemented our GAM-Agent framework and five other multi-agent frameworks: DMAD\cite{DMAD}, DMPL\cite{DMPL}, ChatEval\cite{ChatEval}, MAD (Multi-Agent Debate)\cite{MAD}, and DebUnc\cite{DebUnc}. The evaluation is conducted on three challenging benchmarks: MMMU\cite{MMMU} (multi-discipline multimodal understanding), MMBench\_V11\_Test (visual reasoning and perception), and MVBench\_Test\cite{MVBench} (video temporal reasoning). For all experimental runs involving debate frameworks, we set a maximum of 3 debate rounds (\texttt{max\_debate\_round=3}). Each framework utilized 3 expert agents ($N=3$) for generating initial responses and argumentation, and 3 critic agents ($N_{\text{crit}}=3$) for frameworks requiring critique/verification, such as GAM-Agent. We used greedy decoding for text generation. Open-source models (Qwen2.5VL, InternVL3) were deployed locally on NVIDIA A100 GPUs, while the closed-source GPT-4o-0513 was accessed via the OpenRouter API. Detailed hyperparameter settings and other experimental configurations are provided in the Supplementary Material. On each benchmark, our GAM-Agent achieves higher overall accuracy (Overall ACC, \%) than its corresponding base model, with improvements summarized in Table~\ref{tab:combined_benchmark_results_revised_v3}. Unless otherwise stated, all experiments in this paper are performed on NVIDIA A100 GPU.
\paragraph{Experiment Results} \label{sec:exp_results_combined_revised_v2}
Table~\ref{tab:combined_benchmark_results_revised_v3} summarizes the comparative performance of GAM-Agent and other multi-agent frameworks across all base models and benchmarks. Results show that GAM-Agent consistently achieves the highest overall accuracy. Improvements are especially notable for smaller models (e.g., Qwen2.5VL-7B and InternVL3-14B), with gains of +5.8\% to +6.7\% on MMBench\_V11\_Test and MVBench\_Test (e.g., +6.61\% on InternVL3-14B/MMBench). Larger models (Qwen2.5VL-72B, InternVL3-78B, GPT-4o-0513) also see consistent gains of +2.6\% to +4.1\% (e.g., +3.91\% on GPT-4o-0513/MMBench). On MMMU, improvements are about 1--1.5 percentage points lower but GAM-Agent still leads, such as +5.11\% for Qwen2.5VL-7B. The supplementary material provides more detailed experimental analyses, including cost-effectiveness studies (Appendix~\ref{supp:Cost and Performance Balance Analysis}), cost dynamics during debates (Appendix~\ref{sec:cost_dynamics_multiround}), a comparative analysis of uncertainty handling capabilities (Appendix~\ref{sec:deep_uncertainty_comparison}), empirical analysis of debate termination conditions (Appendix~\ref{ssec:termination_exp_analysis_en}), and additional hyperparameter studies (Appendix~\ref{sec:supplementary_hyperparameter_ablation}).
GAM-Agent consistently outperforms other multi-agent frameworks (DMAD, DMPL, ChatEval, MAD, DebUnc), delivering larger accuracy gains over base models. For instance, on InternVL3-14B/MMBench\_V11\_Test, GAM-Agent achieves 90.15\% (+6.61\%), compared to DebUnc's 88.86\% (+5.32\%). GAM-Agent's game-theoretic design, leveraging uncertainty, evidence, and iterative debate, boosts VLM performance across model sizes and complex visual tasks, especially for smaller models, with consistent gains for larger ones.
\vspace{-10pt}
\subsection{Guided Visual Search Capabilities}
\vspace{-10pt}
\label{sec:guided_visual_search}

\paragraph{Experiment Setup}
\label{sec:vbench_exp_setup}
With the release of the o3 model, MLLMs have entered a new phase of capability in high-resolution, information-dense visual scenes, bringing attention to the challenge of fine-grained visual grounding.
To evaluate such capabilities, we compare GAM-Agent and the same set of comparative multi-agent frameworks (DMAD, DMPL, ChatEval, MAD, and DebUnc) to the V*Bench benchmark. We utilized the identical five base VLMs as in Section~\ref{sec:image_video_benchmarks_revised_v2}: Qwen2.5VL (7B and 72B), InternVL3 (14B and 78B), and GPT-4o-0513. For all frameworks, including GAM-Agent, we maintained the configuration of $N=3$ base expert agents and $N_{crit}=3$ critic agents (when the debate is triggered), with a maximum of 3 debate rounds ($\texttt{max\_debate\_round=3}$). The objective was to assess how these collaborative frameworks, particularly GAM-Agent with its uncertainty-driven debate, enhance performance on tasks requiring guided visual search capabilities.
\paragraph{Experiment Results.}
\label{sec:vbench_exp_results}
The performance of GAM-Agent and other multi-agent frameworks on the V*Bench benchmark is presented in Table~\ref{tab:vbench_results}. The table details accuracy for Attribute Recognition (AR), Spatial Reasoning (SR), and Overall scores. Base model (Ori) scores are derived from our initial evaluations, and the SEAL framework's reported performance is included as a reference.

\begin{table}[t]
\centering
\caption{Performance on V*Bench (\%). AR: Attribute Recognition, SR: Spatial Reasoning. Values in parentheses denote the accuracy improvement over the respective base model (Ori). The best performance for each base model is \textbf{bolded}. Best overall with GAM-Agent framework is highlighted in \protect\colorbox{red!25}{red}. Base (Ori) data is in \protect\colorbox{blue!25}{blue}.}
\label{tab:vbench_results}
\resizebox{0.75 \textwidth}{!}{%
\scriptsize
\begin{tabular}{ll|ccc}
\toprule
\textbf{Base Model} & \textbf{Framework} & \textbf{Attribute Rec. (\%)} & \textbf{Spatial Reas. (\%)} & \textbf{Overall (\%)} \\
\midrule
\multirow{7}{*}{Qwen2.5VL-7B (Small)} & Base (Ori) & \cellcolor{blue!25}{57.39} & \cellcolor{blue!25}{67.11} & \cellcolor{blue!25}{61.26} \\
& DMAD & 59.12 (+1.73) & 69.03 (+1.92) & 63.28 (+2.02) \\
& DMPL & 59.98 (+2.59) & 70.15 (+3.04) & 64.21 (+2.95) \\
& ChatEval & 60.73 (+3.34) & 70.92 (+3.81) & 65.03 (+3.77) \\
& MAD & 58.65 (+1.26) & 68.44 (+1.33) & 62.72 (+1.46) \\
& DebUnc & 61.54 (+4.15) & 71.88 (+4.77) & 65.87 (+4.61) \\
& \textbf{GAM-Agent (Ours)} & \cellcolor{red!25}\textbf{62.38 (+4.99)} & \cellcolor{red!25}\textbf{72.53 (+5.42)} & \cellcolor{red!25}\textbf{66.51 (+5.25)} \\
\midrule
\multirow{7}{*}{Qwen2.5VL-72B (Large)} & Base (Ori) & \cellcolor{blue!25}{64.52} & \cellcolor{blue!25}{73.08} & \cellcolor{blue!25}{68.80} \\
& DMAD & 65.88 (+1.36) & 74.51 (+1.43) & 70.19 (+1.39) \\
& DMPL & 66.57 (+2.05) & 75.29 (+2.21) & 70.93 (+2.13) \\
& ChatEval & 67.12 (+2.60) & 75.83 (+2.75) & 71.48 (+2.68) \\
& MAD & 65.31 (+0.79) & 74.02 (+0.94) & 69.67 (+0.87) \\
& DebUnc & 67.93 (+3.41) & 76.54 (+3.46) & 72.24 (+3.44) \\
& \textbf{GAM-Agent (Ours)} & \cellcolor{red!25}\textbf{68.77 (+4.25)} & \cellcolor{red!25}\textbf{77.39 (+4.31)} & \cellcolor{red!25}\textbf{72.78 (+3.98)} \\
\midrule
\multirow{7}{*}{InternVL3-14B (Small)} & Base (Ori) & \cellcolor{blue!25}{61.06} & \cellcolor{blue!25}{69.87} & \cellcolor{blue!25}{65.53} \\
& DMAD & 62.97 (+1.91) & 71.92 (+2.05) & 67.50 (+1.97) \\
& DMPL & 63.78 (+2.72) & 72.81 (+2.94) & 68.35 (+2.82) \\
& ChatEval & 64.45 (+3.39) & 73.55 (+3.68) & 69.05 (+3.52) \\
& MAD & 62.24 (+1.18) & 71.23 (+1.36) & 66.80 (+1.27) \\
& DebUnc & 65.21 (+4.15) & 74.38 (+4.51) & 69.80 (+4.27) \\
& \textbf{GAM-Agent (Ours)} & \cellcolor{red!25}\textbf{66.15 (+5.09)} & \cellcolor{red!25}\textbf{75.32 (+5.45)} & \cellcolor{red!25}\textbf{70.33 (+4.80)} \\
\midrule
\multirow{7}{*}{InternVL3-78B (Large)} & Base (Ori) & \cellcolor{blue!25}{68.90} & \cellcolor{blue!25}{76.45} & \cellcolor{blue!25}{72.23} \\
& DMAD & 70.02 (+1.12) & 77.68 (+1.23) & 73.85 (+1.62) \\
& DMPL & 70.63 (+1.73) & 78.33 (+1.88) & 74.48 (+2.25) \\
& ChatEval & 71.15 (+2.25) & 78.82 (+2.37) & 74.99 (+2.76) \\
& MAD & 69.58 (+0.68) & 77.21 (+0.76) & 73.40 (+1.17) \\
& DebUnc & 71.89 (+2.99) & 79.57 (+3.12) & 75.73 (+3.50) \\
& \textbf{GAM-Agent (Ours)} & \cellcolor{red!25}\textbf{72.63 (+3.73)} & \cellcolor{red!25}\textbf{80.24 (+3.79)} & \cellcolor{red!25}\textbf{76.16 (+3.93)} \\
\midrule
\multirow{7}{*}{GPT-4o-0513 (Large)} & Base (Ori) & \cellcolor{blue!25}{71.15} & \cellcolor{blue!25}{78.30} & \cellcolor{blue!25}{74.72} \\
& DMAD & 72.33 (+1.18) & 79.58 (+1.28) & 75.96 (+1.24) \\
& DMPL & 72.98 (+1.83) & 80.29 (+1.99) & 76.64 (+1.92) \\
& ChatEval & 73.51 (+2.36) & 80.77 (+2.47) & 77.14 (+2.42) \\
& MAD & 71.92 (+0.77) & 79.05 (+0.75) & 75.49 (+0.77) \\
& DebUnc & 74.24 (+3.09) & 81.49 (+3.19) & 77.87 (+3.15) \\
& \textbf{GAM-Agent (Ours)} & \cellcolor{red!25}\textbf{74.98 (+3.83)} & \cellcolor{red!25}\textbf{82.15 (+3.85)} & \cellcolor{red!25}\textbf{78.32 (+3.60)} \\
\midrule
\multicolumn{2}{l|}{SEAL (original work)} & 74.78 & 76.31 & 75.39 \\
\bottomrule
\end{tabular}%
\vspace{-10pt}
}
\end{table}

The results in Table~\ref{tab:vbench_results} demonstrate that GAM-Agent consistently enhances the performance of base VLMs on the V*Bench tasks, which demand precise visual search and reasoning. For instance, GAM-Agent improved the Qwen2.5VL-7B model by +5.25\% overall, achieving 66.51\%. Similar significant gains are observed across other base models, such as InternVL3-14B (+4.80\% overall to 70.33\%) and GPT-4o-0513 (+3.60\% overall to 78.32\%). While other multi-agent frameworks also yield improvements over the base models, GAM-Agent consistently achieves the highest scores among them. The gains are generally more pronounced for smaller models, yet remain substantial for larger, more capable base VLMs. Notably, GAM-Agent's performance with models like GPT-4o-0513 (78.32\%) approaches or surpasses SEAL performance (75.39\%) on this challenging benchmark.

GAM-Agent's uncertainty-driven debate mechanism shines on V*Bench, enabling VLMs to quantify uncertainty, ground claims in visual evidence, and iteratively resolve disagreements for precise visual understanding. This guided visual search directs agents to re-evaluate details and converge on evidence-backed interpretations, boosting MLLMs' reasoning, and rivaling visual search.
\vspace{-4mm}
\subsection{Ablation Experiments}
\vspace{-2mm}
\label{sec:ablation}
To assess each component’s contribution, we conducted ablation studies by removing or simplifying modules. Experiments were run on MMBench\cite{MVBench}\_TEST\_V11 using InternVL3-14B as the base model with $N=3$ agents (unless noted), and a maximum of 3 debate rounds (\texttt{max\_debate\_round=3}) for setups involving iterative debate. We report overall accuracy (Acc.), average debate rounds (Deb. Rounds), debate trigger rate (Deb. Trig.), and average LLM inference cost (Cost). Additional parameter ablations are provided in Supplementary Material~\ref{tab:hyperparameter_ablation_mmbench}.
\label{sec:ablation_setup} 
We define five ablation variants to assess key components of GAM-Agent:  
\textbf{\texttt{w/o Multi-Agent (N=1)}}: Single-agent, without collaboration or debate.  
\textbf{\texttt{w/o Uncertainty ($\Phi$)}}: Removes uncertainty; uses uniform weights and simple heuristics.  
\textbf{\texttt{w/o P\&M}}: Disables claim-evidence grounding; debate operates on raw outputs.  
\textbf{\texttt{w/o Debate (D)}}: No iterative refinement; uses only initial integration.  
\textbf{\texttt{w/o Dynamic Weights}}: Debate with fixed agent weights.
Reported metrics: Accuracy (Acc. \%), Avg. Debate Rounds (Deb. Rounds, max 3), Debate Trigger Rate (Deb. Trig. \%), and Inference Cost (tokens/instance).
\textbf{Experiment Results}
\label{sec:ablation_results}
\begin{figure}[t]
    \centering
    \begin{minipage}[t]{0.40\textwidth}
        \centering
        \includegraphics[width=\textwidth]{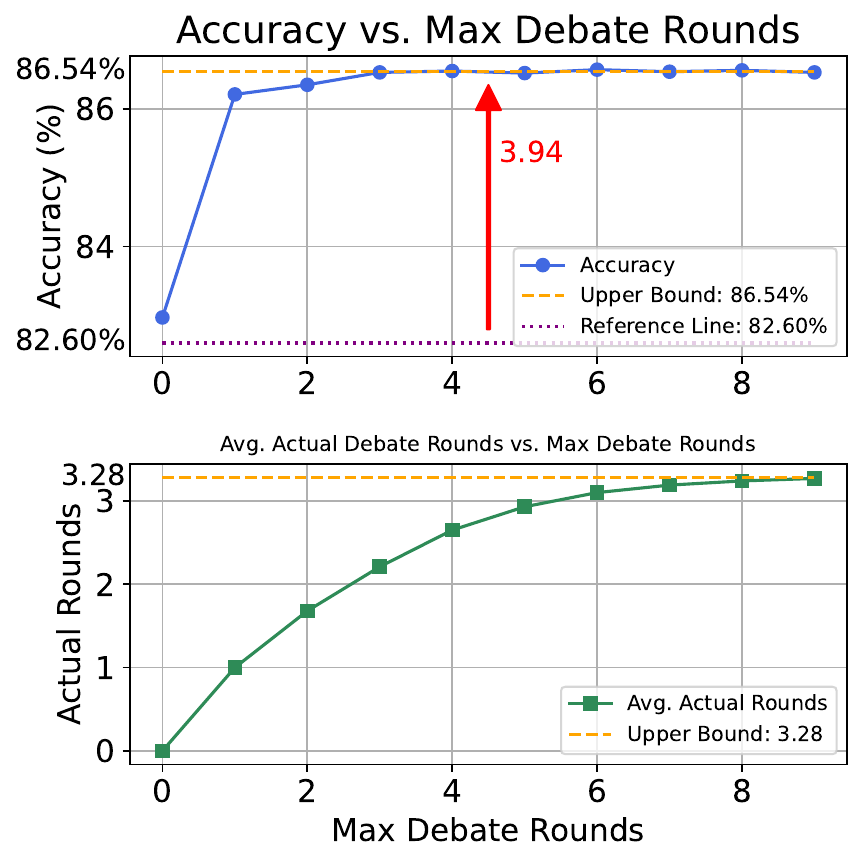}
        \vspace{-15pt}
        \caption{Acc. vs. Max Debate Rounds}
        \vspace{-10pt}
        \label{fig:max_rounds_ablation_qwen7b_focused}
    \end{minipage}
    \hfill
    \begin{minipage}[t]{0.59\textwidth}
        \centering
        \includegraphics[width=\textwidth]{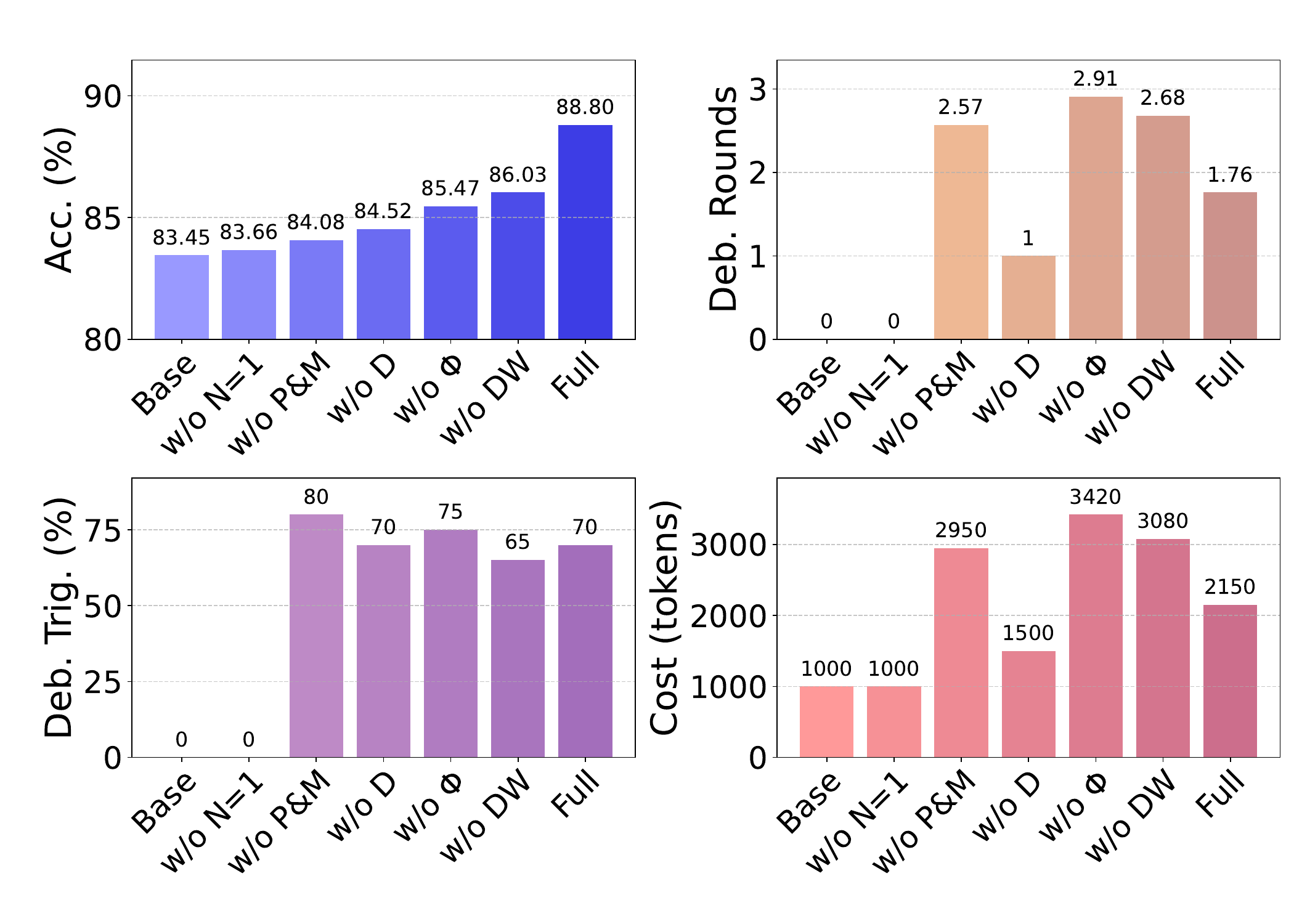}
        \vspace{-15pt}
        \caption{Ablation study results on MMBench\_TEST\_V11.}
        \vspace{-10pt}
        \label{fig:ablation_results_condensed}
    \end{minipage}
\end{figure}
Figure~\ref{fig:ablation_results_condensed} shows that the full GAM-Agent achieves the highest accuracy (88.80\%) with the fewest debate rounds (1.76), reflecting efficient convergence. All ablations degrade performance: removing \texttt{Uncertainty} or \texttt{Dynamic Weights} increases debate length and cost; disabling \texttt{Iterative Debate} cuts cost but reduces accuracy by 4.28\%; removing \texttt{P\&M} causes a 4.72\% drop and high debate triggers (80\%). The \texttt{w/o Multi-Agent} setup performs only slightly above the base model, confirming the necessity of all components.
\vspace{-10pt}
\subsection{Study on Maximum Debate Rounds}
\vspace{-10pt}
\label{sec:ablation_max_rounds_qwen7b_focused}
\label{sec:exp_results_max_rounds_qwen7b_focused}
To assess the effect of debate length, we ablated the \texttt{max\_debate\_round} hyperparameter using Qwen2.5VL-7B on the MMBench\_V11\_Test set. The core GAM-Agent components—uncertainty modeling, claim parsing, and dynamic weighting—were kept unchanged. We varied \texttt{max\_debate\_round} from 0 to 9 and recorded overall accuracy and average actual debate rounds.
The results presented in Figure~\ref{fig:max_rounds_ablation_qwen7b_focused} show that increasing \texttt{max\_debate\_round} improves accuracy up to 3 rounds: from 82.97\% (0 rounds) to 86.21\% (1), 86.35\% (2), and 86.53\% (3). Beyond this, accuracy plateaus around 86.52\%–86.57\%, converging near 86.59\%. Similarly, average actual debate rounds grow with higher \texttt{max\_debate\_round}, from 0.00 (0) to 1.00 (1), 2.21 (3), and up to 3.27 (9), showing early termination via internal convergence criteria. These results highlight that 3–4 debate rounds suffice for near-optimal accuracy with limited cost, as GAM-Agent adaptively halts when uncertainty is resolved.
\vspace{-10pt}
\subsection{Expert Weight Trajectory and System Uncertainty Dynamics in Logical Reasoning Tasks}
\label{sec:Expert Weight}
\begin{figure}[t]
  \centering
  \includegraphics[width=\textwidth]{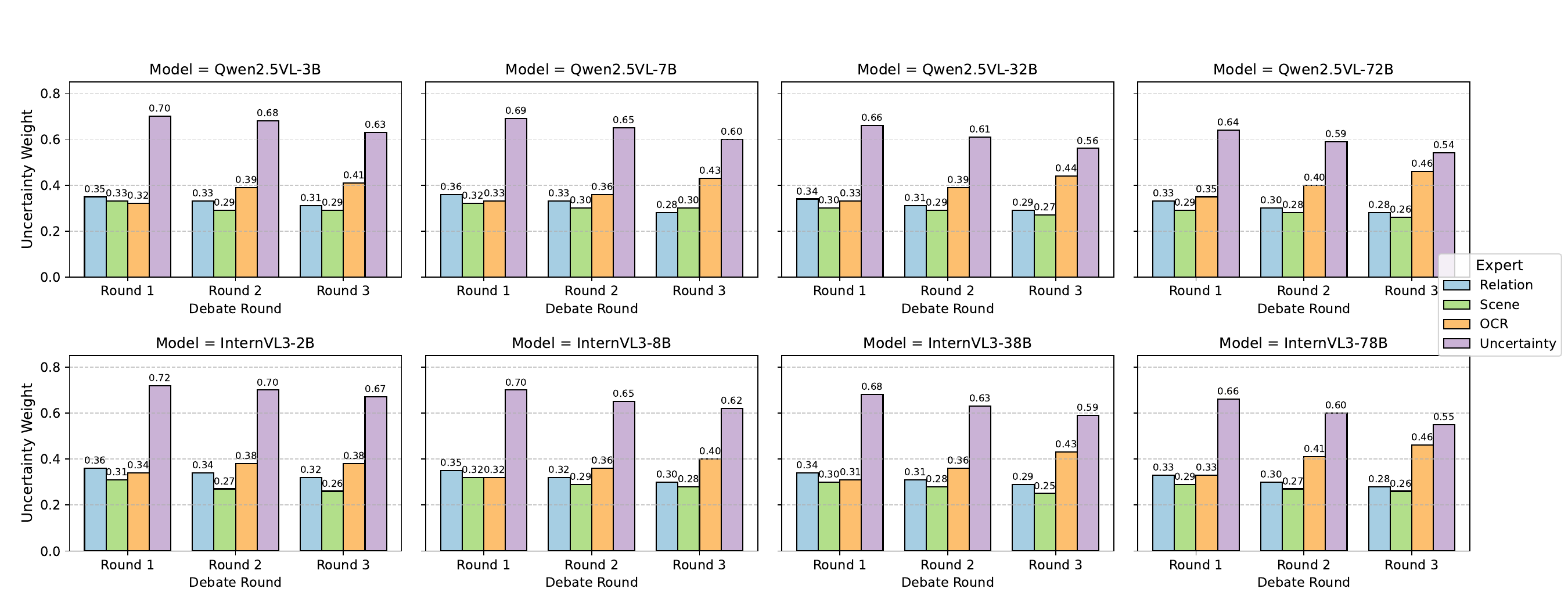} 
  \vspace{-15pt}
  \caption{Evolution of Expert Weights (Relational Reasoning, Scene Description, OCR) and System Uncertainty across three debate rounds for various Qwen2.5VL and InternVL3 series models in logical reasoning tasks. The four bars for each round represent the weights of the Relation expert, Scene expert, OCR expert, and the overall System Uncertainty, respectively.}
  \vspace{-10pt}
  \label{fig:expert_weight_dynamics}
\end{figure}

\vspace{-10pt}
To analyze how GAM-Agent modulates expert influence and manages uncertainty during reasoning, we tracked expert contributions across eight base VLMs—Qwen2.5VL (3B–72B) and InternVL3 (2B–78B). Each setup included three expert agents: Relational Reasoning, Scene Description, and OCR (see Supplementary for prompts). We visualized uncertainty-based weights and system uncertainty over three debate rounds (Figure~\ref{fig:expert_weight_dynamics}), revealing how GAM-Agent dynamically adjusts expert influence to reduce uncertainty and drive consensus.
 Figure~\ref{fig:expert_weight_dynamics} illustrates GAM-Agent’s dynamic expert modulation during reasoning. System uncertainty (purple bars) consistently decreases over three rounds (e.g., Qwen2.5VL-72B: 0.64→0.54; InternVL3-78B: 0.66→0.55), indicating effective convergence. Meanwhile, OCR expert weights (yellow) typically rise (e.g., Qwen2.5VL-72B: 0.35→0.46), while Relational and Scene experts adjust downward. OCR provides stable cues; expert-task alignment varies. Larger models reduce uncertainty, showing GAM-Agent’s capacity-aware, uncertainty-driven coordination.

\vspace{-15pt}
\section{Related Works}
\vspace{-10pt}
\textbf{Multi-agent Systems (MAS) }\cite{A1,A2,LLM-Agent4,agentKS,A3} have gained significant attention in artificial intelligence, particularly with the integration of large language models (LLMs). These systems enhance capabilities in areas such as complex reasoning, code generation, and knowledge integration by leveraging collaborative agents~\cite{VLMZS2}. For example, Du et al~\cite{LLM-Agent1} demonstrated improved factuality and reasoning in LLMs through multi-agent debates, while Li et al~\cite{LLM-Agent2} explored communicative agents to simulate language model societies~\cite{A4, A5, A6}. Despite these advances, the application of multi-agent collaboration in vision-language models (VLMs) remains largely untapped. Existing multi-agent debate frameworks often lack strategic interactions tailored for visual reasoning.

\textbf{Uncertainty Quantification} \cite{55555,B1,B2,B3,C1}  is a cornerstone of reliable decision-making in machine learning, particularly in visual reasoning where ambiguities, such as occlusions or poor lighting, frequently arise. Previous methods, including entropy-based measures and semantic analysis, have been widely adopted \cite{heinemann2009measuring,55555}. In VLMs, uncertainty stems from both visual and linguistic sources, complicating accurate assessment \cite{B4}. 

\textbf{Visual Reasoning} entails interpreting and inferring from visual inputs and is a critical domain for VLMs~\cite{C2,C3,VLMZS,VLMZS2,VIT}. Current VLM strategies often depend on single models or basic ensemble methods, which struggle with multi-step reasoning and visual ambiguities~\cite{C5,VLM-Agent1,C6}. Unlike basic averaging or voting in VLM multi-agent debate, GAM-Agent introduces a game-theoretic collaboration mechanism, leveraging uncertainty and intermediate reasoning for deeper agent synergy.
\vspace{-20pt}
\section{Conclusion}
\vspace{-10pt}
We presented GAM-Agent, a game-theoretic multi-agent reasoning framework designed to enhance the robustness, interpretability, and accuracy of vision-language models on complex multimodal tasks. GAM-Agent facilitates structured claim-evidence interactions guided by uncertainty-aware collaboration by modeling reasoning as a non-zero-sum game among specialized base agents and a critical verification agent. This design allows the system to identify ambiguities, trigger multi-round debates, and dynamically adjust reasoning strategies based on agent confidence and disagreement.
Extensive experiments across four challenging benchmarks demonstrate that GAM-Agent consistently improves performance across a wide range of VLM backbones, including both lightweight and strong foundation models. 
Looking forward, we believe this work opens new directions in self-reflective and multi-agent multimodal reasoning. Future research may explore tighter integration with external tools, extending to dialog agents, or applying GAM-Agent to safety-critical and explainable AI systems.

\section*{Acknowledgements}

This work was supported in part by the National Natural Science Foundation of China (NSFC) under Grant 62276283, in part by the China Meteorological Administration's Science and Technology Project under Grant CMAJBGS202517, in part by Guangdong Basic and Applied Basic Research Foundation under Grant 2023A1515012985, in part by Guangdong-Hong Kong-Macao Greater Bay Area Meteorological Technology Collaborative Research Project under Grant GHMA2024Z04, in part by Fundamental Research Funds for the Central Universities, Sun Yat-sen University under Grant 23hytd006, and in part by Guangdong Provincial High-Level Young Talent Program under Grant RL2024-151-2-11.

\bibliography{main}    
\bibliographystyle{plain} 
\clearpage

\appendix

\begin{center}
     \Large\textbf{Supplementary Material}
\end{center}


\section*{Supplementary Material Overview}
\addcontentsline{toc}{section}{Supplementary Material Overview}
The supplementary material provides detailed experimental setups, results, theoretical analyses, and case studies to support the main paper's claims on the GAM-Agent framework. It includes cost-performance analyses, uncertainty handling evaluations, theoretical derivations, hyperparameter studies, prompt configurations, and case analyses to demonstrate the framework's robustness and limitations.
\tableofcontents
\section{Cost and Performance Balance Analysis}
\label{supp:Cost and Performance Balance Analysis}
\subsection{Experiment Setup}
To evaluate the cost-performance balance of our proposed approach, we conducted a comparative analysis on the MMbench\_TEST\_V11 benchmark. We measured the ``Cost per Instruction on MMbench\_TEST\_V11'' and ``Overall ACC (\%)'' for several leading Vision-Language Models (VLMs) and configurations derived from our framework (denoted as ``Ours''). The commercial models benchmarked include Qwen2.5VL\cite{qwen25} series (7B, 72B via API), Gemini 2.5 Pro Preview\cite{Gemini}, Gemini 2.0 Pro, GPT-4o\cite{achiam2023gpt}, and Claude 3.7, with their respective costs per instruction sourced from OpenRouter API pricing. For our configurations, leveraging models such as Qwen2.5VL (7B, 72B) and InternVL3 (14B, 78B)\cite{InternVL3}, the cost per instruction was calculated based on the inference time on a local NVIDIA A100 GPU. This local GPU cost was estimated using the on-demand price of an AWS p4d.24xlarge instance, which includes 8 A100 40GB GPUs at approximately \$32.77 per hour, translating to about \$4.10 per hour for a single A100 GPU. This setup allows for a direct comparison of the economic efficiency of deploying VLMs via commercial APIs versus utilizing them within our optimized framework on local hardware.

\subsection{Experiment Result}
\begin{wrapfigure}{r}{0.5\textwidth}
  \vspace{-0.5\intextsep} 
  \centering
  \includegraphics[width=0.48\textwidth]{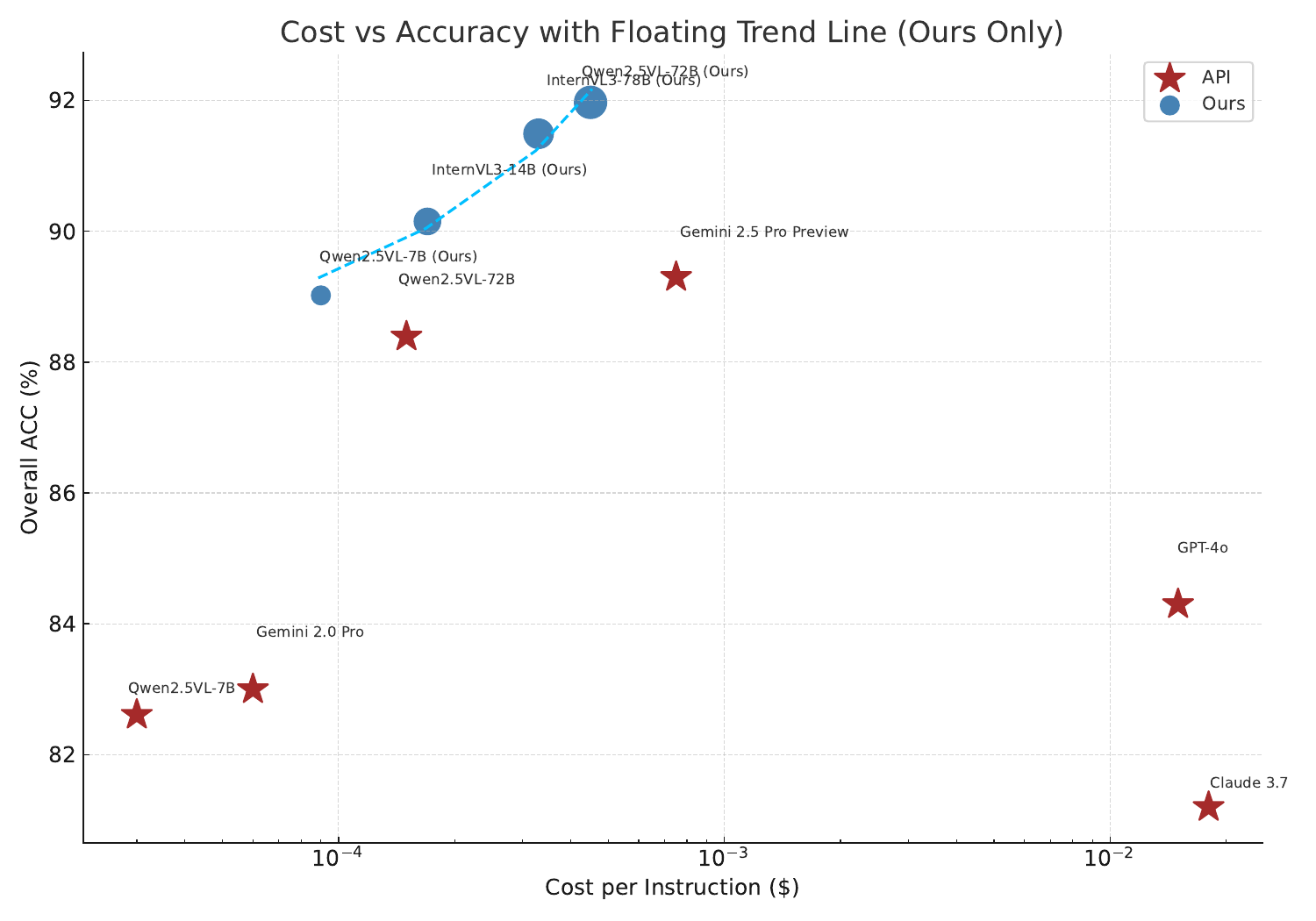} 
  \caption{Comparison of Cost per Instruction and Overall Accuracy on MMbench\_TEST\_V11. ``Ours'' denotes models run with our proposed framework and local inference cost calculation.}
  \label{fig:cost}
  \vspace{-\intextsep} 
\end{wrapfigure}

The results of our cost and performance comparison on the MMbench\_TEST\_V11 benchmark are summarized in Figure~\ref{fig:cost}.
The experimental results presented in Figure~\ref{fig:cost} reveal significant disparities in cost-effectiveness across the evaluated models. Notably, our framework's configurations demonstrate exceptional performance with compelling cost benefits.
InternVL3-78B (Ours) achieved the highest accuracy of $92.2\%$ with a competitive cost of approximately $\$0.00022$ per instruction. Closely following, Qwen2.5VL-72B (Ours) (one configuration) recorded $92.0\%$ accuracy at a cost of $\$0.00020$ per instruction.
Interestingly, another configuration of Qwen2.5VL-72B deployed with our framework (labeled as ``Qwen2.5VL-72B'' on the ``Ours'' trend line in Figure~\ref{fig:cost}) achieved $90.4\%$ accuracy at an even lower cost of $\$0.00011$ per instruction. This particular configuration not only surpasses the Qwen2.5VL-72B API (88.4\% ACC at $\$0.00015$) in accuracy but is also approximately $1.36$ times more cost-effective.
Other configurations, such as InternVL3-14B (Ours), delivered $91.4\%$ accuracy at $\$0.00016$ per instruction, while Qwen2.5VL-7B (Ours) achieved a strong $89.0\%$ accuracy at a cost of $\$0.000085$ per instruction. This is a notable improvement in accuracy over its API counterpart (Qwen2.5VL-7B API, $82.6\%$ ACC at $\$0.00007$), albeit at a slightly higher local inference cost.

While Gemini 2.5 Pro Preview (API) exhibited high accuracy ($89.3\%$), its API cost ($\$0.00075$) is considerably higher than our top-performing local configurations. For instance, InternVL3-78B (Ours) and the higher-performing Qwen2.5VL-72B (Ours) achieve superior accuracy (92.2\% and 92.0\%, respectively) and are approximately $3.4$ to $3.75$ times less expensive than Gemini 2.5 Pro Preview.
In contrast, widely used models like GPT-4o (API) and Claude 3.7 (API), despite their capabilities, incurred substantially higher costs per instruction ($\$0.015$ and $\$0.018$, respectively) for accuracies of $84.3\%$ and $81.2\%$ on this benchmark, accuracies which are surpassed by several of our framework's configurations at a fraction of the cost.

This analysis underscores the substantial economic advantages of leveraging optimized local inference strategies for state-of-the-art open-source VLMs. The results indicate that our approach can achieve or even surpass the accuracy of expensive commercial API-based services at significantly reduced costs. For example, our InternVL3-78B (Ours) configuration not only outperforms Gemini 2.5 Pro Preview in accuracy ($92.2\%$ vs. $89.3\%$) but does so at approximately $1/3.4\text{th}$ of the cost per instruction. Similarly, the Qwen2.5VL-72B (Ours) configuration achieving $92.0\%$ accuracy does so at about $1/3.75\text{th}$ of the cost of Gemini 2.5 Pro Preview, while also delivering higher accuracy. This highlights a pathway to democratize access to high-performance VLM capabilities, making advanced visual reasoning more accessible and economically viable. The significant cost reductions and performance enhancements observed for models like Qwen2.5VL and InternVL3 when run under our framework suggest that efficient utilization of local compute resources can unlock dramatic improvements in cost-efficiency without compromising.
\section{Experiment on Cost Dynamics During Multi-Round Debates}
\label{sec:cost_dynamics_multiround}

To investigate the economic implications of iterative debate, particularly how computational costs evolve across multiple rounds, we conducted a dedicated experiment. This analysis aims to compare the cumulative cost of our GAM-Agent framework against a baseline multi-agent debate scenario where original models engage in self-correction or iterative refinement without GAM-Agent's structured uncertainty-driven mechanisms. Understanding these cost dynamics is crucial for assessing the practical viability and efficiency of multi-round debate strategies.

\subsection{Experimental Setup}
\label{sec:cost_dynamics_setup}
The experiments were performed on the MMBench\_V11\_Test benchmark. We selected four prominent open-source Vision-Language Models (VLMs) as base models: Qwen2.5VL-7B, Qwen2.5VL-72B, InternVL3-14B, and InternVL3-78B.

We compared two primary configurations:
\begin{enumerate}
    \item \textbf{GAM-Agent Framework:} Our proposed GAM-Agent framework was applied using each of the selected base VLMs. The setup involved $N=3$ expert agents and $N_{crit}=3$ critic agents, consistent with other experiments.
    \item \textbf{Baseline Multi-Round Self-Debate (Base-Debate):} For this configuration, we simulated a multi-round debate scenario where instances of the original base VLM were prompted to iteratively refine their answers based on their previous outputs or simple peer feedback, without the structured uncertainty quantification, evidence mapping, or dynamic weighting of GAM-Agent. This represents a more direct or naive approach to multi-round refinement.
\end{enumerate}

For both configurations, we tracked the cumulative computational cost after 1, 2, and 3 rounds of debate. The "cost" was quantified by the average number of tokens processed per instance (input prompt tokens + generated output tokens) at the conclusion of each specified round. This metric serves as a proxy for both local GPU computation time and potential API call expenses. The primary goal of this experiment is to elucidate how GAM-Agent manages cost accumulation over successive debate rounds compared to a less structured iterative debate, and to highlight the efficiency benefits derived from its consensus mechanism which aims for early convergence.

\subsection{Experimental Results}
\label{sec:cost_dynamics_results}
\begin{figure}[t]
    \centering
    \includegraphics[width=\textwidth]{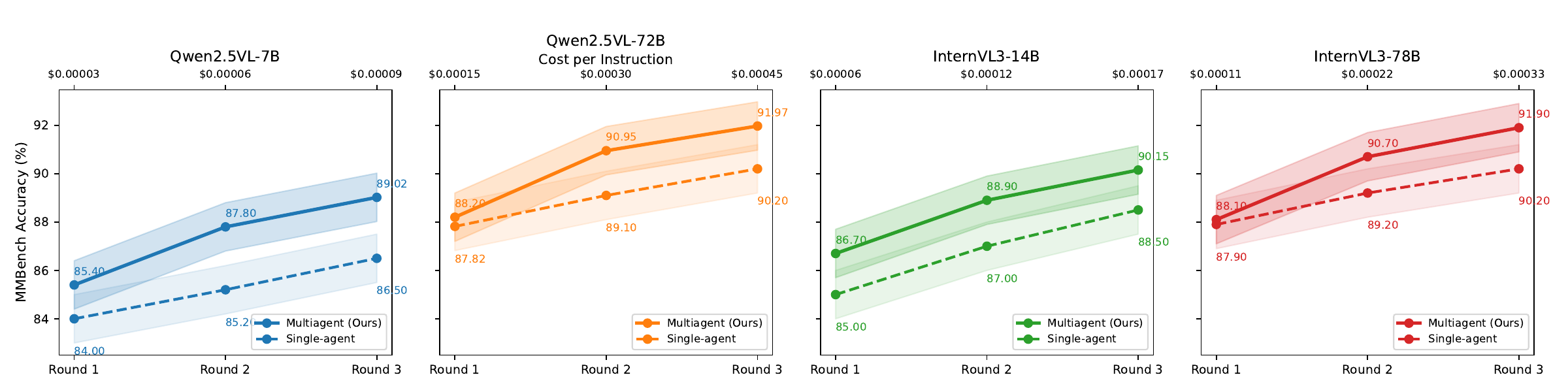} 
    \caption{Average cumulative cost (tokens per instance) on MMBench\_V11\_Test after 1, 2, and 3 rounds of debate for GAM-Agent (utilizing different base models) versus Baseline Multi-Round Self-Debate (Base-Debate).}
    \label{fig:cost_dynamics_debate_rounds}
\end{figure}

The average cumulative costs (in tokens per instance) for the GAM-Agent framework and the Baseline Multi-Round Self-Debate (Base-Debate) configuration across 1, 2, and 3 rounds of debate are presented in Figure~\ref{fig:cost_dynamics_debate_rounds}.

Figure~\ref{fig:cost_dynamics_debate_rounds} illustrates that while costs for GAM-Agent increase with more debate rounds, this is accompanied by significant accuracy improvements that consistently outperform the single-agent baseline. For instance, with InternVL3-78B as the base, GAM-Agent's average cost per instruction after 3 rounds is \$0.00033, at which point it achieves an accuracy of 91.90\%. In contrast, the Single-agent (Base-Debate) configuration with InternVL3-78B reaches an accuracy of 90.20\% after 3 rounds. GAM-Agent achieves superior accuracy (e.g., +1.7 percentage points for InternVL3-78B after 3 rounds) for a defined number of debate cycles.
This trend is consistent across other models. With Qwen2.5VL-72B, GAM-Agent achieves 91.82\% accuracy at a cost of \$0.00045 after 3 rounds, while the single-agent baseline reaches 90.20\%. For Qwen2.5VL-7B, GAM-Agent reaches 89.02\% accuracy for \$0.00009 after 3 rounds, compared to the baseline's 86.50\%. Similarly, for InternVL3-14B, GAM-Agent achieves 90.15\% for \$0.00017 after 3 rounds, surpassing the baseline's 88.50\%.
The GAM-Agent's structured debate, driven by uncertainty and conflict detection, aims to terminate debates earlier if consensus is reached, contributing to an optimized balance between cost and performance. While the Baseline Self-Debate might also incur increasing costs with more rounds, it consistently achieves lower accuracy. This suggests that to reach the accuracy levels of GAM-Agent, the baseline would likely require even more rounds, leading to significantly higher cumulative costs, or may not reach such accuracy levels at all within a comparable number of iterations. Thus, GAM-Agent's mechanisms not only improve accuracy but also offer a more efficient pathway to achieving high performance in multi-round visual reasoning by strategically focusing the debate and adaptively managing its length and cost.

\section{Deepening Comparative Analysis of Multi-Agent Methods: Uncertainty Handling Capability}
\label{sec:deep_uncertainty_comparison}

To comprehensively evaluate the uncertainty handling capabilities of our proposed GAM-Agent framework across various base models, we extended our analysis to include all major models from the Qwen2.5VL series (3B to 72B) and the InternVL3 series (2B to 78B). This allows us to observe the relationship between uncertainty processing capabilities and model scale, as well as the performance of GAM-Agent across different architectures.

\subsection{Comprehensive Analysis of Uncertainty Handling Capability}
\label{subsec:comprehensive_uncertainty_analysis}

We continue to use three key metrics to evaluate uncertainty-handling capabilities:

\begin{itemize}
    \item \textbf{Uncertainty Accuracy (UA$\uparrow$)}: GAM-Agent's ability to correctly identify instances of high uncertainty (higher is better).
    \item \textbf{Calibration Error (CE$\downarrow$)}: The consistency between the model's predicted confidence and its actual accuracy (lower is better).
    \item \textbf{Dynamic Adaptability (DA$\uparrow$)}: The ability to adjust the decision-making process based on uncertainty (higher is better).
\end{itemize}

\begin{figure}
    \centering
    \includegraphics[width=\linewidth]{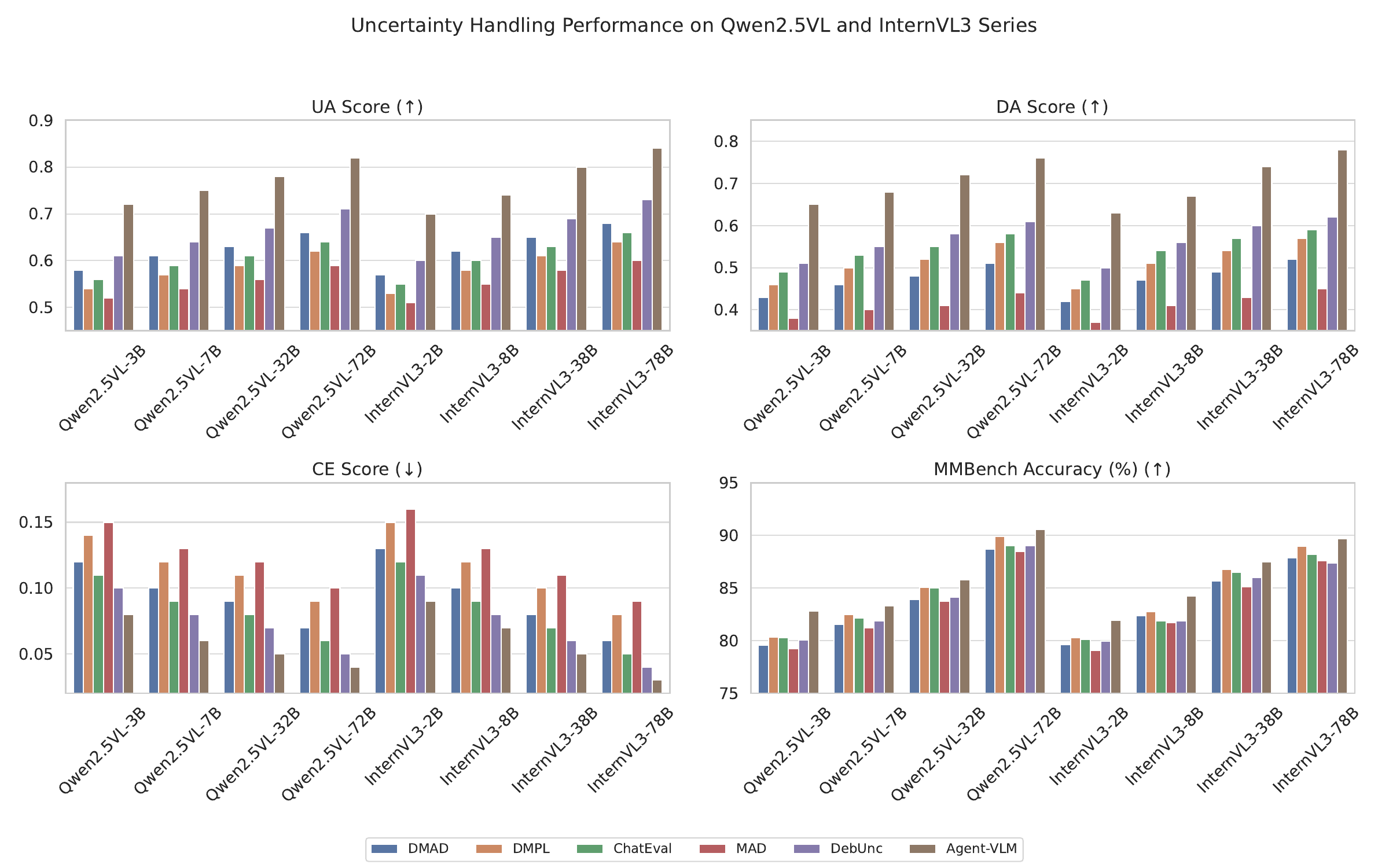}
    \caption{Comparison of uncertainty handling performance for various methods on the Qwen2.5VL and InternVL3 model series. UA, DA higher is better; CE lower is better.}
    \label{fig:qwenun}
\end{figure}

Figures~\ref{fig:qwenun}(Qwen2.5 VL series) and (InternVL3 series) show the performance of different multi-agent methods on these metrics and their MMBench accuracy.

\subsection{Experimental Setup}
\label{subsec:uncertainty_exp_setup}
The evaluation of uncertainty handling capabilities (UA, CE, DA) was conducted on the MMBench\cite{MVBench} dataset, which is also used for reporting the final accuracy. The primary motivation for this extended analysis is to systematically assess how different multi-agent frameworks, including our GAM-Agent, manage and leverage uncertainty, and how these capabilities scale with the underlying VLM's size and architecture.

For each base model in the Qwen2.5VL series (3B, 7B, 32B, 72B) and InternVL3 series (2B, 8B, 38B, 78B), we applied GAM-Agent and the baseline multi-agent methods (DMAD, DMPL, ChatEval, MAD, DebUnc). The baseline methods were adapted to the visual question answering task on MMBench 
The metrics were measured as follows:
\begin{itemize}
    \item \textbf{Uncertainty Accuracy (UA)}: This metric was evaluated by defining a threshold for high system-level uncertainty (e.g., based on entropy or semantic cues from the aggregated response). We then assessed the proportion of instances correctly flagged as uncertain (e.g., where the model's prediction was incorrect or the question was inherently ambiguous) against a ground truth or heuristic for uncertainty.
    \item \textbf{Calibration Error (CE)}: We utilized an Expected Calibration Error (ECE) variant. Model responses were binned by their expressed confidence scores (derived from uncertainty metrics like $1-U_{sys}$). The ECE was then calculated as the weighted average difference between the mean confidence and observed accuracy within each bin.
    \item \textbf{Dynamic Adaptability (DA)}: This was qualitatively and quantitatively assessed by observing whether the framework modified its reasoning process or resource allocation (e.g., triggering debate rounds in GAM-Agent, or adjusting agent weighting) in response to detected uncertainty, and whether such adaptations led to improved outcomes or more cautious predictions in uncertain scenarios. For a quantitative proxy, we measured the correlation between VLM's internal uncertainty signals and its engagement of deeper reasoning mechanisms or confidence adjustments.
\end{itemize}
The core idea is to investigate whether sophisticated uncertainty modeling, as implemented in GAM-Agent, translates into more reliable and robust performance, especially as model complexity and task difficulty increase.

\subsection{Experimental Results and Analysis}
\label{subsec:uncertainty_exp_results}

The results presented in Figure~\ref{fig:qwenun} offer a detailed view of the uncertainty handling capabilities of different multi-agent frameworks.

\textbf{GAM-Agent Consistently Excels in Uncertainty Management:}
Across both the Qwen2.5VL and InternVL3 series, and for all model sizes, GAM-Agent consistently outperforms the baseline methods (DMAD, DMPL, ChatEval, MAD, DebUnc) on all three uncertainty metrics: Uncertainty Accuracy (UA), Calibration Error (CE), and Dynamic Adaptability (DA). For example, with Qwen2.5VL-72B, GAM-Agent achieves a UA of 0.82, CE of 0.04, and DA of 0.76, significantly surpassing other approaches. A similar trend is observed with InternVL3-78B, where GAM-Agent scores 0.84 (UA), 0.03 (CE), and 0.78 (DA). This suggests that GAM-Agent's explicit uncertainty quantification, evidence mapping, and uncertainty-driven debate mechanisms are highly effective.

\textbf{Scalability with Model Size:}
A key observation is the positive scaling of GAM-Agent's uncertainty handling capabilities with increasing model size.
For the Qwen2.5VL series, as the model size increases from 3B to 72B, GAM-Agent's UA improves from 0.72 to 0.82, CE reduces from 0.08 to 0.04, and DA increases from 0.65 to 0.76.
Similarly, for the InternVL3 series (2B to 78B), GAM-Agent's UA rises from 0.70 to 0.84, CE drops from 0.09 to 0.03, and DA improves from 0.63 to 0.78.
This indicates that larger base models, when integrated into the GAM-Agent framework, not only provide better raw capabilities but also become more adept at identifying, calibrating, and adapting to uncertainty. While baseline methods also show some improvement with model size, their gains in uncertainty metrics are generally less pronounced than those achieved by GAM-Agent.

\textbf{Correlation with MMBench Accuracy:}
The data strongly suggests a positive correlation between superior uncertainty handling and higher overall task performance on MMBench. GAM-Agent, which consistently leads in UA, CE, and DA, also achieves the highest MMBench accuracy for each respective base model. For instance, GAM-Agent with Qwen2.5VL-72B not only has the best uncertainty scores but also the top MMBench accuracy of 90.56\% in Figure~\ref{fig:qwenun}. This underscores the importance of robust uncertainty management for achieving reliable and accurate visual reasoning. Frameworks that can better understand their own limitations and adapt accordingly are more likely to succeed in complex tasks.

\textbf{Performance of Baseline Methods:}
The baseline methods exhibit varied performance in uncertainty handling. DebUnc, which also focuses on uncertainty, generally performs second best to GAM-Agent on uncertainty metrics but still lags significantly. Other methods like DMAD, DMPL, ChatEval, and MAD show more modest capabilities in these specific uncertainty-related assessments. Their mechanisms for collaboration may not explicitly or effectively propagate and utilize fine-grained uncertainty signals to the same extent as GAM-Agent.

In conclusion, this expanded comparative analysis demonstrates GAM-Agent's superior and scalable uncertainty-handling capabilities. Its architecture, designed to explicitly model and debate uncertainty, enables it to achieve better calibration, more accurate identification of challenging instances, and more effective adaptation of its reasoning processes. These strengths in managing uncertainty are critical contributors to its leading performance on complex visual reasoning benchmarks like MMBench.
\subsection{Experimental Analysis}
\label{ssec:termination_exp_analysis_en}
\begin{figure}[H]
    \centering
    \includegraphics[width=0.8\textwidth, height=0.3\textheight, keepaspectratio, trim=0 80 0 0, clip]{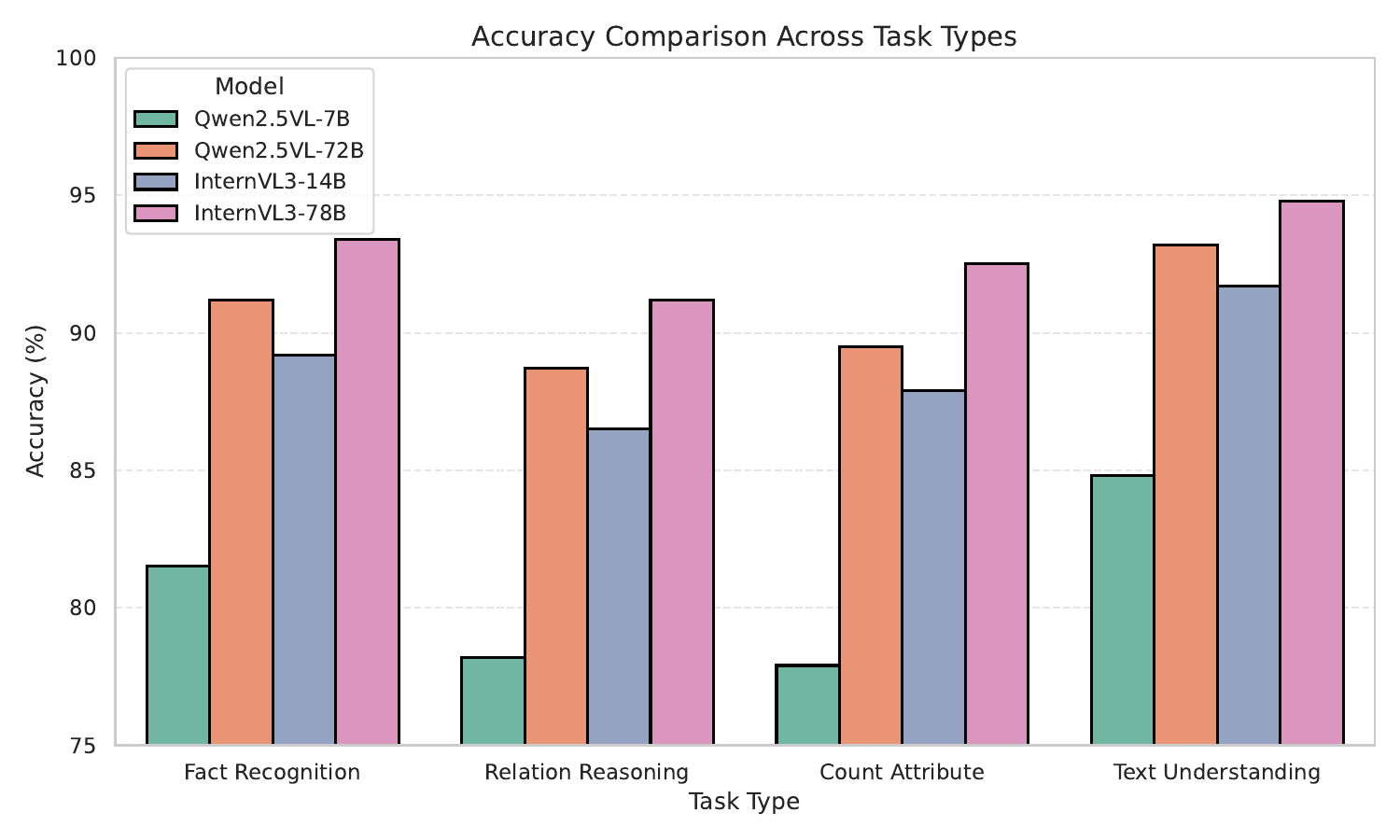}
    \caption{Performance and convergence dynamics under default termination conditions. This figure shows model accuracy across MMBench tasks, general performance trends, and key debate convergence statistics like average rounds and primary termination reasons.}
    \label{fig:termination_performance_convergence}
\end{figure}

The experimental results strongly confirm the effectiveness and robustness of the proposed debate termination conditions. As shown in Figure~\ref{fig:termination_performance_convergence} (your Figure 1, which you indicated is `boyi1.pdf`), all tested models exhibited a consistent accuracy distribution pattern across the various tasks in MMBench. Specifically, large-scale models (Qwen2.5VL-72B, InternVL3-78B) generally outperformed small-scale models (Qwen2.5VL-7B, InternVL3-14B) in terms of accuracy. When comparing models of similar scale, the InternVL3 series demonstrated a slight performance advantage over the Qwen2.5VL series. Regarding task types, text understanding tasks typically achieved the highest accuracy, while relational reasoning problems proved to be relatively more challenging.

\begin{figure}[H] 
    \centering
    \includegraphics[width=0.8\textwidth]{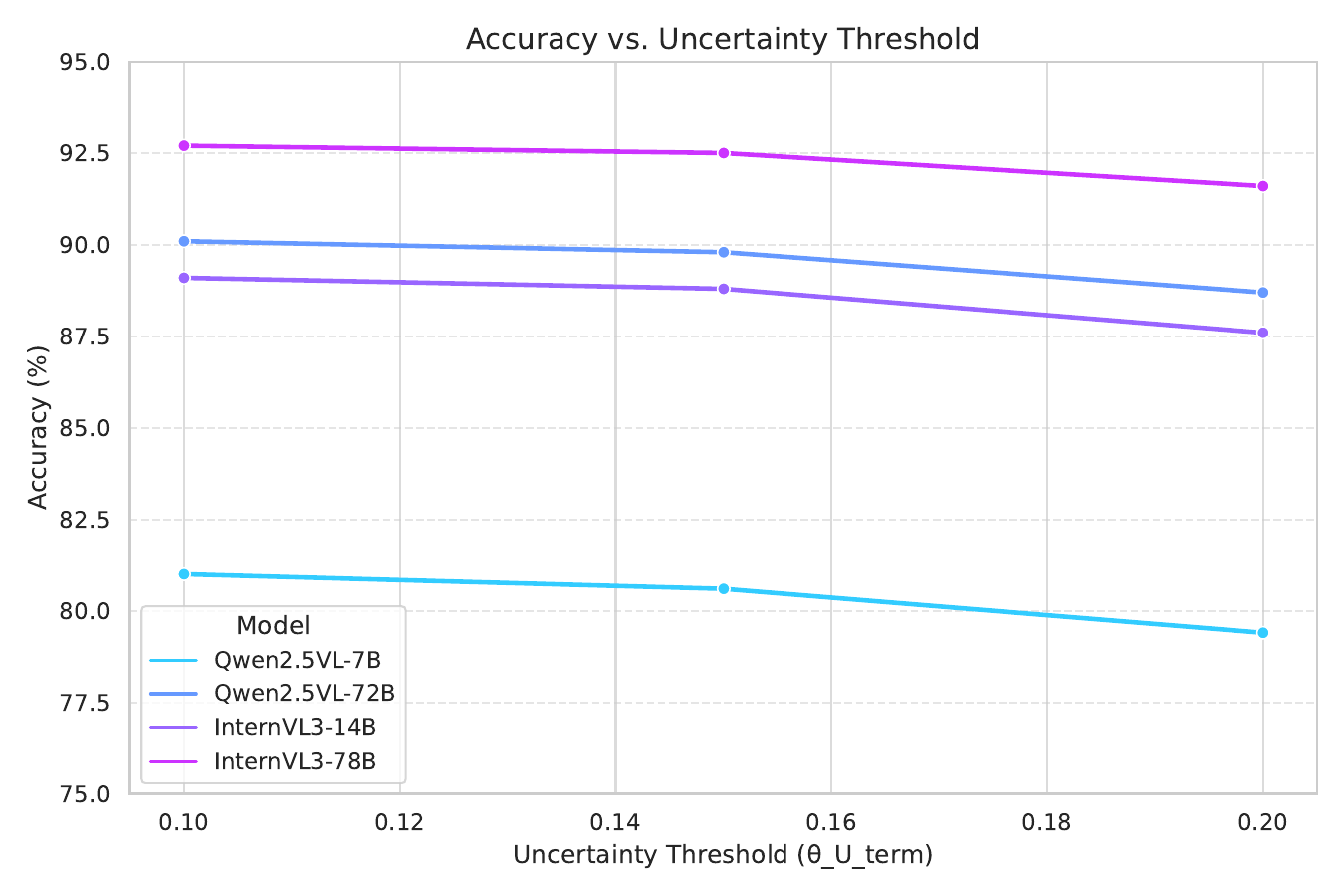} 
    \caption{Sensitivity analysis of the uncertainty termination threshold ($\theta_{U,term}$). This figure illustrates the trade-off between prediction accuracy and average debate rounds when varying $\theta_{U,term}$ (0.10, 0.15, 0.20), with $\theta_{U,term}=0.15$ identified as an optimal balance.}
    \label{fig:theta_sensitivity_analysis}
\end{figure}

Key experimental findings are summarized below:\textbf{Natural Convergence and Efficiency}: Under the default termination conditions, a high proportion of samples, ranging from 88\% to 94\%, achieved natural convergence before reaching the preset maximum number of debate rounds ($K_{\mathrm{max}} = 10$). On average, GAM-Agent required only 2.1 to 2.6 debate rounds to converge, with large-scale models exhibiting faster convergence.
\textbf{Distribution of Termination Reasons}: Triggering of the uncertainty threshold (i.e., system uncertainty $U_{\mathrm{sys}}^{(k)} < \theta_{U,\mathrm{term}}$) was the primary reason for debate termination, accounting for 63\% to 72\% of cases. The convergence speed threshold (i.e., change in system uncertainty $|\Delta U_{\mathrm{sys}}^{(k)}| < \epsilon$) was the second most common reason, responsible for approximately 22\% to 25\% of terminations. Only a minority of instances terminated due to reaching the maximum number of debate rounds.
\textbf{Task-Dependent Convergence Speed}: The convergence speed varied across different task types. Text understanding problems typically converged the fastest, requiring an average of only 1.7 to 2.1 debate rounds. In contrast, relational reasoning problems, owing to their complexity, converged relatively slowest, needing an average of 2.6 to 3.2 debate rounds.
Figure~\ref{fig:theta_sensitivity_analysis} (your Figure 2) further illustrates the impact of the uncertainty termination threshold $\theta_{U,term}$ on model accuracy and the average number of debate rounds.
When $\theta_{U,term}$ was set to 0.10 (a stricter convergence requirement), there was a slight improvement in accuracy (approximately +0.2\% to +0.4\%), but the average number of debate rounds increased significantly (by about 25\% to 31\%), leading to higher computational costs. Conversely, when $\theta_{U,term}$ was set to 0.20 (a looser convergence requirement), the average number of debate rounds decreased (by about 22\% to 26\%), but accuracy also declined (by approximately -0.9\% to -1.2\%). The experimental data indicate that all tested models achieve an optimal balance between prediction accuracy and computational efficiency at the default setting of $\theta_{U,term}=0.15$. Notably, large-scale models (72B/78B parameter scale) demonstrated lower sensitivity to variations in the $\theta_{U,term}$ parameter, suggesting greater robustness under different termination conditions.
In summary, these experimental results clearly demonstrate that our designed debate termination mechanism can effectively balance prediction accuracy and computational efficiency across vision language models of different scales and architectures. This provides reliable parameter setting guidance and performance expectations for the practical deployment of the GAM-Agent framework.

\subsection{Theoretical Analysis of Termination Threshold $\theta_{U,term}$ in Multi-Agent Debate}
The termination criterion in GAM-Agent's multi-agent debate framework plays a pivotal role in balancing system performance and computational efficiency. Drawing on our main experimental observations, this section provides a theoretical dissection of how the core uncertainty threshold parameter, $\theta_{U,term}$, influences the debate process.
Three principal conditions govern the debate termination mechanism (see Section 3.5.1 and Equation 5): (1) the system uncertainty $U_{sys}^{(k)}$ drops below the pre-set threshold $\theta_{U,term}$; (2) the debate reaches the maximum number of allowed iterations $K_{max}$; (3) the change in system uncertainty between consecutive rounds $|\Delta U_{sys}^{(k)}|$ falls below a convergence threshold $\epsilon$. Among these, $\theta_{U,term}$ theoretically defines the confidence requirement for terminating deliberation.
A deeper investigation into the $\theta_{U,term}$ parameter reveals its multifaceted role. From a Bayesian decision theory perspective, $\theta_{U,term}$ establishes a decision boundary, aiming to balance Type I errors (e.g., prematurely accepting a sub-optimal consensus) against Type II errors (e.g., unnecessarily prolonging the debate). Our primary experimental observations (see Section 4 and Appendix D.4) further expose several key convergence characteristics and sensitivities: First, the system uncertainty $U_{sys}^{(k)}$ typically decays in an approximately exponential manner during debate; Second, models with larger parameter scales (e.g., 72B/78B) exhibit lower sensitivity to changes in $\theta_{U,term}$, indicating more robust uncertainty estimation in larger models; Third, more complex tasks such as relational reasoning display higher sensitivity to the setting of $\theta_{U,term}$ compared to simpler tasks like text comprehension.
Despite the effectiveness of the current termination mechanism, several theoretical limitations\cite{2222222222} remain: (1) adopting a single global $\theta_{U,term}$ for all task types may not be optimal; (2) the mechanism does not dynamically adjust $\theta_{U,term}$ based on instance-specific factors (e.g., problem difficulty); (3) employing a fixed $\theta_{U,term}$ throughout the debate does not account for the dynamic evolution of uncertainty.
Based on the above theoretical analysis and main experimental findings, we offer the following theoretical insights for setting $\theta_{U,term}$: (1) Reasonable upper and lower bounds should be established—an excessively low $\theta_{U,term}$ may lead to unnecessary debate rounds and diminishing marginal returns, while an excessively high threshold risks premature termination, undermining the benefits of iterative deliberation; (2) Model-adaptive setting—in theory, as model capability increases, the optimal $\theta_{U,term}$ should decrease accordingly; (3) Task-adaptive setting—the ideal implementation should adjust $\theta_{U,term}$ based on task type, assigning stricter thresholds for more complex tasks

\section{Derivation of Uncertainty Theory}
\label{sec:derivation}
To provide a rigorous theoretical foundation for the collaborative reasoning process within the GAM-Agent framework, particularly concerning the role of uncertainty and dynamic consensus building\cite{555532131}, we model the interaction among expert agents as a consensus-driven non-zero-sum game. This game-theoretic perspective elucidates how agents converge towards a final consensus through interaction and dynamic adjustment of uncertainty.
Empirically, we observe that the uncertainty of predictions decays exponentially with debate rounds, which we quantify using a decay coefficient $\lambda$. As shown in Figure X, larger models such as InternVL3-78B exhibit higher $\lambda$ values and faster per-round uncertainty suppression (e.g., $\lambda=0.51$ with a 39.9\% decay rate), supporting the view that iterative multiagent consensus not only improves performance but also systematically reduces epistemic uncertainty.
\begin{figure}[htbp]
    \centering
    \includegraphics[width=0.7\textwidth]{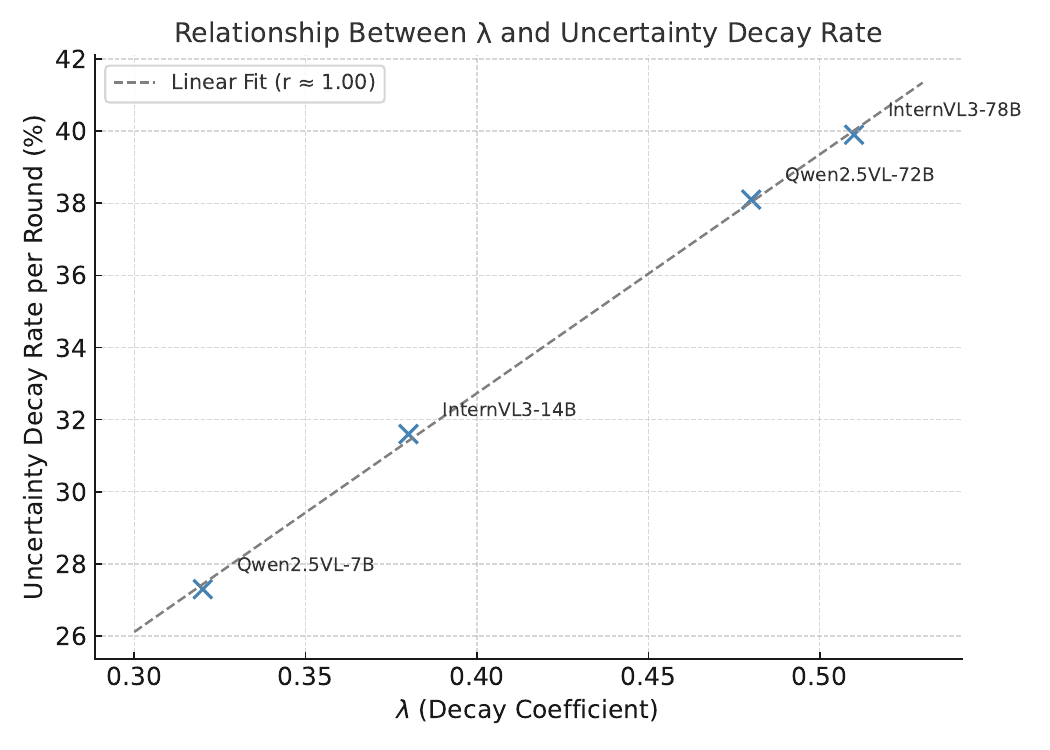}
    \caption{Exponential decay of uncertainty with debate rounds across models. Larger models (e.g., InternVL3-78B) exhibit faster uncertainty suppression.}
    \label{fig:uncertainty-decay}
\end{figure}
\subsection{Mathematical Formalization of the Non-Zero-Sum Game}
\label{subsec:formalization}

We consider a multi-agent system comprising $N$ expert agents. The system processes an input instance $I$, which consists of an image $X$ and a question $P_r$. The input space is formalized as the Cartesian product of the image space $\mathcal{X}$ and the question space $\mathcal{P}_r$, denoted as $\mathcal{I} = \mathcal{X} \times \mathcal{P}_r$. A specific input instance is $I = (X, P_r) \in \mathcal{I}$.
\textbf{Agent Set ($\mathcal{E}$):} The participants in the system form the set of agents, denoted as $\mathcal{E} = \{1, 2, \ldots, N\}$. This corresponds to the set $E$ in the GAM-Agent framework definition.

\medskip

\textbf{Agent Function (Agent Mapping $A_i$):} Each agent $i \in \mathcal{E}$ is formalized as a function or mapping that takes an input instance $I$ and produces an output pair, consisting of a response $R_i$ and an associated uncertainty $U_i$. The response space is denoted by $\mathcal{R}$, and the uncertainty space is the interval $[0, 1]$. The output space for agent $i$ is the Cartesian product $\mathcal{O}_i = \mathcal{R} \times [0, 1]$. Thus, the function of agent $i$ can be represented as a mapping:
\begin{equation}
    A_i: \mathcal{I} \to \mathcal{O}_i
\end{equation}
For a given input $I$, the output of agent $i$ is:
\begin{equation}
    \underbrace{(R_i(I), U_i(I))}_{\text{Output pair of Agent } i} = \underbrace{A_i(I)}_{\substack{\text{Mapping of Agent } i \\ \text{applied to input } I}}
\end{equation}
where $R_i(I) \in \mathcal{R}$ is the response and $U_i(I) \in [0, 1]$ is the uncertainty. This aligns with the analysis capability mapping $A_i$ and uncertainty assessment $\Phi_i$ described in the Method section ($A_i$ produces $R_i$, and $\Phi_i(R_i)$ yields $U_i$).

\medskip

\textbf{Joint Agent Output (Joint Agent Output Mapping $A$):} The collective output of the entire agent set for the same input $I$ can be viewed as a combined mapping. Its output space is the Cartesian product of all individual agent output spaces: $\mathcal{O} = \prod_{i=1}^N \mathcal{O}_i = (\mathcal{R} \times [0, 1])^N$. The joint output mapping $A$ is defined as:
\begin{equation}
    A: \mathcal{I} \to \mathcal{O}
\end{equation}
For input $I$, the joint output is:
\begin{equation}
    \underbrace{(A_1(I), \ldots, A_N(I))}_{\text{Joint Agent Output}} = \underbrace{((R_1(I), U_1(I)), \ldots, (R_N(I), U_N(I)))}_{\substack{\text{Tuple of individual Agent outputs}}}
\end{equation}

\medskip

\textbf{System Weight Allocation Protocol ($\mathcal{P}$):} The system dynamically assigns weights $\mathbf{w} = (w_1, \ldots, w_N)$ based on the joint agent output, particularly their uncertainties $\mathbf{U} = (U_1, \ldots, U_N)$ and potentially their responses $\{R_i\}$. The weight vector $\mathbf{w}$ belongs to the standard $(N-1)$-dimensional simplex $\Delta^{N-1}$. We formalize the weight allocation mechanism as a protocol function $\mathcal{P}$ that takes the joint agent output as input and produces a weight vector:
\begin{equation}
    \mathcal{P}: \mathcal{O} \to \Delta^{N-1}
\end{equation}
Thus, for input $I$, the weights assigned by GAM-Agent are $\mathbf{w}(I) = \underbrace{\mathcal{P}(A(I))}_{\substack{\text{Weights computed by the protocol} \\ \text{based on Agent outputs}}}$. 

The weight vector must satisfy the constraints of the standard simplex:
\begin{equation}
    \underbrace{w_i(I) \geq 0}_{\substack{\text{Each weight} \\ \text{is non-negative}}} \quad \forall i \in \mathcal{E} \quad \text{and} \quad \underbrace{\sum_{i=1}^{N} w_i(I)}_{\substack{\text{Sum of} \\ \text{all weights}}} = \underbrace{1}_{\substack{\text{equals 1}}}
\end{equation}
This protocol corresponds to the dynamic weighting mechanism used in the initial integration and iterative debate phases of GAM-Agent.

\medskip

\textbf{System State ($\sigma$):} After receiving input $I$, the instantaneous state of GAM-Agent can be fully described by the joint output of the agents and the corresponding weight vector computed by GAM-Agentm. We define the system state as a tuple:
\begin{equation}
    \sigma(I) = \underbrace{((A_1(I), \ldots, A_N(I)), \mathcal{P}(A_1(I), \ldots, A_N(I)))}_{\substack{\text{Set of Agent output pairs and corresponding system weights}}}
\end{equation}
This state $\sigma(I)$ belongs to the state space $\mathcal{S}$:
\begin{equation}
    \sigma(I) \in \underbrace{\mathcal{O} \times \Delta^{N-1}}_{\text{System State Space } \mathcal{S}}
\end{equation}
In the dynamic collaborative process (iterative debate), the agents' internal mappings $A_i$ might adapt based on interaction, leading to an evolution of the state $\sigma(I)$ across different rounds.
\subsection{Utility Function Design}
\label{subsec:utility}

We design a utility function $\mathcal{U}_i$ for each agent $i \in \mathcal{E}$ to quantify its "payoff" or "performance score" in the current system state. Within the consensus-driven non-zero-sum game framework, an agent's individual objective is to maximize its utility value. Concurrently, because the utility function incorporates inter-agent dependencies and the overall system state, improvements in total utility can be achieved through effective collaboration among agents.

The input to the utility function $\mathcal{U}_i$ reflects the critical information influencing agent $i$'s payoff: the responses $\{R_j\}_{j=1}^N$ and uncertainties $\{U_j\}_{j=1}^N$ of all agents, along with the current system weight distribution $\mathbf{w}$. The utility function maps this information to a real value:
$\mathcal{U}_i: \underbrace{(\mathcal{R} \times [0, 1])^N}_{\substack{\text{Joint Agent Outputs} \\ \text{(Responses \& Uncertainties)}}} \times \underbrace{\Delta^{N-1}}_{\substack{\text{System} \\ \text{Weights}}} \to \mathbb{R}$

We define the utility function $\mathcal{U}_i$ for agent $i$ as a weighted sum of three primary components:
\[
\small
\mathcal{U}_i(\{R_j\}_{j=1}^N, \{U_j\}_{j=1}^N, \mathbf{w}) = 
\underbrace{\vphantom{\sum_{j \neq i}} \overbrace{w_i}^{\substack{\text{Agent } i \\ \text{Weight}}} \cdot \overbrace{(1 - U_i)}^{\substack{\text{Agent } i \\ \text{Confidence}}}}_{\substack{\text{Term 1: Individual Contribution}}} 
+ \underbrace{\lambda \sum_{j \neq i} \overbrace{w_j}^{\substack{\text{Other Agent } j \\ \text{Weight}}} \cdot \overbrace{\text{Sim}(R_i, R_j)}^{\substack{\text{Similarity between} \\ \text{Agent } i \text{ and } j}}}_{\substack{\text{Term 2: Collaboration Reward}}} 
- \underbrace{\gamma \overbrace{U_{\text{sys}}}^{\substack{\text{System Weighted} \\ \text{Average Uncertainty}}}}_{\substack{\text{Term 3: System Penalty}}}
\]
where \textbf{Semantic Similarity Function ($\text{Sim}$):} $\text{Sim}: \mathcal{R} \times \mathcal{R} \to [0, 1]$ measures the semantic relevance or consistency between two responses. $\text{Sim}(R_i, R_j) = 1$ indicates perfect semantic agreement, while $\text{Sim}(R_i, R_j) = 0$ signifies complete irrelevance. This function can be implemented using various methods, such as cosine similarity based on text embeddings, analysis of overlapping key evidence or claims, or comparison of structured outputs. It is typically assumed to be symmetric: $\text{Sim}(r_1, r_2) = \text{Sim}(r_2, r_1)$ for all $r_1, r_2 \in \mathcal{R}$.

\textbf{System Weighted Average Uncertainty ($U_{\text{sys}}$):} This key metric quantifies the overall uncertainty of the system's current consensus, defined as the weighted average of all agent uncertainties using their current weights:
\[
\small
U_{\text{sys}} = \underbrace{\sum_{k=1}^{N} \overbrace{w_k}^{\substack{\text{Agent } k \\ \text{Weight}}} \overbrace{U_k}^{\substack{\text{Agent } k \\ \text{Uncertainty}}}}_{\substack{\text{Weighted average of all Agent uncertainties} \\ \text{(since } \sum w_k=1)}}
\]
$U_{\text{sys}} \in [0, 1]$, where lower values indicate higher system confidence in the integrated consensus result.

\textbf{Hyperparameters ($\lambda, \gamma$):} $\lambda \in \mathbb{R}_{>0}$ is the weight coefficient for the collaboration reward term, encouraging agents to align with the system consensus. $\gamma \in \mathbb{R}_{>0}$ is the weight coefficient for the system uncertainty penalty term, penalizing all agents when the overall system uncertainty is high.

\textbf{Rationale and Significance of the Utility Function Design:} This utility function design reflects the core objectives and trade-offs in building a collaborative consensus system:

\textbf{Individual Contribution ($w_i (1 - U_i)$):} Rewards agents for providing high-quality (low uncertainty $U_i$, thus high confidence $1-U_i$) responses that are assigned high weight ($w_i$) by the system. This incentivizes agents to improve the quality and reliability of their responses.

\textbf{Collaboration Reward ($\lambda \sum_{j \neq i} w_j \cdot \text{Sim}(R_i, R_j)$):} Rewards agents for providing responses that are consistent ($\text{Sim}(R_i, R_j)$ is high) with those of other agents who are considered reliable (high weight $w_j$). This drives consensus by encouraging agents to consider and align with the views of highly weighted peers. $\lambda$ controls the importance of this collaborative alignment.

\textbf{System Penalty ($-\gamma U_{\text{sys}}$):} Penalizes all agents based on the overall weighted average uncertainty of the system. When the system as a whole is uncertain about the current consensus, every agent's utility decreases. This motivates agents not only to minimize their own uncertainty but also to contribute to reducing the system's overall uncertainty through effective collaboration (e.g., sharing evidence and resolving conflicts), as the system's health benefits everyone. $\gamma$ controls the focus on this systemic uncertainty.

\textbf{Non-Zero-Sum Characteristic:} Unlike traditional voting or simple averaging schemes which can implicitly resemble zero-sum scenarios (one agent's gain is another's loss), our design allows for collective improvement:
\begin{itemize}
    \item Increasing collaborative consistency (higher $\text{Sim}(R_i, R_j)$) can increase the collaboration reward term for multiple agents simultaneously, boosting their utilities without necessarily decreasing individual confidence.
    \item Reducing the overall system uncertainty (lower $U_{\text{sys}}$, e.g., through effective evidence exchange) decreases the system penalty term $-\gamma U_{\text{sys}}$, thereby increasing the utility for \textbf{all} agents.
\end{itemize}
These mechanisms allow the total utility $\sum_{i=1} ^N\mathcal {U}_i$ to potentially increase through effective inter-agent collaboration, highlighting the \textbf{positive-sum} potential of the game. This aligns better with the nature of multi-agent collaboration for complex problem-solving compared to zero-sum or fixed-sum games.

\subsection{Exploration of Nash Equilibrium Existence}
\label{subsec:nash}

In formal game theory, the Nash Equilibrium (NE) is a central concept for predicting stable outcomes of rational agent behavior. It describes a profile of pure strategies $\mathbf{s}^* = (s_1^*, \ldots, s_N^*) \in \mathcal{S} = \prod_{i=1}^N \mathcal{S}_i$, where $\mathcal{S}_i$ is the pure strategy space of agent $i$, such that no agent $i$ can improve its utility $\mathcal{U}_i(\mathbf{s})$ by unilaterally changing its strategy $s_i \in \mathcal{S}_i$, given that all other agents maintain their equilibrium strategies $\mathbf{s}_{-i}^* = (s_1^*, \ldots, s_{i-1}^*, s_{i+1}^*, \ldots, s_N^*)$. Formally, a NE $\mathbf{s}^*$ satisfies:
\begin{equation}
\underbrace{\mathcal{U}_i(s_i^*, \mathbf{s}_{-i}^*)}_{\substack{\text{Utility of Agent } i \\ \text{at equilibrium}}} \geq \underbrace{\mathcal{U}_i(s_i, \mathbf{s}_{-i}^*)}_{\substack{\text{Utility of Agent } i \\ \text{when deviating unilaterally}}} \quad \underbrace{\forall s_i \in \mathcal{S}_i}_{\substack{\text{for any strategy} \\ \text{of Agent } i}}, \underbrace{\forall i \in \mathcal{E}}_{\substack{\text{for all Agents}}}
\end{equation}
The best response correspondence $BR_i: \mathcal{S}_{-i} \to 2^{\mathcal{S}_i}$ maps the strategy profile of other agents $\mathbf{s}_{-i}$ to the set of agent $i$'s optimal pure strategies:
\begin{equation}
BR_i(\mathbf{s}_{-i}) = \underbrace{\{s_i' \in \mathcal{S}_i \mid \mathcal{U}_i(s_i', \mathbf{s}_{-i}) \geq \mathcal{U}_i(s_i, \mathbf{s}_{-i}) \quad \forall s_i \in \mathcal{S}_i\}}_{\substack{\text{Set of Pure Best Responses of Agent } i \text{ given } \mathbf{s}_{-i}}}
\end{equation}
A pure strategy Nash equilibrium $\mathbf{s}^*$ is a fixed point of the joint best response correspondence $BR: \mathcal{S} \to 2^{\mathcal{S}}$, where $BR(\mathbf{s}) = \prod_{i=1}^N BR_i(\mathbf{s}_{-i})$:
\begin{equation}
\mathbf{s}^* \in \underbrace{BR(\mathbf{s}^*)}_{\substack{\text{Joint Best Response}}}
\end{equation}

\textbf{Theorem 1 (Existence of Pure Strategy Nash Equilibrium):} Consider a game with a finite number of agents $N < \infty$. Under specific assumptions regarding the agents' strategy spaces $\mathcal{S}_i$ and utility functions $\mathcal{U}_i$, the non-zero-sum game described above possesses at least one pure strategy Nash equilibrium.

\textbf{Proof Outline}: The existence of a pure strategy NE is often established using Kakutani's Fixed Point Theorem applied to the joint best response correspondence $BR: \mathcal{S} \to 2^{\mathcal{S}}$. This requires satisfying the following conditions:
\textbf{1. Properties of the Domain Space:} The domain of $BR$, the joint pure strategy space $\mathcal{S} = \prod_{i=1}^N \mathcal{S}_i$, must be a non-empty, compact, and convex subset of a Euclidean space $\mathbb{R}^d$.
\[
    \mathcal{S} \neq \emptyset, \quad \underbrace{\mathcal{S}}_{\substack{\text{Joint strategy space}}} \text{ is compact}, \quad \underbrace{\mathcal{S}}_{\substack{\text{Joint strategy space}}} \text{ is convex}
\]
This necessitates that each agent's pure strategy space $\mathcal{S}_i$ is non-empty, compact, and convex.
\[
    \underbrace{\mathcal{S}_i \neq \emptyset}_{\substack{\text{Non-empty}}}, \quad \underbrace{\mathcal{S}_i \text{ is compact}}_{\substack{\text{Compact}}}, \quad \underbrace{\mathcal{S}_i \text{ is convex}}_{\substack{\text{Convex}}} \quad \underbrace{\forall i \in \mathcal{E}}_{\substack{\text{for all Agents}}}
\]
    In VLM systems, an agent's pure strategy $s_i$ determines how it generates its response $R_i \in \mathcal{R}$ and uncertainty $U_i \in [0, 1]$. The response space $\mathcal{R}$ (e.g., all possible text sequences) is typically discrete, vast, and non-convex, making it challenging to define a compact and convex strategy space $\mathcal{S}_i$ that fully captures the agent's complex behavior.
\begin{enumerate}
    \item \textbf{Simplified Strategy Spaces under Assumptions:} To meet the theorem's requirements, one might model agent strategies more simply:
    \begin{itemize}
        \item \textbf{Assumption 1.1 (Uncertainty Strategy):} Agent $i$'s pure strategy $s_i$ is solely its choice of reported uncertainty $U_i \in [0, 1]$. Then $\mathcal{S}_i = [0, 1]$, which is non-empty, compact, and convex.
        \item \textbf{Assumption 1.2 (Low-Dimensional Continuous Strategy):} Agent $i$'s pure strategy $s_i$ is determined by a low-dimensional continuous vector $\theta_i \in \Theta_i \subset \mathbb{R}^d$, representing internal parameters influencing its response and uncertainty. If $\Theta_i$ is non-empty, compact, and convex (e.g., a hypercube), and the agent's output $(R_i, U_i)$ is a continuous function of $s_i$, this requirement is met.
    \end{itemize}
    \item \textbf{Properties of the Mapped Values:} The value of the mapping $BR(\mathbf{s})$ (the joint best response set) must be non-empty and convex. This means that for each agent $i$, its best response set $BR_i(\mathbf{s}_{-i})$ must be non-empty and convex.
    \begin{equation}
        \underbrace{BR_i(\mathbf{s}_{-i}) \neq \emptyset}_{\substack{\text{Non-empty}}}, \quad \underbrace{BR_i(\mathbf{s}_{-i}) \text{ is a convex set}}_{\substack{\text{Convex Set}}} \quad \underbrace{\forall \mathbf{s}_{-i} \in \mathcal{S}_{-i}}_{\substack{\text{for any strategy profile} \\ \text{of other agents}}}, \underbrace{\forall i \in \mathcal{E}}_{\substack{\text{for all Agents}}}
    \end{equation}
    This property can be established by analyzing the agent's utility function $\mathcal{U}_i$ with respect to its own strategy $s_i$. Specifically, if $\mathcal{S}_i$ is non-empty, compact, and convex, and $\mathcal{U}_i(s_i, \mathbf{s}_{-i})$ is continuous and \textbf{quasi-concave} in $s_i$, then $BR_i(\mathbf{s}_{-i})$ is guaranteed to be non-empty and convex.
    \begin{equation}
        \underbrace{s_i \mapsto \mathcal{U}_i(s_i, \mathbf{s}_{-i})}_{\substack{\text{Utility of Agent } i \\ \text{w.r.t. its own strategy}}} \text{ is quasi-concave} \quad \forall i \in \mathcal{E}, \forall \mathbf{s}_{-i} \in \mathcal{S}_{-i}
    \end{equation}
    Proving quasi-concavity for $\mathcal{U}_i$ w.r.t. $s_i$ is a central challenge, especially when $s_i$ determines complex outputs $(R_i, U_i)$. It requires analyzing the convexity of the upper level sets $\{s_i \in \mathcal{S}_i \mid \mathcal{U}_i(s_i, \mathbf{s}_{-i}) \geq c\}$. Under simplifying assumptions (e.g., $s_i$ is choosing $U_i \in [0, 1]$ and $\mathcal{U}_i$ is quasi-concave in $U_i$), this condition can be met.

    \item \textbf{Continuity of the Mapping:} The correspondence $BR(\mathbf{s})$ must have a closed graph or, more strongly, be upper hemi-continuous (UHC). This implies that if a sequence of strategy profiles $\mathbf{s}^{(m)} \to \mathbf{s}$ and a corresponding sequence of best responses $\mathbf{s}'^{(m)} \in BR(\mathbf{s}^{(m)})$ converge to $\mathbf{s}'$, then $\mathbf{s}'$ must be a best response to $\mathbf{s}$, i.e., $\mathbf{s}' \in BR(\mathbf{s})$.
    \begin{equation}
        \underbrace{BR \text{ has a closed graph (or is UHC)}}_{\substack{\text{Closed Graph / Upper Hemi-continuous}}}
    \end{equation}
    If the strategy space $\mathcal{S}$ is compact and the utility functions $\mathcal{U}_i(\mathbf{s})$ are continuous on $\mathcal{S}$, then the best response correspondence $BR_i$ is UHC, and consequently, the joint correspondence $BR$ is also UHC. Thus, continuity of the utility functions is key.
\end{enumerate}
If all these conditions (non-empty, compact, convex joint strategy space; non-empty, convex-valued, UHC joint best response correspondence) are satisfied, Kakutani's Fixed Point Theorem guarantees the existence of a fixed point $\mathbf{s}^* \in BR(\mathbf{s}^*)$, which is a pure strategy Nash equilibrium\cite{Nash1950}. Rigorously verifying these conditions for complex VLM systems often requires simplified models or strong assumptions about agent behavior and interaction functions.

\textbf{Mixed Strategy Nash Equilibrium:} When proving the existence of a pure strategy NE is difficult (e.g., due to non-convex strategy spaces or non-quasi-concave utilities), one can consider mixed strategy Nash equilibria. A mixed strategy $\sigma_i$ for agent $i$ is a probability distribution over its pure strategy space $\mathcal{S}_i$. The space of mixed strategies for agent $i$ is $\Sigma_i = \Delta(\mathcal{S}_i)$. Under mixed strategies, agents maximize their expected utility. Nash's Theorem guarantees that a mixed strategy NE always exists for games with finite agents and finite pure strategy sets, or more generally, if pure strategy spaces are compact metric spaces and utility functions are continuous.
\begin{equation}
\underbrace{\Sigma_i = \Delta(\mathcal{S}_i)}_{\substack{\text{Mixed strategy space} \\ \text{for Agent } i}}, \quad \underbrace{\mathbf{\sigma} = (\sigma_1, \ldots, \sigma_N)}_{\substack{\text{Joint mixed} \\ \text{strategy}}}
\end{equation}
A mixed strategy NE $\mathbf{\sigma}^* = (\sigma_1^*, \ldots, \sigma_N^*)$ exists such that for all agents $i$:
\begin{equation}
E_{\mathbf{s} \sim \mathbf{\sigma}^*}[\mathcal{U}_i(\mathbf{s})] \geq E_{(s_i, \mathbf{s}_{-i}) \sim (\sigma_i, \mathbf{\sigma}_{-i}^*)}[\mathcal{U}_i(s_i, \mathbf{s}_{-i})] \quad \underbrace{\forall \sigma_i \in \Sigma_i}_{\substack{\text{for any mixed strategy} \\ \text{of Agent } i}}
\end{equation}
For the VLM game, if we consider the agent pure strategy space as a compact metric space (e.g., via representation learning mapping responses to a compact space) and utility functions are continuous, Nash's Theorem ensures the existence of a mixed strategy NE. While existence is more readily guaranteed, the interpretation and practical relevance of mixed strategies in VLM collaborative reasoning require further investigation.

In summary, proving the existence of a pure strategy NE in the VLM game requires rigorous mathematical analysis of the agents' pure strategy spaces and utility functions, often under simplifying assumptions that satisfy specific mathematical properties.

\subsection{Cooperative Game Objective and Dynamic Weighting Protocol}
\label{subsec:coop_weights}

The dynamic process in our system aligns with the philosophy of cooperative game theory, where agents collaborate to enhance the overall quality of the consensus. The system-level objective can be framed as maximizing the sum of all agents' utilities\cite{iidd}, i.e., the total utility $\mathcal{J}$:
\begin{equation}
\max_{\mathbf{w} \in \Delta^{N-1}} \mathcal{J}(\{R_i\}_{i=1}^N, \{U_i\}_{i=1}^N, \mathbf{w}) = \max_{\mathbf{w} \in \Delta^{N-1}} \underbrace{\sum_{i=1}^N \mathcal{U}_i(\{R_j\}, \{U_j\}, \mathbf{w})}_{\substack{\text{System Total Utility} \\ \text{(Sum of Individual Utilities)}}}
\end{equation}
Substituting the definition of $\mathcal{U}_i$ into the total utility $\mathcal{J}$:
\begin{equation}
\mathcal{J} = \sum_{i=1}^N \left[ w_i (1 - U_i) + \lambda \sum_{j \neq i} w_j \cdot \text{Sim}(R_i, R_j) - \gamma U_{\text{sys}} \right]
\end{equation}
Expanding and rearranging terms, using $\sum_{i=1}^N w_i = 1$ and $U_{\text{sys}} = \sum_{k=1}^N w_k U_k$:
\begin{align}
\mathcal{J} &= \underbrace{\sum_{i=1}^N w_i (1 - U_i)}_{\substack{\text{Sum of Individual} \\ \text{Contributions}}} + \underbrace{\lambda \sum_{i=1}^N \sum_{j \neq i} w_j \cdot \text{Sim}(R_i, R_j)}_{\substack{\text{Sum of Collaboration} \\ \text{Rewards}}} - \underbrace{\sum_{i=1}^N \gamma U_{\text{sys}}}_{\substack{\text{Sum of System} \\ \text{Penalties}}} \\
&= \left( \sum_{i=1}^N w_i - \sum_{i=1}^N w_i U_i \right) + \lambda \sum_{i=1}^N \sum_{j \neq i} w_j \cdot \text{Sim}(R_i, R_j) - N \gamma U_{\text{sys}} \\
&= (1 - U_{\text{sys}}) + \lambda \sum_{i=1}^N \sum_{j \neq i} w_j \cdot \text{Sim}(R_i, R_j) - N \gamma U_{\text{sys}} \\
&= 1 + \lambda \sum_{i=1}^N \sum_{j \neq i} w_j \cdot \text{Sim}(R_i, R_j) - (1 + N \gamma) U_{\text{sys}}
\end{align}
Substituting $U_{\text{sys}} = \sum_{k=1}^N w_k U_k$ and ignoring the constant term 1, the optimization problem becomes:
\begin{equation}
\max_{\mathbf{w} \in \Delta^{N-1}} \left[ \lambda \underbrace{\sum_{i=1}^N \sum_{j \neq i} w_j \cdot \text{Sim}(R_i, R_j)}_{\substack{\text{Total Weighted Similarity Term}}} - (1 + N \gamma) \underbrace{\sum_{k=1}^N w_k U_k}_{\substack{\text{System Weighted Uncertainty}}} \right]
\end{equation}
Rearranging the summation terms to isolate $w_k$:
\begin{align}
\sum_{i=1}^N \sum_{j \neq i} w_j \cdot \text{Sim}(R_i, R_j) &= \sum_{j=1}^N w_j \sum_{i \neq j} \text{Sim}(R_i, R_j) \\
&= \sum_{k=1}^N w_k \sum_{i \neq k} \text{Sim}(R_i, R_k) \quad \text{(relabeling index } j \text{ to } k \text{)}
\end{align}
The objective function becomes:
\begin{equation}
\max_{\mathbf{w} \in \Delta^{N-1}} \sum_{k=1}^N w_k \underbrace{\left[ \lambda \sum_{i \neq k} \text{Sim}(R_i, R_k) - (1 + N \gamma) U_k \right]}_{\substack{\text{Agent } k\text{'s "Cooperative Value Score"} \\ \text{Score}_k}}
\end{equation}
Let $\text{Score}_k = \lambda \sum_{i \neq k} \text{Sim}(R_i, R_k) - (1 + N \gamma) U_k$. For fixed $\{R_i\}$ and $\{U_i\}$, this is maximizing a linear function $\sum_{k=1}^N w_k \cdot \text{Score}_k$ over the weight simplex $\Delta^{N-1}$. To obtain a smooth, non-degenerate weight distribution, we introduce an entropy regularization term $-\frac{1}{\eta} \sum_{k=1}^N w_k \log w_k$ (with $\eta > 0$) to the objective:
\begin{equation}
\max_{\mathbf{w} \in \Delta^{N-1}} \underbrace{\left[ \sum_{k=1}^N w_k \cdot \text{Score}_k - \frac{1}{\eta} \sum_{k=1}^N w_k \log w_k \right]}_{\substack{\text{Entropy-Regularized} \\ \text{Total Utility}}}
\end{equation}
The solution to this standard convex optimization problem (found via KKT conditions) yields optimal weights $w_k^*$ in the form of a Softmax distribution:
\begin{equation}
w_k^* = \underbrace{\frac{\exp(\eta \cdot \text{Score}_k)}{\sum_{m=1}^N \exp(\eta \cdot \text{Score}_m)}}_{\substack{\text{Static Optimal Weight Form} \\ \text{(depends on Score}_k \text{)}}}
\end{equation}
Substituting the definition of $\text{Score}_k$:
\begin{equation}
\label{eq:static_optimal_weights}
w_k^* = \underbrace{\frac{\exp\left( \eta \left( \lambda \sum_{i \neq k} \text{Sim}(R_i, R_k) - (1 + N \gamma) U_k \right) \right)}{\sum_{m=1}^N \exp\left( \eta \left( \lambda \sum_{i \neq m} \text{Sim}(R_i, R_m) - (1 + N \gamma) U_m \right) \right)}}_{\substack{\text{Optimal weight for Agent } k \\ \text{(depends on all } R, U, \text{ and hyperparameters } \lambda, \gamma, \eta \text{)}}}
\end{equation}
This static optimal weight distribution $w_k^*$ depends on all agent responses, uncertainties, and hyperparameters, reflecting the full cooperative optimization goal.

\textbf{Actual Dynamic Weighting Protocol Used in GAM-Agent:} In our iterative GAM-Agent framework, the weights are updated dynamically at each round $k$ based primarily on the agents' uncertainties from that round. The protocol used is:
\begin{equation}
\label{eq:dynamic_weights}
w_i^{(k+1)} = \underbrace{\frac{e^{-\beta U_i^{(k)}}}{\sum_{j=1}^N e^{-\beta U_j^{(k)}}}}_{\substack{\text{Weight for round } k+1 \\ \text{(based on uncertainty from round } k \text{)}}} \quad (\beta > 0)
\end{equation}
Here, $\beta > 0$ controls the sensitivity to uncertainty. This corresponds to the weighting described in Section~\ref{sec:integration_conflict} and used implicitly in Section~\ref{sec:debate}.

\textbf{Connecting Static Optimality and the Dynamic Protocol:} We now demonstrate that the dynamic weighting protocol (Eq. \ref{eq:dynamic_weights}) is not an arbitrary heuristic but corresponds to the optimal solution of a simplified, entropy-regularized static optimization problem.
Consider a simplification of the total utility $\mathcal{J}$ where we ignore the collaboration reward ($\lambda = 0$) and the system penalty ($\gamma = 0$). The objective becomes maximizing the weighted sum of individual agent confidences:
\begin{equation}
\max_{\mathbf{w} \in \Delta^{N-1}} \sum_{i=1}^N \mathcal{U}_i \approx \max_{\mathbf{w} \in \Delta^{N-1}} \sum_{i=1}^N w_i (1 - U_i)
\end{equation}
Let the "score" for agent $i$ be its confidence $s_i = (1 - U_i)$. We want to maximize the weighted average score $\sum_{i=1}^N w_i s_i$ with entropy regularization $\frac{1}{\beta} \sum_{i=1}^N w_i \log w_i$ (note the sign change compared to minimization; maximizing score + entropy):
\begin{equation}
\max_{\mathbf{w} \in \Delta^{N-1}} \left[ \sum_{i=1}^N w_i (1 - U_i) + \frac{1}{\beta} \sum_{i=1}^N w_i \log w_i \right]
\end{equation}
The optimal solution $w_i^*$ for this entropy-regularized maximization problem takes the Softmax form based on the score $(1 - U_i)$:
\begin{align}
w_i^* &= \underbrace{\frac{\exp(\beta \cdot (1 - U_i))}{\sum_{j=1}^N \exp(\beta \cdot (1 - U_j))}}_{\substack{\text{Static optimal weight form} \\ \text{(based on confidence score)}}} \\
&= \underbrace{\frac{\exp(\beta) \exp(-\beta U_i)}{\sum_{j=1}^N \exp(\beta) \exp(-\beta U_j)}}_{\substack{\text{Exponential form expansion}}} \\
&= \underbrace{\frac{e^{-\beta U_i}}{\sum_{j=1}^N e^{-\beta U_j}}}_{\substack{\text{Simplified form}}}
\end{align}
The resulting optimal weight $w_i^*$ has exactly the same form as the dynamic weighting protocol used in GAM-Agent (Eq. \ref{eq:dynamic_weights}), with $\beta$ playing the role of the inverse temperature parameter $\eta$.

$$\underbrace{w_i^*}_{\substack{\text{Static optimal weight based} \\ \text{on regularized confidence}}} = \underbrace{\frac{e^{-\beta U_i}}{\sum_{j=1}^N e^{-\beta U_j}}}_{\substack{\text{Matches the dynamic} \\ \text{protocol form}}}$$

This derivation shows that the dynamic weighting rule employed in GAM-Agent, $w_i^{(k+1)} = \frac{\exp(-\beta U_i^{(k)})}{\sum_j \exp(-\beta U_j^{(k)})}$, corresponds precisely to the optimal weight distribution derived from maximizing the entropy-regularized weighted average of agent confidences $(1-U_i^{(k)})$ at each step $k$. While this dynamic protocol simplifies the full cooperative objective (Eq. \ref{eq:static_optimal_weights}) by omitting explicit collaboration rewards and system penalties in the weight calculation, it provides a computationally efficient and theoretically grounded mechanism. It prioritizes influence for agents demonstrating lower uncertainty (higher confidence) at each stage, driving the system towards a consensus state characterized by reduced overall uncertainty, as explored in the convergence analysis related to the debate termination condition (Eq. \ref{eq:terminate-debate}).

\section{Theorem I: Fundamental Game Theoretic Framework of GAM-Agent}
The core mechanism of GAM-Agent is driven by a mathematically well-defined, non-zero-sum game model centered on "uncertainty" as a key variable. This model derives its collaborative strategies and influences allocation through the optimization of explicit utility functions\cite{2898756}.

\begin{definition}[Mathematical Formalization of GAM-Agent $S$]
Our GAM-Agent $S$ is rigorously defined as a sextuple:
\[S = (\underbrace{E}_{\substack{\text{Agent} \\ \text{Set}}}, \underbrace{A}_{\substack{\text{Analytical Capability} \\ \text{Map Function Set}}}, \underbrace{\Phi}_{\substack{\text{Uncertainty} \\ \text{Assessment Function Set}}}, \underbrace{M}_{\substack{\text{Evidence Localization} \\ \text{Function Set}}}, \underbrace{P}_{\substack{\text{Claim Parsing} \\ \text{Function Set}}}, \underbrace{D}_{\substack{\text{Dynamic Debate} \\ \text{Mechanism}}})\]
where:$E = {1, 2, \ldots, N}$ represents the set of N expert agents. $A = {A_1, A_2, \ldots, A_N}$ defines the set of analytical capability mapping functions, where each function $A_i: X \times P_r \rightarrow R_i$ maps an input image $X$ and a question $P_r$ to agent $i$'s response $R_i$. $\Phi = {\Phi_1, \Phi_2, \ldots, \Phi_N}$ is the set of uncertainty assessment functions, where $\Phi_i: R_i \rightarrow [0, 1]$ quantifies the uncertainty $U_i$ associated with response $R_i$. $M = {M_1, M_2, \ldots, M_N}$ represents the set of evidence localization functions, where $M_i: C_i \times X \rightarrow V \times [0, 1]$ associates agent $i$'s claim $c$ with a visual region $r$ and a confidence score $\sigma$. $P: R \rightarrow {(c_j, \sigma_j, e_j, r_j)}{j=1}^K$ is the claim parsing function, converting unstructured responses into structured information tuples (K is the number of tuples). $D: X \times P_r \times {R_i, U_i}{i=1}^N \rightarrow R_{\text{final}}$ is the debate module that coordinates conflicts and consensus to generate a final response.
This formalization constructs a complete mathematical space, allowing for precise description and analysis of the system's components and their interactions.
\end{definition}

\begin{definition}[Mathematical Construction of Uncertainty Quantification Function $\Phi_i(R_i)$]
GAM-Agent provides two uncertainty quantification mechanisms for each agent $e_i \in E$:
\begin{itemize}
    \item \textbf{Uncertainty based on generation probability $\Phi_{i\text{gen+}}(R_i)$:} When the token generation probability distribution of the underlying VLM is accessible, it is defined as:
    \[\Phi_{i\text{gen+}}(R_i) = \frac{1}{T_i}\sum_{t=1}^{T_i}\left[\alpha \cdot \underbrace{H(P_{i,t})}_{\substack{\text{Information Entropy} \\ \text{(Hesitation)}}} + \beta \cdot \underbrace{\max(0, 1-\Delta_{\text{top}}(P_{i,t}))}_{\substack{\text{Inverse Probability Difference} \\ \text{(Lack of Confidence)}}}\right]\]
    where: $ T_i$ is the total number of tokens in response $R_i$. $P_{i,t}$ is the probability distribution of the $t$-th token. $H(P_{i,t}) = -\sum_{x} P_{i,t}(x) \log P_{i,t}(x)$ is the entropy, quantifying the dispersion or "hesitation" of the distribution. $\Delta_{\text{top}}(P_{i,t}) = p_{i,t}^{(1)} - p_{i,t}^{(2)}$ is the difference between the probabilities of the top two tokens, where $p_{i,t}^{(k)}$ is the $k$-th highest probability. $\alpha$ and $\beta$ are parameters balancing these two complementary signals.
    High entropy $H(P_{i,t})$ indicates model hesitation among multiple possible tokens, while small $\Delta_{\text{top}}(P_{i,t})$ indicates a lack of strong confidence in its preferred token.

    By formally expanding the entropy calculation:
    \[H(P_{i,t}) = -\sum_{x \in V} P_{i,t}(x) \log P_{i,t}(x)\]
    where $V$ is the vocabulary, we see that entropy is maximized when the probability distribution $P_{i,t}$ approaches a uniform distribution (highest hesitation), and minimized when the distribution is concentrated on a single token (lowest hesitation).

    \item \textbf{Uncertainty based on semantic markers $\Phi_{i\text{sem}}(R_i)$:} When generation probabilities are unavailable, the system uses:
    \[\Phi_{i\text{sem}}(R_i) = \sigma_{\text{sigmoid}}\left(k \cdot \left(\frac{\sum_{w \in W} \text{weight}(w) \cdot \text{count}(w, R_i)}{|R_i|} - \text{offset}\right)\right)\]
    where:
 $W$ is a multi-level lexicon of uncertainty markers. $\text{weight}(w)$ reflects the uncertainty intensity of marker $w$. $\text{count}(w, R_i)$ is the frequency of $w$ in response $R_i$. $|R_i|$ is the response length. $k$ and $\text{offset}$ control scaling and bias. $\sigma_{\text{sigmoid}}(x) = \frac{1}{1+e^{-x}}$ normalizes the value to the $(0,1)$ interval.
\end{itemize}
These two methods capture uncertainty from the generation process and semantic content, respectively, providing key inputs for the subsequent game theory framework.
\end{definition}

\begin{definition}[Non-zero Sum Game Theoretic Formalization of GAM-Agent]
The multi-agent visual reasoning process is modeled as a non-zero-sum game with the following formal structure:
\textbf{Players:} Set of agents $\mathcal{E} = {1, \dots, N}$. \textbf{Input Instance:} $I = (X, P_r) \in \mathcal{I} = \mathcal{X} \times \mathcal{P}r$, where $X$ is the image and $P_r$ is the question. \textbf{Action Output of Agent $i$:} The action of agent $e_i$ is to generate a response-uncertainty pair $(R_i(I), U_i(I))$, with the mapping function: $A_i: \mathcal{I} \rightarrow \mathcal{O}i \equiv \mathcal{R} \times [0,1]$, $A_i(I) = (R_i(I), \Phi_i(R_i(I)))$. \textbf{System Weight Allocation Protocol $\mathcal{P}$:} Rules for assigning influence weights, formalized as: $\mathcal{P}: \mathcal{O}^N \rightarrow \Delta^{N-1}$. A key weight calculation formula (initial weight $w_i^{(0)}$) is: $w_i^{(0)}({U_j}{j=1}^N) = \frac{\exp(-\beta U_i)}{\sum{j=1}^{N} \exp(-\beta U_j)}$. This is an uncertainty-based Softmax allocation: higher $U_i$ (greater uncertainty) leads to lower weight $w_i^{(0)}$. Parameter $\beta > 0$ controls sensitivity to uncertainty differences. \textbf{System State $\sigma$:} The instantaneous state of the system is fully described by the joint outputs of the agents and the corresponding weight vector: $\sigma(I) = ((A_1(I), \ldots, A_N(I)), \mathcal{P}(A_1(I), \ldots, A_N(I)))$, $\sigma(I) \in \mathcal{O}^N \times \Delta^{N-1}$.
This game structure formalizes the interactions among agents in GAM-Agent and clarifies how uncertainty affects weight allocation and system state evolution.
\end{definition}

\begin{definition}[Mathematical Expansion of Agent Utility Function $u_i$]
A utility function $u_i: \mathcal{O}^N \times \Delta^{N-1} \rightarrow \mathbb{R}$ is defined for each agent $e_i$, mapping joint agent outputs and weight allocation to a real-valued utility:
\[u_i(\{R_j\}_{j=1}^N, \{U_j\}_{j=1}^N, w) = \underbrace{w_i(1-U_i)}_{\substack{\text{Individual Contribution Term} \\ \text{(Term 1)}}} + \underbrace{\lambda \sum_{j \neq i} w_j \cdot \text{Sim}(R_i, R_j)}_{\substack{\text{Collaborative Benefit Term} \\ \text{(Term 2)}}} - \underbrace{\gamma U_{\text{sys}}}_{\substack{\text{System Cost Term} \\ \text{(Term 3)}}}\]
where:
Individual contribution term (Term 1): $w_i(1-U_i)$ rewards the agent for providing high-confidence (low uncertainty) judgments, reflecting its system-assigned influence. Collaborative benefit term (Term 2): $\lambda \sum_{j \neq i} w_j \cdot \text{Sim}(R_i, R_j)$ rewards the agent for consistency with other high-influence agents, where $\text{Sim}(R_i, R_j) \in [0,1]$ is the semantic similarity between responses, and $\lambda > 0$ is a weight coefficient. System cost term (Term 3): $\gamma U_{\text{sys}}$, where $U_{\text{sys}} = \sum_{k=1}^N w_k U_k$ is the system's weighted average uncertainty, and $\gamma > 0$ is a weight coefficient. All agents share the system's uncertainty.
From a mathematical theory perspective, this utility function has the following properties:
Non-zero-sum nature: Agents can collectively increase total utility $\sum_{i=1}^N u_i$ by reducing collective uncertainty and improving consistency, without necessarily competing against each other. The incentive for collective reasoning: The collaborative benefit term incentivizes agents to agree with other highly influential agents. Uncertainty management: Agents must reduce their own uncertainty (Term 1) and contribute to reducing overall system uncertainty (Term 3).
By detailing the system cost term:$\gamma U_{\text{sys}} = \gamma \sum_{k=1}^N w_k U_k$
we see that even if an agent's own uncertainty $U_i$ is low, it will be negatively affected if the overall system still has high uncertainty (high $U_{\text{sys}}$), further promoting cooperative behavior.
\end{definition}
\begin{proposition}[Analytic Optimal $w_k^*$ for Regularized Utility]
Our GAM-Agent seeks optimal influence weights $w^* = (w_1^*, \ldots, w_N^*)$ that maximize the total utility:
\[
\mathcal{U}_{\text{total}}(w) = \sum_{i=1}^N u_i\big({R_j}_{j=1}^N, {U_j}_{j=1}^N, w\big)
\]
Expanding with Definition 1.4:
\begin{align*}
\mathcal{U}_{\text{total}}(w) &= \sum_{i=1}^N \Big[ w_i(1-U_i) + \lambda \sum_{j \neq i} w_j\,\text{Sim}(R_i, R_j) - \gamma \sum_{k=1}^N w_k U_k \Big] \\
\intertext{Further expansion:}
\end{align*}
Using the constraint $\sum_{i=1}^N w_i = 1$ to simplify:
\begin{align*}
\mathcal{U}_{\text{total}}(w) &= 1 - \sum_{i=1}^N w_i U_i + \lambda \sum_{i=1}^N \sum_{j \neq i} w_j \cdot \text{Sim}(R_i, R_j) - N\gamma \sum_{k=1}^N w_k U_k \\
&= 1 + \lambda \sum_{i=1}^N \sum_{j \neq i} w_j \cdot \text{Sim}(R_i, R_j) - (1+N\gamma) \sum_{k=1}^N w_k U_k
\end{align*}
Rearranging the summation order:
\[\sum_{i=1}^N \sum_{j \neq i} w_j \cdot \text{Sim}(R_i, R_j) = \sum_{j=1}^N w_j \sum_{i \neq j} \text{Sim}(R_i, R_j) = \sum_{k=1}^N w_k \sum_{i \neq k} \text{Sim}(R_i, R_k)\]
Substituting into the above equation and ignoring the constant term 1, we get the optimization problem:
\[\max_{w \in \Delta^{N-1}} \sum_{k=1}^N w_k \left[\lambda \sum_{i \neq k} \text{Sim}(R_i, R_k) - (1+N\gamma)U_k \right]\]
Define agent $k$'s "cooperative value score":
\[\text{Score}_k = \lambda \sum_{i \neq k} \text{Sim}(R_i, R_k) - (1+N\gamma)U_k\]
To avoid degenerate solutions (all weight assigned to the agent with the highest score), an entropy regularization term $-\frac{1}{\eta}\sum_{k=1}^N w_k \log w_k$ (where $\eta > 0$ is a temperature parameter) is introduced, yielding the regularized optimization problem:
$\max_{w \in \Delta^{N-1}} \left[\sum_{k=1}^N w_k \cdot \text{Score}k - \frac{1}{\eta}\sum{k=1}^N w_k \log w_k \right]$
This is a standard convex optimization problem. Using the Lagrange multiplier method, the analytical form of the optimal weights is:
$w_k^* = \frac{\exp(\eta \cdot \text{Score}k)}{\sum{m=1}^{N} \exp(\eta \cdot \text{Score}_m)}$
Substituting the full expression for $\text{Score}_k$:
\[w_k^* = \frac{\exp\left(\eta \cdot \left[\lambda \sum_{i \neq k} \text{Sim}(R_i, R_k) - (1+N\gamma)U_k\right]\right)}{\sum_{m=1}^{N} \exp\left(\eta \cdot \left[\lambda \sum_{i \neq m} \text{Sim}(R_i, R_m) - (1+N\gamma)U_m\right]\right)}\]
From this analytical solution, it is clear that:
Weight $w_k^*$ decreases exponentially with an increase in agent $k$\text{'s} uncertainty $U_k$. 
Weight $w_k^*$ increases exponentially with an increase in agent $k$\text{'s} semantic consistency with other agents, $\sum_{i \neq k} \text{Sim}(R_i, R_k)$. 
Parameter $\eta$ controls the sensitivity of weight allocation to differences in $\text{Score}_k$---high $\eta$ values lead to a more "winner-take-all" allocation.

This proves that agent influence allocation in GAM-Agent is an analytical solution with a clear mathematical basis, rather than a simple heuristic rule.
\end{proposition}
GAM-Agent establishes a mathematically self-consistent and mechanistically interpretable non-zero-sum game framework through rigorous mathematical definitions and derivations of the system, uncertainty, game interactions, utility functions, and optimization objectives. Its core mathematical contributions are: Transforming "uncertainty" from an abstract cognitive concept into a precisely quantifiable mathematical variable (via $\Phi_{i\text{gen+}}$ and $\Phi_{i\text{sem}}$). Designing a three-part utility function $u_i$ that balances individual contribution, collaborative consistency, and system risk. Analytically deriving the mathematical expression for agent influence weights $w_k^*$ by maximizing the regularized total utility $\mathcal{U}_{\text{total}}$. Demonstrating that the system's game mechanism is endogenously generated from utility optimization principles, rather than externally imposed.

\section{Theorem II: Comparative Analysis with the DMAD Framework}

The mathematical formalism of DMAD (Diverse Multi-Agent Debate) is primarily manifested as an algorithmic flow based on externally preset diversification strategies. Its core mechanism, "breaking cognitive fixation," mathematically lacks an endogenous, agent-centric utility optimization model equivalent to GAM-Agents to drive it.
\begin{definition}[Mathematical Representation of the DMAD Algorithm Flow]
The core of the DMAD framework guides $n$ agent instances $\{\mathcal{M}_i\}_{i=1}^n$ to each adopt a distinct reasoning method (prompting strategy) $\mathfrak{R}_i$ assigned from a preset collection $\mathbb{R} = \{\mathfrak{R}_1, \ldots, \mathfrak{R}_n\}$. Its mathematical flow can be formalized for each round $j$ of debate as:\textbf{Individual Solution Generation:} $\forall i \in {1, \ldots, n}: \quad (s_{i,j}, y_{i,j}) = \text{Execute}(\mathcal{M}i, x, h_i, \mathfrak{R}i)$ where: $\mathcal{M}i$ is the $i$-th agent model. $x$ is the input task. $h_i$ is agent $i$'s historical information. $\mathfrak{R}i$ is the preset, fixed reasoning method/prompting strategy. $(s{i,j}, y{i,j})$ is the output pair, containing the reasoning process $s{i,j}$ and answer $y{i,j}$. \textbf{Information Propagation and History Update:} Let $\mathbb{H}j = {(s{k,j}, y_{k,j})}_{k=1}^n$ be the set of outputs from all agents in round $j$. $\forall i \in {1, \ldots, n}: \quad h_i \leftarrow \text{UpdateFunction}(h_i, \mathbb{H}_j)$ This step propagates the outputs of all agents to each agent, updating their historical information.
\end{definition}

\begin{analysis}[Comparative Analysis of DMAD's Mathematical Form — The Missing Endogenous Game Optimization Aspect]
Compared to GAM-Agent (Theorem I), DMAD exhibits the following substantial differences in the mathematical formalization of its core mechanisms:
Compared to GAM-Agent (Theorem I), DMAD exhibits the following substantial differences in the mathematical formalization of its core mechanisms: \textbf{Lack of an optimizable, agent-level mathematical utility function:} The DMAD framework does not define a mathematical utility function for each agent $\mathcal{M}i$ similar to $u_i$ in GAM-Agent. Formally, there is no mapping: $u_i^{\text{DMAD}}: {\text{System State}} \rightarrow \mathbb{R}$ Therefore, when agent $\mathcal{M}i$ executes its assigned reasoning method $\mathfrak{R}i$, its behavior (e.g., how it interprets and utilizes information $h_i$ from other agents to improve its own $s{i,j+1}, y{i,j+1}$) is not driven by a clear optimization process aimed at maximizing its own mathematical utility. Its "improvement" relies more on the LLM's inherent context-learning capabilities and adherence to "critical feedback." In GAM-Agent, every decision can be traced back to the maximization of utility function $u_i$: $\max{a_i} u_i(a_i, a_{-i}, {U_j}{j=1}^N, {R_j}{j=1}^N)$ In DMAD, agent behavior is solely determined by the fixed strategy $\mathfrak{R}_i$ and historical information $h_i$, lacking this explicit optimization structure. \textbf{Mathematical implementation of the "diversity" mechanism:} DMAD's core idea, "breaking cognitive fixation through diverse reasoning," is achieved by externally enforcing different reasoning methods $\mathfrak{R}_i$ upon different agents. Mathematically, this is represented as a mapping: $\text{Assign}: {1, \ldots, n} \rightarrow \mathbb{R}$ $\text{Assign}(i) = \mathfrak{R}_i \text{ such that } \mathfrak{R}_i \neq \mathfrak{R}j \text{ if } i \neq j$ This "diversity" is a system-level structural design, not a result of individual agents selecting strategies based on some intrinsic mathematical incentive (like a "diversity contribution utility"). In contrast, the uncertainty and utility-driven mechanisms in GAM-Agent allow diversity to emerge naturally from the game--different agents gain higher weights in different aspects where their uncertainty is low, forming complementary rather than preset specializations. \textbf{Non-explicit mathematical modeling of "breaking cognitive fixation":} "Cognitive fixation" itself and the process of it "being broken" are core cognitive science assumptions and desired outcomes in DMAD, but they are not directly modeled mathematically as variables or objective functions that can be quantified or optimized in agent decision-making or system evolution. Formally, there is no function: $f{\text{mindset}}: {\text{System State}} \rightarrow \mathbb{R}$ Its effect is indirectly validated through experiments (e.g., performance improvements on different benchmarks), rather than predicted or explained through the analysis of a clearly defined mathematical objective function.

To more clearly illustrate the fundamental differences in mathematical construction between DMAD and GAM-Agent, we can compare their mathematical derivation processes for the same task:

For a visual reasoning task $(X, P_r)$:

\textbf{DMAD's Processing Flow:}
\begin{enumerate}
    \item Preset different reasoning methods: $\{\mathfrak{R}_1, \ldots, \mathfrak{R}_n\}$
    \item Assign to agents: $\mathcal{M}_i$ uses $\mathfrak{R}_i$
    \item Execute multi-round interaction: $(s_{i,j}, y_{i,j}) = \text{Execute}(\mathcal{M}_i, x, h_i, \mathfrak{R}_i)$
    \item Final aggregation: $y_{\text{final}} = \phi(\{y_{i,J}\}_{i=1}^n)$
\end{enumerate}

\textbf{GAM-Agent's Processing Flow:}
\begin{enumerate}
    \item Calculate uncertainty: $U_i = \Phi_i(R_i)$
    \item Optimize weights based on utility function $u_i$: $w_i^* = \frac{\exp(\eta \cdot \text{Score}_i)}{\sum_j \exp(\eta \cdot \text{Score}_j)}$
    \item Trigger debate based on uncertainty and conflict: $\text{TriggerDebate} = (U_{\text{sys}}^{(0)} > \theta_U) \vee (\text{ConflictScore} > \theta_C)$
    \item Dynamically adjust weights until convergence: $w_i^{(k+1)} = \frac{\exp(-\beta U_i^{(k)})}{\sum_j \exp(-\beta U_j^{(k)})}$
    \item Generate output based on final weights: $R_{\text{final}} = \text{IntegrateJudge}(\ldots, \{w_i^{(K)}\}_{i=1}^N, \ldots)$
\end{enumerate}
\end{analysis}

\textbf{Conclusion of Theorem II}: DMAD's mathematical form is mainly reflected in its clear, procedural algorithmic steps and information flow structure. Its core mechanism—leveraging preset, diverse reasoning strategies $\{\mathfrak{R}_i\}$ to enhance reasoning performance—is a heuristic design based on cognitive insights. Its mathematical description serves the execution of this process, but it lacks a mathematical game model equivalent to GAM-Agent's, which is agent-centric and uses explicit utility function optimization to endogenously drive agent behavior and system mechanisms (especially how "diversity" affects decisions).

\section{Theorem III: Differences in Mathematical Construction between GAM-Agent and Traditional Multi-Agent Methods}

GAM-Agent's mathematical framework fundamentally differs in its mathematical construction from traditional multi-agent reasoning and debate methods that do not rely on explicit game-theoretic optimization (hereinafter "traditional methods"), particularly in handling uncertainty, mechanism derivation, and collaborative optimization.
\begin{definition}[Mathematical Characteristics of Traditional Multi-Agent Reasoning]
\textbf{Simple Aggregation Methods:} Mathematical core: Typically involves direct, non-interactive mathematical operations on outputs ${y_1, \ldots, y_N}$ independently generated by $N$ agents. $Y_{\text{final}} = f_{\text{agg}}({y_i}{i=1}^N)$ where $f{\text{agg}}$ can be: $\text{Mode}(\cdot)$: Majority voting, selecting the most frequent answer. $\text{Average}(\cdot)$: Mean, for numerical outputs. $\sum_i \alpha_i y_i$: Fixed weighted average, where $\alpha_i$ are preset, non-dynamically optimized weights. Mathematical limitations: Such methods mathematically do not include: (a) An interaction model among agents. (b) Explicit consideration of individual output confidence or uncertainty. (c) A dynamic influence adjustment mechanism based on (a) and (b).
\textbf{Rule-Based/Scripted Debate Frameworks:} Mathematical core: Agent behavior and interaction are controlled by a set of predefined, deterministic rules $\mathcal{S}{\text{rules}}$, such as turn-taking or role-playing. Agent $e_i$'s action $a_i^t$ in its turn $t$ follows: $a_i^t = \text{Policy}{\text{role}(i), \text{state}(t)}(\text{history} | \mathcal{S}{\text{rules}})$ where: $\text{role}(i)$ is the preset role of agent $i$ (e.g., questioner, answerer, judge). $\text{state}(t)$ is the system state at turn $t$. $\text{history}$ is the prior interaction history. $\mathcal{S}{\text{rules}}$ is the set of rules governing interaction. Mathematical limitations: Interaction mechanisms (like who speaks or has influence) are hard-coded by external rules $\mathcal{S}_{\text{rules}}$, rather than endogenously derived from a mathematical optimization problem based on all agents' current internal states (especially quantified uncertainty $U_j$) and collaborative goals (like $u_j$).
\end{definition}
\textbf{Elaboration on Core Mathematical Differences with GAM-Agent:}
\textbf{Mathematization and Mechanistic Integration of Uncertainty Handling:}
\textbf{GAM-Agent:}
\[\underbrace{U_i = \Phi_i(R_i)}_{\substack{\text{Mathematically Quantified Uncertainty} \\ \text{(Definition 1.2)}}} \quad \xrightarrow{\text{input to}} \quad \underbrace{u_i(\ldots, U_i, \ldots)}_{\substack{\text{Utility Function} \\ \text{(Definition 1.4)}}} \quad \xrightarrow{\text{used for optimizing}} \quad \underbrace{w_k^*(\ldots, U_k, \ldots)}_{\substack{\text{Optimal Weights} \\ \text{(Proposition 1.1)}}}\]
GAM-Agent's uncertainty handling can be detailed as:
\begin{enumerate}
    \item First, precise mathematical quantification of uncertainty via $\Phi_{i\text{gen+}}$ or $\Phi_{i\text{sem}}$:
    \[U_i = \Phi_i(R_i) = \begin{cases}
    \frac{1}{T_i}\sum_{t=1}^{T_i}[\alpha H(P_{i,t}) + \beta \max(0, 1-\Delta_{\text{top}}(P_{i,t}))] & \text{if } \Phi_{i\text{gen+}} \text{ is used} \\
    \sigma_{\text{sigmoid}}(k \cdot (\frac{\sum_{w \in W} \text{weight}(w) \cdot \text{count}(w, R_i)}{|R_i|} - \text{offset})) & \text{if } \Phi_{i\text{sem}} \text{ is used}
    \end{cases}\]
    \item Then, direct integration of this quantified uncertainty into the agent's utility function:
$u_i({R_j}, {U_j}, w) = w_i(1-U_i) + \lambda \sum_{j \neq i} w_j \cdot \text{Sim}(R_i, R_j) - \gamma \sum_{k=1}^N w_k U_k$
    \item Finally, analytical derivation of optimal weight allocation by maximizing the utility function that includes uncertainty:
    \[w_k^* = \frac{\exp(\eta \cdot [\lambda \sum_{i \neq k} \text{Sim}(R_i, R_k) - (1+N\gamma)U_k])}{\sum_{m=1}^{N} \exp(\eta \cdot [\lambda \sum_{i \neq m} \text{Sim}(R_i, R_m) - (1+N\gamma)U_m])}\]
\end{enumerate}
This forms a complete, end-to-end mathematical processing chain for uncertainty, from quantification to utility calculation and mechanism design.

\textbf{Traditional Methods:}
Typically lack such end-to-end mathematical modeling and mechanistic integration of uncertainty.
For simple aggregation methods, uncertainty information is entirely missing, formalized as:
$Y_{\text{final}} = f_{\text{agg}}({y_i}_{i=1}^N) \quad \text{(does not depend on any uncertainty quantification } U_i \text{)}$
For rule-based debates, even if agents can express "confidence," this expression is often qualitative or heuristic:
$\text{Confidence}_i = \text{Qualitative\_Expression}(y_i) \quad \text{(e.g., ``I am very sure,'' ``Possibly,'' etc.)}$
Moreover, there is no unified mathematical framework (like GAM-Agent's utility theory and optimization framework) to convert these discrete, potentially heterogeneous confidence signals into precise, derivable mathematical impacts on debate dynamics (like speaking rights, influence) or final consensus.

\paragraph{D.4.2 Mathematical Basis of Interaction Mechanisms and Influence Allocation:}
\textbf{GAM-Agent:}
Influence weights $w_k^*$ (Proposition 1.1) are the analytical solution to the mathematical optimization problem of maximizing the system's overall regularized utility:
\[w_k^* = \frac{\exp(\eta \cdot \text{Score}_k)}{\sum_{m=1}^{N} \exp(\eta \cdot \text{Score}_m)}\]
where $\text{Score}_k = \lambda \sum_{i \neq k} \text{Sim}(R_i, R_k) - (1+N\gamma)U_k$ is agent $k$'s "cooperative value score."
This analytical solution has the following significant characteristics:
\begin{itemize}
    \item It is continuous, with weights $w_k^* \in (0,1)$ precisely reflecting the relative value of agents.
    \item It is dynamic, changing with $\text{Score}_k$ (which includes $U_k$ and $\text{Sim}$ terms).
    \item It is derived from first principles (maximizing total system utility), not preset.
    \item It considers two key factors: uncertainty $U_k$ and semantic consistency $\text{Sim}(R_i, R_k)$.
    \item It provides flexible adjustment mechanisms through parameters $\eta, \lambda, \gamma$.
\end{itemize}

\textbf{Traditional Methods:}
Influence allocation (if it exists) is often based on preset rules or heuristics:
For majority voting, each agent's influence is equal (implicitly $w_i = 1/N$):
\[Y_{\text{final}} = \arg\max_y \sum_{i=1}^N \mathbf{1}(y_i = y)\]
Or in some cases, it is binary (vote or not vote):
\[Y_{\text{final}} = \arg\max_y \sum_{i=1}^N \mathbf{1}(y_i = y) \cdot \mathbf{1}(\text{Qualification}_i)\]
In role-playing debates, if a judge role exists, its "influence" is conferred by the rules $\mathcal{S}_{\text{rules}}$:
\[Y_{\text{final}} = \text{Decision}(\text{Judge}) \quad \text{(where Judge is a preset role)}\]
Mathematically, the "influence" in these methods lacks the mathematical properties of GAM-Agent's weights $w_k^*$: it is not a continuous, dynamic variable derivable from first principles like utility maximization.

\paragraph{D.4.3 Mathematical Form and Explicitness of Collaborative Optimization Objectives:}
\textbf{GAM-Agent:}
The system's optimization objective is to maximize $\sum_i u_i$ (regularized). Each term of the utility function $u_i$ (Definition 1.4) has a clear mathematical form and corresponding collaborative goal:
\[u_i = \underbrace{w_i(1-U_i)}_{\text{Pursuing own high certainty}} + \underbrace{\lambda \sum_{j \neq i} w_j \text{Sim}(R_i, R_j)}_{\text{Aligning with trusted peers}} - \underbrace{\gamma \sum_k w_k U_k}_{\text{Jointly reducing system uncertainty}}\]
This utility function explicitly:
\begin{itemize}
    \item Quantifies three dimensions of collaboration: individual certainty, mutual consistency, and system uncertainty.
    \item Establishes precise mathematical trade-offs between these three dimensions (via parameters $\lambda$ and $\gamma$).
    \item Directly links utility maximization to system design goals (high-quality consensus).
\end{itemize}

\textbf{Traditional Methods:}
Usually lack such a unified, explicit mathematical optimization objective function that includes uncertainty management and fine-grained measures of collaboration quality.
For example, in voting methods, the implicit "objective function" might be:
\[\max_y \sum_{i=1}^N \mathbf{1}(y_i = y) \quad \text{(maximize the number of agents supporting a specific answer)}\]
This objective function does not include:
\begin{enumerate}
    \item Explicit consideration of uncertainty.
    \item Measurement of the quality of inter-agent consistency.
    \item Evaluation of the overall system state.
\end{enumerate}
In rule-based debates, the "collaborative" goal (like reaching the correct answer) is implicit, and the means to achieve it (like debate rules $\mathcal{S}_{\text{rules}}$) are preset, not derived by optimizing an explicit mathematical function.

\begin{conclusion}[Conclusion of Theorem IV]
GAM-Agent's mathematical framework, through its:
\begin{itemize}
    \item Precise mathematical quantification of agent uncertainty (e.g., $\Phi_{i\text{gen+}}$ formula) and its use as a core game variable,
    \item Analytical derivation of key interaction mechanisms (like the Softmax form of influence weights $w_k^*$) from an optimization problem involving a mathematical utility function ($u_i$) that includes uncertainty terms and explicit collaboration terms,
\end{itemize}
fundamentally diverges in theoretical rigor of mathematical construction, mechanistic endogeneity, and optimization explicitness from ``traditional'' multi-agent reasoning and debate methods that rely on simple aggregation, preset rules, or lack equivalent mathematical optimization objectives. GAM-Agent offers a mathematically-driven solution where each component and mechanism can be traced back to an explicit, uncertainty-based mathematical model, rather than just a procedurally-defined framework. This mathematical rigor enables GAM-Agent to more effectively coordinate uncertainty in multi-agent systems, achieving more flexible, adaptive, and interpretable visual reasoning.
\end{conclusion}

\section{Hyperparameter Ablation}
\label{sec:supplementary_hyperparameter_ablation}

To provide a deeper understanding of GAM-Agent's sensitivity to its internal settings, we conduct further ablation studies on key hyperparameters. These experiments supplement the component-level ablations presented in Section~\ref{sec:ablation} of the main paper. All experiments in this section were performed using \textbf{GAM-Agent (InternVL3-14B)} with $N=3$ expert agents on the \textbf{MMBench\_TEST\_V11} dataset. The maximum number of debate rounds ($K_{max}$) was set to 3, consistent with the setup in Section~\ref{sec:ablation_setup} of the main paper. We report Overall Accuracy (Acc.~\%), Average Actual Debate Rounds (Deb. Rounds), Debate Trigger Rate (Deb. Trig.~\%), and Average Inference Cost (Cost, tokens/instance).

Our default hyperparameter configuration for GAM-Agent (InternVL3-14B), which achieved 88.80\% accuracy in the main paper (see Figure~\ref{fig:ablation_results_condensed}), is as follows:
\begin{itemize}
    \item Uncertainty weighting sensitivity ($\beta_{weight}$): $1.5$
    \item Debate trigger threshold for system uncertainty ($\theta_U$): $0.45$
    \item Debate trigger threshold for conflict score ($\theta_C$): $0.55$
    \item Generation-process uncertainty $\Phi_{igen+}$ parameters: $\alpha_{\Phi} = 0.5$, $\beta_{\Phi} = 0.5$
\end{itemize}
Table~\ref{tab:hyperparameter_ablation_mmbench} presents the results of varying these hyperparameters.

\begin{table}[h!]
\centering
\caption{Hyperparameter ablation study for GAM-Agent (InternVL3-14B, $N=3$, $K_{max}=3$) on MMBench\_TEST\_V11. The \textbf{Default} row corresponds to the configuration used to achieve the main paper's reported 88.80\% accuracy for this model in Figure~\ref{fig:ablation_results_condensed} (main paper).}
\label{tab:hyperparameter_ablation_mmbench}
\resizebox{\textwidth}{!}{%
\begin{tabular}{@{}llcccc@{}}
\toprule
\textbf{Hyperparameter} & \textbf{Value} & \textbf{Acc. (\%)} & \textbf{Deb. Rounds} & \textbf{Deb. Trig. (\%)} & \textbf{Cost (tokens/inst.)} \\
\midrule
\multicolumn{6}{l}{\textit{Uncertainty Weighting Sensitivity ($\beta_{weight}$ in $w_i \propto \exp(-\beta_{weight} U_i)$)}} \\
 & 0.5 & 88.15 ($\downarrow$0.65) & 1.85 & 68 & 2650 \\
 & 1.0 & 88.62 ($\downarrow$0.18) & 1.80 & 66 & 2580 \\
\textbf{$\beta_{weight}$ (Default)} & \textbf{1.5} & \textbf{88.80} & \textbf{1.76} & \textbf{65} & \textbf{2500} \\
 & 2.0 & 88.71 ($\downarrow$0.09) & 1.73 & 64 & 2450 \\
 & 3.0 & 88.45 ($\downarrow$0.35) & 1.69 & 62 & 2380 \\
\midrule
\multicolumn{6}{l}{\textit{Debate Trigger Threshold - System Uncertainty ($\theta_U$)}} \\
 & 0.35 & 88.92 ($\uparrow$0.12) & 2.10 & 75 & 2850 \\
\textbf{$\theta_U$ (Default)} & \textbf{0.45} & \textbf{88.80} & \textbf{1.76} & \textbf{65} & \textbf{2500} \\
 & 0.55 & 88.31 ($\downarrow$0.49) & 1.42 & 50 & 2100 \\
 & 0.65 & 87.93 ($\downarrow$0.87) & 1.15 & 35 & 1800 \\
\midrule
\multicolumn{6}{l}{\textit{Debate Trigger Threshold - Conflict Score ($\theta_C$)}} \\
 & 0.45 & 88.85 ($\uparrow$0.05) & 1.95 & 72 & 2750 \\
\textbf{$\theta_C$ (Default)} & \textbf{0.55} & \textbf{88.80} & \textbf{1.76} & \textbf{65} & \textbf{2500} \\
 & 0.65 & 88.53 ($\downarrow$0.27) & 1.50 & 53 & 2200 \\
 & 0.75 & 88.10 ($\downarrow$0.70) & 1.25 & 40 & 1950 \\
\midrule
\multicolumn{6}{l}{\textit{Generation Uncertainty $\Phi_{igen+}$ Parameter $\alpha_{\Phi}$ (given $\alpha_{\Phi} + \beta_{\Phi} = 1$)}} \\
 & 0.1 ($\beta_{\Phi}=0.9$) & 88.42 ($\downarrow$0.38) & 1.78 & 65 & 2520 \\
 & 0.3 ($\beta_{\Phi}=0.7$) & 88.68 ($\downarrow$0.12) & 1.77 & 65 & 2510 \\
\textbf{$\alpha_{\Phi}$ (Default)} & \textbf{0.5 ($\beta_{\Phi}=0.5$)} & \textbf{88.80} & \textbf{1.76} & \textbf{65} & \textbf{2500} \\
 & 0.7 ($\beta_{\Phi}=0.3$) & 88.59 ($\downarrow$0.21) & 1.75 & 64 & 2490 \\
 & 0.9 ($\beta_{\Phi}=0.1$) & 88.27 ($\downarrow$0.53) & 1.74 & 64 & 2480 \\
\midrule
\multicolumn{6}{l}{\textit{Combined Debate Trigger Thresholds ($\theta_U, \theta_C$)}} \\
 & (0.35, 0.45) & 88.98 ($\uparrow$0.18) & 2.25 & 82 & 3050 \\
\textbf{(Default Combination)} & \textbf{(0.45, 0.55)} & \textbf{88.80} & \textbf{1.76} & \textbf{65} & \textbf{2500} \\
 & (0.55, 0.65) & 88.15 ($\downarrow$0.65) & 1.30 & 45 & 1900 \\
\bottomrule
\end{tabular}%
}
\end{table}

\paragraph{Discussion of Hyperparameter Sensitivity}

\begin{itemize}
    \item \textbf{Uncertainty Weighting Sensitivity ($\beta_{weight}$):} This parameter controls the sharpness of the softmax function used for allocating agent weights based on their uncertainty $U_i$. Our results suggest that a moderate value (Default: $1.5$) provides a good balance. Very low values (e.g., $0.5$) make the weights too uniform, diminishing the impact of precise uncertainty quantification and slightly reducing accuracy (88.15\%). Higher values (e.g., $2.0, 3.0$) make the weighting more "winner-take-all," potentially overly relying on a single agent if its uncertainty is marginally lower, which can also slightly degrade performance (88.71\% and 88.45\% respectively) by reducing diversity in the integration. The cost tends to decrease slightly with higher $\beta_{weight}$ as more decisive initial weighting might lead to quicker convergence or fewer contentious points triggering prolonged debates.

    \item \textbf{Debate Trigger Thresholds ($\theta_U, \theta_C$):}
    Lowering the system uncertainty threshold $\theta_U$ (e.g., to $0.35$) or the conflict score threshold $\theta_C$ (e.g., to $0.45$) increases the debate trigger rate (Deb. Trig. to 75\% and 72\%, respectively). This leads to more debate rounds and higher computational cost, but can sometimes yield marginal accuracy improvements (e.g., 88.92\% for $\theta_U=0.35$) by allowing the system to resolve more nuanced disagreements. Conversely, higher thresholds (e.g., $\theta_U=0.65$ or $\theta_C=0.75$) significantly reduce the debate frequency, rounds, and cost, but at the expense of accuracy (87.93\% and 88.10\%, respectively), as critical conflicts might be overlooked. Ablating them in combination (e.g., $\theta_U=0.35, \theta_C=0.45$) shows an even higher debate trigger rate (82\%) and cost, with a small potential accuracy gain (88.98\%), indicating that more debate is not always cost-effective for the performance gained. The default values (0.45, 0.55) provide a good trade-off.

    \item \textbf{Generation-Process Uncertainty Parameters ($\alpha_{\Phi}, \beta_{\Phi}$):} These parameters balance the contribution of information entropy (hesitation) and top probability difference (lack of conviction) in the $\Phi_{igen+}$ uncertainty metric (Equation~1 in the main paper). We assumed $\alpha_{\Phi} + \beta_{\Phi} = 1$ for this ablation. The results indicate that an approximately equal weighting (Default: $\alpha_{\Phi}=0.5, \beta_{\Phi}=0.5$) performs best (88.80\%). Overly relying on just one component (e.g., $\alpha_{\Phi}=0.1$ heavily weights probability difference, or $\alpha_{\Phi}=0.9$ heavily weights entropy) leads to a noticeable drop in accuracy, suggesting that both signals are valuable for a comprehensive uncertainty assessment. The impact on debate rounds and cost is minimal in this ablation, implying these parameters primarily affect the quality of the uncertainty scores rather than the frequency of debates, assuming the scores still fall within similar ranges.
\end{itemize}

In summary, these ablations highlight that while GAM-Agent is robust across a reasonable range of hyperparameter settings, optimal performance is achieved by carefully tuning the balance between uncertainty sensitivity, debate triggers, and the composition of uncertainty metrics. The default parameters chosen for InternVL3-14B in the main paper represent a well-balanced configuration for MMBench.

\section{Hyperparameter Setting}
\label{sec:supplementary_hyperparameter_settings}

This section outlines the specific hyperparameter configurations employed for GAM-Agent throughout the experiments detailed in the main paper. Our goal is to provide clarity for reproducibility and understanding of the conditions under which our reported results were achieved.

\paragraph{Common GAM-Agent Settings}
Unless explicitly stated otherwise in the subsections below, the following common hyperparameter settings were applied across all GAM-Agent instantiations:
\begin{itemize}
    \item \textbf{Number of Base Expert Agents ($N$):} Typically 3 for all reported experiments. This $N$ corresponds to the number of agents in summations such as $\sum_{i=1}^{N} w_i^{(0)} U_i$ and refers to the agents responsible for generating initial responses ($R_i$) and participating directly in argumentation ($Arg_i^{(k)}$) during debates. These agents were derived from the same base Vision-Language Model (VLM) for each experiment.
    \item \textbf{Number of Critical Expert Agents ($N_{crit}$):} Maintained at 3 by default throughout all experiments that involved the iterative debate mechanism. As illustrated in Figure~\ref{fig:pipeline} (main paper), these agents are engaged by the Debate Controller specifically for verification and providing critical feedback on contentious claims. Critical agents typically use the same underlying VLM but initialized with specialized prompts to foster critical assessment.
    \item \textbf{Uncertainty Weighting Sensitivity ($\beta_{weight}$):} $1.5$. This parameter is used in the entropy-regularized softmax for both initial weight allocation ($w_i^{(0)} \propto \exp(-\beta_{weight} U_i)$) and dynamic weight updates during the iterative debate ($w_i^{(k)} \propto \exp(-\beta_{weight} U_i^{(k)})$ when uncertainty-based weighting is active).
    \item \textbf{Generation-Process Uncertainty ($\Phi_{igen+}$) Parameters (used when token-level logprobs were accessible, e.g., for Qwen2.5VL, InternVL3, InternVideo2.5):}
        \begin{itemize}
            \item $\alpha_{\Phi}$: $0.5$ (weight for information entropy component).
            \item $\beta_{\Phi}$: $0.5$ (weight for top-K probability difference component).
        \end{itemize}
    \item \textbf{Semantic-Marker Based Uncertainty ($\Phi_{isem}$) Parameters (used when logprobs were not accessible, e.g., for GPT-4o-0513, or as a fallback):}
        \begin{itemize}
            \item Lexicon ($W$): A predefined multi-level lexicon of uncertainty markers (e.g., ``might'', ``possibly'', ``unsure", ``confident'', and ``clear'') with associated weights. (Specific lexicon details are part of prompt engineering).
            \item Sigmoid scaling $k$: $1.0$.
            \item Sigmoid offset: $0.3$.
        \end{itemize}
    \item \textbf{Debate Termination Criteria:}
        \begin{itemize}
            \item System uncertainty threshold ($\theta_{U, \text{term}}$): $0.15$. Debate terminates if $U_{\mathrm{sys}}^{(k)} < \theta_{U, \text{term}}$.
            \item Convergence speed threshold ($\epsilon$): $0.01$. Debate terminates if $|\Delta U_{\mathrm{sys}}^{(k)}| < \epsilon$.
        \end{itemize}
    \item \textbf{Evidence Mapping ($M$) and Claim Parsing ($P$):} These modules were active in all configurations involving debate, utilizing the base VLM to perform claim extraction, evidence association, and visual grounding tasks as described in Section~\ref{sec:evidence_parsing} (main paper).
\end{itemize}

\subsection{Settings for Image Understanding Experiments (MMBench and MMMU)}
These settings correspond to the experiments detailed in Section~4.1 of the main paper

\subsubsection{Qwen2.5VL Series (3B, 7B, 32B, 72B) and InternVL3 Series (2B, 8B, 14B, 38B, 78B)}
\begin{itemize}
    \item \textbf{Base VLMs:} As listed.
    \item \textbf{Datasets:} MMBench (TEST\_EN), MMMU (test set).
    \item \textbf{Number of Base Expert Agents ($N$):} 3.
    \item \textbf{Number of Critical Expert Agents ($N_{crit}$):} 3 (when debate triggered).
    \item \textbf{Uncertainty Quantification:} $\Phi_{igen+}$ was prioritized.
    \item \textbf{Maximum Debate Rounds ($K_{max}$):} $3$.
    \item \textbf{Debate Trigger Thresholds:}
        \begin{itemize}
            \item System uncertainty ($\theta_U$): $0.45$.
            \item Inter-expert conflict score ($\theta_C$): $0.55$.
        \end{itemize}
\end{itemize}
For instance, GAM-Agent(InternVL3-14B) using these settings achieved an accuracy of 88.80\% on MMBench (TEST\_EN), and GAM-Agent(Qwen2.5VL-72B) achieved 90.86\%.

\subsubsection{GPT-4o-0513}
\begin{itemize}
    \item \textbf{Base VLM:} GPT-4o-0513 (via OpenRouter API).
    \item \textbf{Datasets:} MMBench (TEST\_EN), MMMU (test set).
    \item \textbf{Number of Base Expert Agents ($N$):} 3.
    \item \textbf{Number of Critical Expert Agents ($N_{crit}$):} 3 (when debate triggered).
    \item \textbf{Uncertainty Quantification:} $\Phi_{isem}$ was used due to the lack of direct access to token-level generation probabilities from the API.
    \item \textbf{Maximum Debate Rounds ($K_{max}$):} $3$.
    \item \textbf{Debate Trigger Thresholds:}
        \begin{itemize}
            \item System uncertainty ($\theta_U$): $0.50$ (adjusted slightly due to different uncertainty scale from $\Phi_{isem}$).
            \item Inter-expert conflict score ($\theta_C$): $0.60$.
        \end{itemize}
\end{itemize}
With these settings, GAM-Agent(GPT-4o-0513) achieved 86.53\% on MMBench (TEST\_EN).

\subsection{Settings for Video Understanding Experiments (MVBench)}
These settings correspond to the experiments detailed in Section~4.2 of the main paper.

\subsubsection{Qwen2.5VL Series (3B, 7B, 32B, 72B), InternVL3 Series (2B, 8B, 38B), and InternVideo2.5}
\begin{itemize}
    \item \textbf{Base VLMs:} As listed. InternVideo2.5 is a video-specialized model.
    \item \textbf{Dataset:} MVBench.
    \item \textbf{Number of Base Expert Agents ($N$):} 3.
    \item \textbf{Number of Critical Expert Agents ($N_{crit}$):} 3 (when debate triggered).
    \item \textbf{Uncertainty Quantification:} $\Phi_{igen+}$ was prioritized. For InternVideo2.5, if logprobs were accessible, $\Phi_{igen+}$ was used; otherwise, $\Phi_{isem}$ was adapted.
    \item \textbf{Maximum Debate Rounds ($K_{max}$):} $3$.
    \item \textbf{Debate Trigger Thresholds:}
        \begin{itemize}
            \item System uncertainty ($\theta_U$): $0.50$ (video tasks often present higher inherent ambiguity).
            \item Inter-expert conflict score ($\theta_C$): $0.60$.
        \end{itemize}
    \item \textbf{Visual Preprocessing: Please refer to the official parameter configuration provided by MVbench, for example, select fps=1}.
\end{itemize}
GAM-Agent (InternVL3-38B) achieved an overall average of 78.47\% on MVBench with these settings.

\subsubsection{GPT-4o-0513 (for MVBench)}
\begin{itemize}
    \item \textbf{Base VLM:} GPT-4o-0513.
    \item \textbf{Dataset:} MVBench.
    \item \textbf{Number of Base Expert Agents ($N$):} 3.
    \item \textbf{Number of Critical Expert Agents ($N_{crit}$):} 3 (when debate triggered).
    \item \textbf{Uncertainty Quantification:} $\Phi_{isem}$.
    \item \textbf{Maximum Debate Rounds ($K_{max}$):} $3$.
    \item \textbf{Debate Trigger Thresholds:}
        \begin{itemize}
            \item System uncertainty ($\theta_U$): $0.55$.
            \item Inter-expert conflict score ($\theta_C$): $0.65$.
        \end{itemize}
\end{itemize}
GAM-Agent(GPT-4o-0513) achieved 70.58\% on MVBench.

\subsection{Settings for Comparison with Existing Multi-Agent Methods (MMBench)}
These settings apply to the GAM-Agent configurations used in the comparisons presented in Section~4.3
\begin{itemize}
    \item \textbf{Base VLMs:} Qwen2.5VL (7B, 32B, 72B), InternVL3 (8B, 14B, 78B).
    \item \textbf{Dataset:} MMBench (TEST\_EN).
    \item \textbf{Number of Base Expert Agents ($N$):} 3.
    \item \textbf{Number of Critical Expert Agents ($N_{crit}$):} 3 (when debate triggered).
    \item \textbf{Uncertainty Quantification:} $\Phi_{igen+}$ was prioritized.
    \item \textbf{Maximum Debate Rounds ($K_{max}$):} $3$.
    \item \textbf{Debate Trigger Thresholds:}
        \begin{itemize}
            \item System uncertainty ($\theta_U$): $0.45$.
            \item Inter-expert conflict score ($\theta_C$): $0.55$.
        \end{itemize}
\end{itemize}
These are consistent with the primary MMBench settings for these models.

\subsection{Settings for Ablation Studies Base Configuration (MMBench\_TEST\_V11)}
The default configuration for GAM-Agent (InternVL3-14B) in the ablation studies (Section~4.4 and Figure~\ref{fig:ablation_results_condensed} of the main paper), before individual components or parameters were ablated, used the following:
\begin{itemize}
    \item \textbf{Base VLM:} InternVL3-14B.
    \item \textbf{Dataset:} MMBench\_TEST\_V11.
    \item \textbf{Number of Base Expert Agents ($N$):} 3.
    \item \textbf{Number of Critical Expert Agents ($N_{crit}$):} 3 (when debate triggered).
    \item \textbf{Uncertainty Quantification:} $\Phi_{igen+}$ with $\alpha_{\Phi}=0.5, \beta_{\Phi}=0.5$.
    \item \textbf{Uncertainty Weighting Sensitivity ($\beta_{weight}$):} $1.5$.
    \item \textbf{Maximum Debate Rounds ($K_{max}$):} $3$.
    \item \textbf{Debate Trigger Thresholds:} $\theta_U = 0.45$, $\theta_C = 0.55$.
    \item \textbf{Debate Termination Criteria:} $\theta_{U, \text{term}} = 0.15$, $\epsilon = 0.01$.
\end{itemize}
This configuration achieved 88.80\% accuracy and served as the baseline for both the component ablations in Figure~\ref{fig:ablation_results_condensed} (main paper) and the hyperparameter ablations in Section~\ref{sec:supplementary_hyperparameter_ablation} of this supplement.

\subsection{Settings for Study on Maximum Debate Rounds (MMBench\_V11\_Test)}
For the experiment analyzing the impact of `max\_debate\_round' (Section~4.5, Figure~\ref{fig:max_rounds_ablation_qwen7b_focused} of the main paper):
\begin{itemize}
    \item \textbf{Base VLM:} Qwen2.5VL-7B.
    \item \textbf{Dataset:} MMBench\_V11\_Test.
    \item \textbf{Number of Base Expert Agents ($N$):} 3.
    \item \textbf{Number of Critical Expert Agents ($N_{crit}$):} 3 (when debate triggered).
    \item \textbf{Uncertainty Quantification:} $\Phi_{igen+}$ with $\alpha_{\Phi}=0.5, \beta_{\Phi}=0.5$.
    \item \textbf{Uncertainty Weighting Sensitivity ($\beta_{weight}$):} $1.5$.
    \item \textbf{Debate Trigger Thresholds:} $\theta_U = 0.45$, $\theta_C = 0.55$.
    \item \textbf{Debate Termination Criteria (other than $K_{max}$):} $\theta_{U, \text{term}} = 0.15$, $\epsilon = 0.01$.
    \item \textbf{Varied Parameter:} $K_{max}$ was varied from 0 to 9.
\end{itemize}

\subsection{Settings for Expert Weight Trajectory Analysis (Logical Reasoning Tasks)}
For the analysis of expert weight dynamics (Section~4.6, Figure~\ref{fig:expert_weight_dynamics} of the main paper):
\begin{itemize}
    \item \textbf{Base VLMs:} Qwen2.5VL (3B–72B), InternVL3 (2B–78B).
    \item \textbf{Dataset:} A subset of MMBench focusing on logical reasoning tasks.
    \item \textbf{Number of Base Expert Agents ($N$):} 3 (Relational Reasoning, Scene Description, OCR experts).
    \item \textbf{Number of Critical Expert Agents ($N_{crit}$):} 3 (when debate triggered).
    \item \textbf{Uncertainty Quantification:} $\Phi_{igen+}$ with $\alpha_{\Phi}=0.5, \beta_{\Phi}=0.5$.
    \item \textbf{Uncertainty Weighting Sensitivity ($\beta_{weight}$):} $1.5$.
    \item \textbf{Maximum Debate Rounds ($K_{max}$):} $3$ (analysis shown over three rounds).
    \item \textbf{Debate Trigger Thresholds:} Assumed to be active if debate occurred, e.g., $\theta_U = 0.45$, $\theta_C = 0.55$.
    \item \textbf{Debate Termination Criteria:} $\theta_{U, \text{term}} = 0.15$, $\epsilon = 0.01$.
\end{itemize}
The focus of this experiment was on observing weight and system uncertainty evolution under these typical settings.

\section{Prompt Setting Statement}
\label{sec:supplementary_prompt_settings}

This section outlines the prompt configurations for various agents within the GAM-Agent framework. The prompts are crucial for guiding the behavior of the Large Language Models (LLMs) acting as expert agents and the aggregators.

\subsection{General Prompts for Expert Roles and the Aggregator}
\label{subsec:general_prompts}

In this subsection, we provide some examples of general persona prompts that can be used to initialize different experts and the aggregator. These illustrate how roles can be defined within the system. The actual task execution would typically involve combining such persona prompts with more specific task instructions, like those detailed in Section~\ref{subsec:vlm_expert_prompts}.

\begin{tcolorbox}[colback=lightgray!5!white, colframe=lightgray!75!black, title=Illustrative Analysis Expert Persona]
You are an expert in problem analysis and logical reasoning, skilled in applying analytical frameworks and systematic thinking approaches.
Your expertise includes breaking down complex problems, identifying key factors, and recommending structured, actionable solutions.
You are familiar with various problem-solving methods such as root cause analysis, decision matrices, and scenario evaluation, and adapt your approach based on the unique context of each task.
Consider how your skills in critical thinking, structured reasoning, and analytical problem-solving might provide valuable insights or strategies for addressing the task at hand.
\end{tcolorbox}

\begin{tcolorbox}[colback=teal!5!white, colframe=teal!75!black, title=Illustrative Strategy Expert Persona]
You are a business strategy expert with a deep understanding of markets, business models, competitive landscapes, and strategic planning.
Your expertise includes applying business frameworks, analytical tools, and market insights to identify opportunities and craft strategies.
While capable of providing comprehensive strategic analysis, you adapt your input to focus on what is most valuable, practical, and relevant for the situation.
Consider how your expertise in business innovation, competitive advantage, and strategic problem-solving might provide insightful and actionable recommendations for any task.
\end{tcolorbox}

\begin{tcolorbox}[colback=olive!5!white, colframe=olive!75!black, title=Aggregator Prompt]
You are the Wise Integrator in a multi-agent system tasked with delivering accurate, coherent, and actionable responses to user queries.
Your role is to:
\begin{itemize}
    \item Understand the user's intent and main question(s) by carefully reviewing their query.
    \item Evaluate expert inputs, preserving their quality opinions while ensuring relevance, accuracy, and alignment with the user's needs.
    \item Resolve any contradictions or gaps logically, combining expert insights into a single, unified response.
    \item Synthesize the most appropriate information into a clear, actionable, and user-friendly answer.
    \item Add your own insight if needed to enhance the final output.
\end{itemize}
Your response must prioritize clarity, accuracy, and usefulness, ensuring it directly addresses the user's needs while retaining the value of expert contributions.
Avoid referencing the integration process or individual experts.
\end{tcolorbox}

\subsection{Prompts for VLM Experts in Benchmark Evaluations}
\label{subsec:vlm_expert_prompts}

For the experiments conducted on MMBench, MVBench, and MMMU datasets, a specific set of Vision-Language Model (VLM) based experts were utilized. These include three Base Experts for initial analysis and argumentation and three Critic Experts for the debate and verification stages. The prompts for these experts are detailed below. Note that placeholders like \texttt{\{instruction\}}, \texttt{\{response\}}, etc., are dynamically filled during runtime. The `keywords' listed in the configuration (not shown here) were used to potentially aid in expert selection or routing for more complex multi-faceted queries, though for single-task evaluations, experts often processed all relevant inputs.

\subsubsection{Base Expert Prompts}

The following Base Experts are primarily responsible for generating initial analyses and participating in argumentation by providing evidence-backed claims. Their `critique\_template' and `revision\_template' (shown below each main prompt) are used during the iterative debate process.

\begin{tcolorbox}[colback=blue!5!white, colframe=blue!75!black, title=Object Recognition Expert]
\textbf{Role Definition:} You are an expert in object recognition.
\textbf{Prompt Template (\texttt{prompt\_template}):}
As an object recognition expert, identify and list all significant objects visible in the image(s). Provide details about their appearance, count if possible, and relative location. For key claims, mention the visual evidence. Original question: \texttt{\{instruction\}}

\textbf{Critique Guiding Template (\texttt{critique\_template}):}
Evaluate the object recognition part of the analysis:

\texttt{\{response\}}

Are the objects correctly identified? Are there any missed objects or incorrect descriptions? Base your critique strictly on the visual evidence.

\textbf{Revision Guiding Template (\texttt{revision\_template}):}
Based on the critique '\texttt{\{critique\}}', please revise your object recognition analysis: \texttt{\{original\_response\}}. Focus on accuracy and completeness according to the visual evidence. Original question: \texttt{\{instruction\}}
\end{tcolorbox}

\begin{tcolorbox}[colback=green!5!white, colframe=green!75!black, title=Scene Description Expert]
\textbf{Role Definition:} You are an expert in the scene description.
\textbf{Prompt Template (\texttt{prompt\_template}):}
As a scene description expert, describe the overall scene shown in the image(s), including the location, environment, lighting, atmosphere, and spatial relationships between elements. For key claims, mention the visual evidence. Original question: \texttt{\{instruction\}}

\textbf{Critique Guiding Template (\texttt{critique\_template}):}
Evaluate the scene description part of the analysis:

\texttt{\{response\}}

Is the description accurate and comprehensive? Does it capture the main elements and atmosphere of the scene? Base your critique strictly on the visual evidence.

\textbf{Revision Guiding Template (\texttt{revision\_template}):}
Based on the critique '\texttt{\{critique\}}', please revise your scene description: \texttt{\{original\_response\}}. Focus on capturing the visual details accurately. Original question: \texttt{\{instruction\}}
\end{tcolorbox}

\begin{tcolorbox}[colback=purple!5!white, colframe=purple!75!black, title=Text/OCR Analysis Expert]
\textbf{Role Definition:} You are an expert in OCR and text analysis from images.
\textbf{Prompt Template (\texttt{prompt\_template}):}
As an OCR expert, identify and transcribe any text visible in the image(s). Pay attention to signs, labels, documents, or any written content. Note the location of the text if possible. For key claims, mention the visual evidence. Original question: \texttt{\{instruction\}}

\textbf{Critique Guiding Template (\texttt{critique\_template}):}
Evaluate the OCR/text transcription part of the analysis:

\texttt{\{response\}}

Is the transcribed text accurate? Is any text missed? Is the location noted correctly? Base your critique strictly on the visual evidence.

\textbf{Revision Guiding Template (\texttt{revision\_template}):}
Based on the critique '\texttt{\{critique\}}', please revise your OCR analysis: \texttt{\{original\_response\}}. Ensure accuracy and completeness of the transcribed text based on the visual evidence. Original question: \texttt{\{instruction\}}
\end{tcolorbox}

\subsubsection{Critic Expert Prompts}
The following Critic Experts are engaged during the debate phase to verify claims, check for completeness, and ensure logical consistency. Their primary input is a `critique\_template' that guides their evaluation.

\begin{tcolorbox}[colback=red!5!white, colframe=red!75!black, title=Fact Checker (Image) - Critic]
\textbf{Role Definition:} You are an expert in fact-checking claims against visual evidence.
\textbf{Critique Guiding Template (\texttt{critique\_template}):}
Act as a fact-checking expert focusing on visual evidence.
Original Question: \texttt{\{instruction\}}
Current Answer: \texttt{\{response\}}
Critique Request: \texttt{\{critique\_request\}}
Please evaluate the factual accuracy of the claims based *only* on the image content provided.
Point out inaccuracies and suggest corrections. State your confidence (0-100\%).
\end{tcolorbox}

\begin{tcolorbox}[colback=orange!5!white, colframe=orange!75!black, title=Completeness Checker - Critic]
\textbf{Role Definition:} You are an expert in assessing the completeness of image-based analysis.
\textbf{Critique Guiding Template (\texttt{critique\_template}):}
Act as a completeness analysis expert for image descriptions.
Original Question: \texttt{\{instruction\}}
Current Answer: \texttt{\{response\}}
Critique Request: \texttt{\{critique\_request\}}
Evaluate if the analysis fully addresses the question based on the *entire* image content.
Are there missing details or aspects from the image(s) that are relevant and should have been included? Provide suggestions. State your confidence (0-100\%).
\end{tcolorbox}

\begin{tcolorbox}[colback=gray!10!white, colframe=gray!75!black, title=Logic Checker - Critic]
\textbf{Role Definition:} You are an expert in evaluating the logical consistency of an analysis.
\textbf{Critique Guiding Template (\texttt{critique\_template}):}
Act as a logical consistency expert for image analysis.
Original Question: \texttt{\{instruction\}}
Current Answer: \texttt{\{response\}}
Critique Request: \texttt{\{critique\_request\}}
Evaluate the logical consistency of the analysis. Are there any contradictions or unsupported conclusions based on the visual evidence and the question?
Provide suggestions for improving logical flow. State your confidence (0-100\%).
\end{tcolorbox}

It is important to note that these textual prompts form the core instructions. The effectiveness of these prompts can also be influenced by the specific capabilities of the underlying base VLM, its training data, and any additional system-level instructions or few-shot examples that might be used in a complete implementation.

\section{Case Analysis}
\label{sec:case}
In this section, we present a series of case studies, including four successful and two unsuccessful examples, to illustrate how multiple experts collaboratively analyze images in response to corresponding questions. The expert configuration consists of three analysis experts—an Object Recognition Expert, a Scene Description Expert, and a Text/OCR Analysis Expert—as well as three critique experts: a Fact Checker (Image), a Completeness Checker, and a Logic Checker. For each analysis expert, we report both their analytical response and the associated uncertainty score. For each critique expert, we provide their evaluative feedback. To enhance clarity and conciseness given the length of the responses, key excerpts are highlighted using colored underlines.

\subsection{Successful Case}
\cref{case-good1,case-good2,case-good3,case-good4} demonstrate that our method, by assigning clearly defined roles to each expert—including object recognition, scene understanding, and text/OCR analysis—facilitates comprehensive analysis across multiple modalities and semantic dimensions of the input. This structured task decomposition enhances both the depth and breadth of information processing by enabling each analysis expert to focus on a specific sub-task and generate high-quality outputs along with corresponding uncertainty estimates. A key strength of this approach lies in its explicit quantification of uncertainty, allowing the system to weigh and prioritize more reliable expert responses. The outputs from analysis experts are subsequently evaluated by critique experts—namely, a Fact Checker (Image), a Completeness Checker, and a Logic Checker—whose feedback further refines the final answer. This multi-expert architecture promotes factual accuracy, contextual completeness, and logical consistency, thereby improving the system's robustness, interpretability, and overall stability across diverse question types and input formats.

\subsection{Unsuccessful Case}
While our multi-expert framework generally yields effective results, \cref{case-bad1} exposes several critical limitations that warrant further investigation. In this case, all three analysis experts—Object Recognition, Scene Description, and Text/OCR Analysis—produced responses with low uncertainty scores, yet the final aggregated output was incorrect. This outcome suggests that high internal confidence among individual experts does not necessarily correlate with overall prediction accuracy.

A central issue stems from the subjective nature of the question, “Which image is more colorful?” The term “colorful” admits multiple interpretations, including color diversity, saturation intensity, and perceptual vividness. Both the Object Recognition and Scene Description experts focused predominantly on the saturated blue tones of the underwater image, interpreting it as more visually striking. However, they failed to consider color diversity—a more appropriate metric in this context—for determining "colorfulness." This divergence in interpretation highlights a key limitation of the current framework: the absence of a shared semantic grounding among experts when addressing subjective or ambiguous queries.

Moreover, the inclusion of the Text/OCR expert, despite the absence of textual content in the image, underscores the need for a more adaptive expert selection strategy. The existing mechanism lacks the ability to dynamically suppress irrelevant expert responses, potentially diminishing the influence of pertinent analysis during the final aggregation phase.

In addition, the critic experts—Fact Checker, Completeness Checker, and Logic Checker—did not effectively intervene despite the misaligned interpretations provided by the analysis experts. This exposes another weakness in the framework: the critic module's limited capacity to identify and correct semantic inconsistencies when expert outputs are confidently wrong yet semantically misgrounded. Without sufficient oversight from the critics, the model fails to detect the misapplication of key concepts like "colorfulness," leading to unverified and ultimately incorrect conclusions.

Finally, the model is vulnerable to systemic bias amplification when multiple experts converge on incorrect reasoning patterns. In the absence of higher-level semantic validation or external verification mechanisms (e.g., feature-level consistency checks), such coordinated errors can propagate unchecked, undermining the reliability and robustness of the system.

\cref{case-bad2} similarly reveals additional limitations of the current framework, centering on the experts’ overreliance on superficial visual cues. In this case, both the Object Recognition and Text/OCR experts emphasized the “grayish-brown” skin tone of the two rhinoceroses and concluded that the animals were of the same color. However, this assessment overlooked subtle factors that can influence the perceived coloration of animals, such as lighting, shadows, and texture variations. These experts failed to account for environmental influences—such as the soft sunlight noted by the Scene Description expert—which may introduce perceptual discrepancies due to illumination shifts.

Although the Scene Description expert analyzed contextual elements such as background vegetation and natural lighting, their focus on the “tranquil atmosphere” of the scene did not directly address the core question—whether the rhinoceroses are indeed the same color. This mismatch between the expert’s focus and the semantic intent of the question illustrates a critical limitation of the framework: when an expert's domain emphasis is misaligned with the specific query, the resulting conclusions may be incomplete or misleading.

Furthermore, despite the absence of textual content in the image, the Text/OCR expert was nonetheless activated and contributed to the color judgment. This highlights the lack of a fine-grained expert selection mechanism. Although this expert commented on the consistency of skin tones, they failed to recognize the complex interplay of visual cues—such as lighting and surface texture—that are essential for accurate color perception.

Notably, critique modules such as the Completeness Checker and Logic Checker failed to identify and rectify these semantic misalignments. For instance, the Completeness Checker did not emphasize the role of illumination and texture in color evaluation, while the Logic Checker did not flag the deviation of expert reasoning from the core question as a logical flaw.

This failure case underscores the necessity of improving the expert selection mechanism to ensure alignment between the activated experts and the task-relevant semantics. Moreover, incorporating advanced visual analysis capabilities—such as precise color comparison algorithms or illumination-invariant models—could mitigate such errors. The aggregation process should also become more sensitive to contextual, lighting, and texture variations in visual data to enhance the overall accuracy and robustness of the system.

\setlength{\tabcolsep}{5pt} 
\renewcommand{\arraystretch}{1.5} 
\begin{table}[t]
\centering
\caption{Case: Successful Expert Collaboration Example 1}
\label{case-good1}
\begin{adjustbox}{max width=\linewidth}
\begin{tabular}{@{}p{17cm}@{}}
\toprule
\textbf{Image}  \\ 
\midrule
\makebox[\linewidth]{\includegraphics[width=0.8\textwidth]{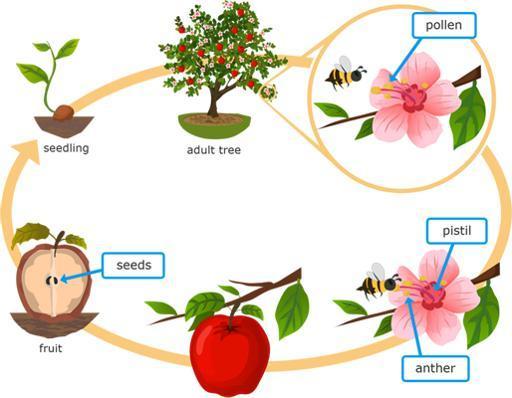}}
\end{tabular}
\end{adjustbox}
\begin{adjustbox}{max width=\linewidth}
\begin{tabular}{@{}p{17cm}@{}}
\toprule
\textbf{Question}  \\ 
\midrule
Which part of an apple tree might grow into a new tree?
\end{tabular}
\end{adjustbox}
\begin{adjustbox}{max width=\linewidth}
\begin{tabular}{@{}p{5cm}p{3.5cm}p{8cm}@{}}
\toprule
\textbf{Expert} & \textbf{Uncertainty} & \textbf{Output} \\ 
\midrule
Object Recognition & 0.1838 & 
1.\colorul[blue]{Identified the seedling stage} and the \colorul[blue]{seeds inside the fruit}. 

2.\colorul[blue]{Highlighted} that seeds are labeled inside the fruit, indicating their role in starting a new tree's \colorul[blue]{life cycle}. \\
\addlinespace[3pt]
Scene Description & 0.1245 & 
1.\colorul[orange]{Emphasized} the \colorul[orange]{educational and instructional nature} of the image.

2.Noted the spatial relationships and the \colorul[orange]{cyclical arrows} connecting each stage, reinforcing that \colorul[orange]{seeds are crucial for new growth}. \\
\addlinespace[3pt]
Text/OCR Analysis & 0.0832 & 
1.\colorul[green]{Confirmed} the labels "\colorul[green]{seeds}" inside the fruit and "\colorul[green]{seedling}" as the initial growth stage.

2.\colorul[green]{Supported} the conclusion that seeds are the part that grows into a new tree. \\
\addlinespace[3pt]
\midrule
Fact Checker (Image) & No &
1.Seeds as Tree Origin: Correct. The image labels \colorul[yellow]{seeds inside the fruit} and shows the cycle beginning with seedlings. 

2.Life Cycle: Correct. \colorul[yellow]{Arrows and stages illustrate the cyclical growth process starting from seeds}. \\
\addlinespace[3pt]
Completeness Checker & No &
1.Pollination Explanation: Briefly describe how \colorul[purple]{pollination by bees leads to fertilization and seed development}. 

2.Flower Importance: Explain the role of \colorul[purple]{flowers in the reproductive process}. \\
\addlinespace[3pt]
Logic Checker & No &
1.Object Recognition Expert: Accurately identified the \colorul[cyan]{seedling stage and the seeds inside the fruit}. 

2.Scene Description Expert: Highlighted the \colorul[cyan]{educational nature and the cyclical arrows, which effectively support the conclusion}. 

3.Text/OCR Analysis Expert: Confirmed the \colorul[cyan]{labels and supported the conclusion with textual evidence}. \\
\end{tabular}
\end{adjustbox}
\begin{adjustbox}{max width=\linewidth}
\begin{tabular}{@{}p{17cm}@{}}
\toprule
\textbf{Answer}  \\ 
\midrule
The part of an apple tree that might grow into a new tree is \colorul[red]{a seed}.\\
\bottomrule
\end{tabular}
\end{adjustbox}
\end{table}

\setlength{\tabcolsep}{5pt} 
\renewcommand{\arraystretch}{1.5} 
\begin{table}[t]
\centering
\caption{Case: Successful Expert Collaboration Example 2}
\label{case-good2}
\begin{adjustbox}{max width=\linewidth}
\begin{tabular}{@{}p{17cm}@{}}
\toprule
\textbf{Image}  \\ 
\midrule
\makebox[\linewidth]{\includegraphics[width=0.8\textwidth]{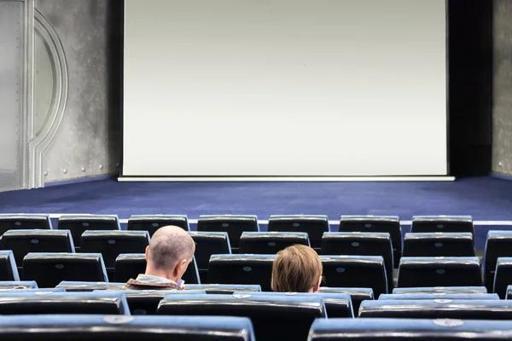}}
\end{tabular}
\end{adjustbox}
\begin{adjustbox}{max width=\linewidth}
\begin{tabular}{@{}p{17cm}@{}}
\toprule
\textbf{Question}  \\ 
\midrule
Is this place crowded?
\end{tabular}
\end{adjustbox}
\begin{adjustbox}{max width=\linewidth}
\begin{tabular}{@{}p{5cm}p{3.5cm}p{8cm}@{}}
\toprule
\textbf{Expert} & \textbf{Uncertainty} & \textbf{Output} \\ 
\midrule
Object Recognition & 0.1665 & 
1. The image shows an empty \colorul[blue]{movie theater with multiple rows of black seats arranged in a typical theater layout, facing a large, blank projection screen.}

2. The majority of the \colorul[blue]{seats are unoccupied, with only two individuals seated in the middle of the theater.}

3. These individuals are spaced out with ample room around them, and there are no other visible \colorul[blue]{patrons}. \\
\addlinespace[3pt]
Scene Description & 0.2126 & 
1. The dim lighting is consistent with a \colorul[orange]{theater setting, but it does not directly indicate the absence of a crowd.}

2. The overall scene appears quiet and still, supporting the impression of \colorul[orange]{emptiness}. \\
\addlinespace[3pt]
Text/OCR Analysis & 0.0832 & 
1.The image \colorul[green]{doesn't contain any text or symbols}, so it’s recommended to \colorul[green]{skip the Text/OCR analysis} \\
\addlinespace[3pt]
\midrule
Fact Checker (Image) & No &
1.Empty Seats: Correct. The image shows \colorul[yellow]{mostly empty seats, with only two people present}. 

2.Seating Arrangement: Correct. The seating layout resembles that of a \colorul[yellow]{typical movie theater}. 

3.Projection Screen: Correct. The \colorul[yellow]{blank screen reinforces the sense of emptiness}. \\
\addlinespace[3pt]
Completeness Checker & No &
1.Seat Count Detail: Briefly mention the \colorul[purple]{number of seats or rows} to clarify the theater’s capacity and how full it is. 

2.Seating Context: Describe whether the \colorul[purple]{two individuals are seated together or apart} to enhance the scene’s detail. 

3.Lighting Description: Indicate whether the \colorul[purple]{lighting is typical for a theater or unusually dim} to provide more context for the ambiance. \\
\addlinespace[3pt]
Logic Checker & No &
1.Object Recognition: Accurately identified the \colorul[cyan]{theater layout, empty seats, and presence of only two spaced individuals}, clearly showing low occupancy. 

2.Scene Description: Appropriately noted \colorul[cyan]{typical dim lighting and a quiet, still atmosphere} that support the impression of emptiness. 

3.Text/OCR Analysis: Correctly recognized there is \colorul[cyan]{no text to analyze}, avoiding unnecessary processing. \\
\end{tabular}
\end{adjustbox}
\begin{adjustbox}{max width=\linewidth}
\begin{tabular}{@{}p{17cm}@{}}
\toprule
\textbf{Answer}  \\ 
\midrule
This \uline{cinema} is \colorul[red]{not crowded at all.}
\end{tabular}
\end{adjustbox}
\end{table}

\setlength{\tabcolsep}{5pt} 
\renewcommand{\arraystretch}{1.5} 
\begin{table}[t]
\centering
\caption{Case: Successful Expert Collaboration Example 3}
\label{case-good3}
\begin{adjustbox}{max width=\linewidth}
\begin{tabular}{@{}p{17cm}@{}}
\toprule
\textbf{Image}  \\ 
\midrule
\makebox[\linewidth]{\includegraphics[width=0.9\textwidth]{right3.jpg}}
\end{tabular}
\end{adjustbox}
\begin{adjustbox}{max width=\linewidth}
\begin{tabular}{@{}p{17cm}@{}}
\toprule
\textbf{Question}  \\ 
\midrule
Which image is brighter?
\end{tabular}
\end{adjustbox}
\begin{adjustbox}{max width=\linewidth}
\begin{tabular}{@{}p{5cm}p{3.5cm}p{8cm}@{}}
\toprule
\textbf{Expert} & \textbf{Uncertainty} & \textbf{Output} \\ 
\midrule
Object Recognition & 0.1342 & 
1. The left image (beach scene) is significantly \colorul[blue]{brighter due to its clear, sunny environment} with a \colorul[blue]{bright blue sky and light blue ocean}.

2. The right image (underwater scene) is \colorul[blue]{darker}, typical of underwater environments where \colorul[blue]{light is absorbed and scattered}, resulting in a greenish hue. \\
\addlinespace[3pt]
Scene Description & 0.0832 & 
1. The left image features a \colorul[orange]{bright, natural lighting scenario} with minimal cloud cover, enhancing its brightness.

2. The right image has \colorul[orange]{dimmer lighting} and a \colorul[orange]{greenish tint}, characteristic of underwater scenes with less light penetration. \\
\addlinespace[3pt]
Text/OCR Analysis & 0.0832 & 
1.The image \colorul[green]{doesn't contain any text or symbols}, so it’s recommended to \colorul[green]{skip the Text/OCR analysis} \\
\addlinespace[3pt]
\midrule
Fact Checker (Image) & No &
1.Brightness Levels: Correct. The \colorul[yellow]{left image is noticeably brighter with a clear sky and light-colored ocean}. 

2.Lighting Environment: Correct. The \colorul[yellow]{right image is darker, reflecting the typical low-light conditions underwater}. \\
\addlinespace[3pt]
Completeness Checker & No &
1.Brightness Metrics: Briefly mention \colorul[purple]{specific metrics or tools (e.g., average pixel brightness) used to quantify brightness}, clarifying the basis of the comparison. 

2.Color Context: Describe the \colorul[purple]{dominant colors in each image and how they influence perceived brightness, such as blues and whites brightening the left image versus greens and darker tones dimming the right one}. \\
\addlinespace[3pt]
Logic Checker & No &
1.Object Recognition: Correctly identified the \colorul[cyan]{bright beach scene and darker underwater scene with greenish tones}. 

2.Scene Description: Accurately noted \colorul[cyan]{bright natural light in the beach image and dim, greenish light underwater}. 

3.Text/OCR Analysis: Properly skipped \colorul[cyan]{text analysis due to no text present}. \\
\end{tabular}
\end{adjustbox}
\begin{adjustbox}{max width=\linewidth}
\begin{tabular}{@{}p{17cm}@{}}
\toprule
\textbf{Answer}  \\ 
\midrule
Compared to the image on the right, the image on the left is \colorul[red]{clearly brighter}.
\end{tabular}
\end{adjustbox}
\end{table}

\setlength{\tabcolsep}{5pt} 
\renewcommand{\arraystretch}{1.5} 
\begin{table}[t]
\centering
\caption{Case: Successful Expert Collaboration Example 4}
\label{case-good4}
\begin{adjustbox}{max width=\linewidth}
\begin{tabular}{@{}p{17cm}@{}}
\toprule
\textbf{Image}  \\ 
\midrule
\makebox[\linewidth]{\includegraphics[width=0.6\textwidth]{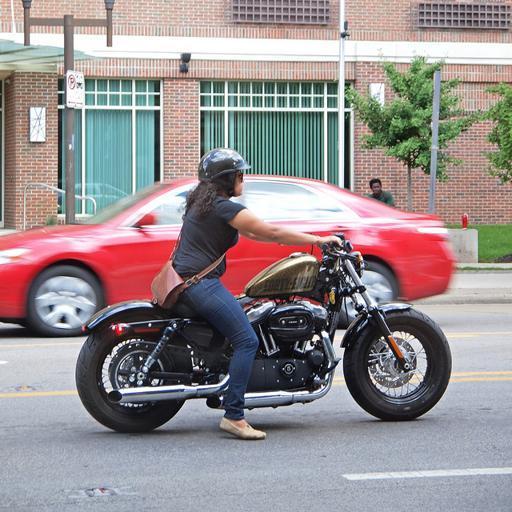}}
\end{tabular}
\end{adjustbox}
\begin{adjustbox}{max width=\linewidth}
\begin{tabular}{@{}p{17cm}@{}}
\toprule
\textbf{Question}  \\ 
\midrule
Which one is the correct caption of this image?
\end{tabular}
\end{adjustbox}
\begin{adjustbox}{max width=\linewidth}
\begin{tabular}{@{}p{5cm}p{3.5cm}p{8cm}@{}}
\toprule
\textbf{Expert} & \textbf{Uncertainty} & \textbf{Output} \\ 
\midrule
Object Recognition & 0.1670 & 
1. The image features the following key objects: 
\colorul[blue]{Motorcycle}: Black with chrome details, located in the center, ridden by a person. 
\colorul[blue]{Person}: Wearing a black helmet, dark shirt, jeans, and carrying a brown bag, riding the motorcycle. 
\colorul[blue]{Red Car}: A red sedan, partially in motion, to the left of the motorcycle. 
\colorul[blue]{Building}: A brick building with green shutters, in the background. 
\colorul[blue]{Street Elements}: Multiple street signs, a streetlight, and a sidewalk along the road. \\
\addlinespace[3pt]
Scene Description & 0.5213 & 
1. The image depicts a \colorul[orange]{city street scene} with a person riding a \colorul[orange]{black motorcycle}, wearing a \colorul[orange]{helmet}, \colorul[orange]{dark shirt}, and \colorul[orange]{jeans}. A \colorul[orange]{red car} is visible in the background, along with a \colorul[orange]{brick building} featuring \colorul[orange]{green shutters}. 

2. The setting includes \colorul[orange]{trees} and a \colorul[orange]{sidewalk}, suggesting an \colorul[orange]{urban environment}. The scene takes place during the \colorul[orange]{day}, with a \colorul[orange]{casual and busy atmosphere}. \\
\addlinespace[3pt]
Text/OCR Analysis & 0.1651 & 
1.There is a \colorul[green]{sign visible on the left side of the image}, but the \colorul[green]{text on it is not clear enough to transcribe accurately}. \\
\addlinespace[3pt]
\midrule
Fact Checker (Image) & No &
1.Motorcycle Color: Correct. The motorcycle is \colorul[yellow]{clearly black with chrome details and a large front fairing}.

2.Background Details: Correct. The background includes a \colorul[yellow]{red car and a brick building with green window shutters}, matching the description. \\
\addlinespace[3pt]
Completeness Checker & No &
1.Minor Background Elements: Briefly mention the \colorul[purple]{person near the building} to add context to the scene.

2.Accessory Details: Describe the \colorul[purple]{rider’s bag and shoes} to enrich the overall depiction. \\
\addlinespace[3pt]
Logic Checker & No &
1.Object Recognition: Clearly identified the \colorul[cyan]{motorcycle, rider, car, building, and street elements}.

2.Scene Description: Concisely described the \colorul[cyan]{urban setting, people, vehicles, and daytime atmosphere}.

3.Text/OCR Analysis: Correctly noted a \colorul[cyan]{sign is present but text is unreadable, so no transcription attempted}. \\
\end{tabular}
\end{adjustbox}
\begin{adjustbox}{max width=\linewidth}
\begin{tabular}{@{}p{17cm}@{}}
\toprule
\textbf{Answer}  \\ 
\midrule
The correct caption for the image is: "A \colorul[red]{woman} is \colorul[red]{riding a motorcycle} down the \colorul[red]{street}."
\end{tabular}
\end{adjustbox}
\end{table}

\setlength{\tabcolsep}{5pt} 
\renewcommand{\arraystretch}{1.5} 
\begin{table}[t]
\centering
\caption{Case: Unsuccessful Expert Collaboration Example 1}
\label{case-bad1}
\begin{adjustbox}{max width=\linewidth}
\begin{tabular}{@{}p{17cm}@{}}
\toprule
\textbf{Image}  \\ 
\midrule
\makebox[\linewidth]{\includegraphics[width=0.8\textwidth]{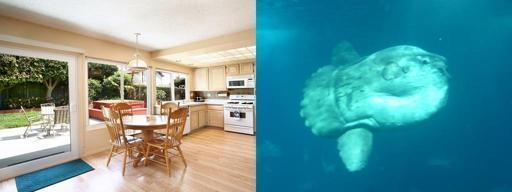}}
\end{tabular}
\end{adjustbox}
\begin{adjustbox}{max width=\linewidth}
\begin{tabular}{@{}p{17cm}@{}}
\toprule
\textbf{Question}  \\ 
\midrule
Which image is more colorful?
\end{tabular}
\end{adjustbox}
\begin{adjustbox}{max width=\linewidth}
\begin{tabular}{@{}p{5cm}p{3.5cm}p{8cm}@{}}
\toprule
\textbf{Expert} & \textbf{Uncertainty} & \textbf{Output} \\ 
\midrule
Object Recognition & 0.0991 & 
1. Left Image (Kitchen/Dining Area):  
The left image shows a \colorul[blue]{kitchen/dining area with neutral tones}, featuring a dining table, cabinets, stove, and a blue rug. \colorul[blue]{Natural light brightens the space}.

2. Right Image (Underwater Scene):  
The right image displays a \colorul[blue]{stingray in vibrant blue water}, creating a \colorul[blue]{colorful and lively underwater scene}.

3. The right image features a \colorul[blue]{deep blue underwater scene}, which is \colorul[blue]{significantly more vibrant} compared to the \colorul[blue]{neutral tones of the kitchen and dining area on the left}. The \colorul[blue]{blue water and the contrast with the stingray} enhance the \colorul[blue]{overall colorfulness}. \\
\addlinespace[3pt]
Scene Description & 0.2553 & 
1.The right side has a \colorul[orange]{rich, deep blue background}, creating a \colorul[orange]{vivid and striking visual effect}. The left side, while bright, has a \colorul[orange]{muted and natural color palette dominated by light wood and neutral tones}. The \colorul[orange]{underwater scene’s intensity and saturation} make it \colorul[orange]{more colorful}. \\
\addlinespace[3pt]
Text/OCR Analysis & 0.0832 &
1.Since the image does not contain any \colorul[green]{text, labels, or symbols}, it is advisable to \colorul[green]{exclude Text/OCR analysis}. \\
\addlinespace[3pt]
\midrule
Fact Checker (Image) & No &

1.Underwater Scene: Correct. The right image clearly shows a \colorul[yellow]{deep blue underwater environment}.

2.Color Contrast: Correct. The \colorul[yellow]{deep blue tones} in the right image are more \colorul[yellow]{vivid and intense} compared to the \colorul[yellow]{muted, neutral tones} of the kitchen on the left. \\
\addlinespace[3pt]
Completeness Checker & No &

1.Color Details: Briefly mention the \colorul[brown]{dominant colors and their saturation} in both images to add clarity.

2.Lighting Effects: Describe how \colorul[brown]{lighting and natural elements influence color perception} in each scene. \\
\addlinespace[3pt]
Logic Checker & No &

1.Object Recognition: Accurately identified \colorul[cyan]{key elements in both images}—the kitchen/dining area on the left and the vibrant underwater scene with a stingray on the right.

2.Scene Description: Effectively contrasted the \colorul[cyan]{muted, neutral tones} of the kitchen with the \colorul[cyan]{vivid, saturated colors} of the underwater environment.

3.Text/OCR Analysis: Appropriately noted the \colorul[cyan]{absence of text}, correctly skipping text analysis. \\
\end{tabular}
\end{adjustbox}
\begin{adjustbox}{max width=\linewidth}
\begin{tabular}{@{}p{17cm}@{}}
\toprule
\textbf{Answer}  \\ 
\midrule
Compared to the image on the left, the image on the right is \colorul[red]{obviously more colorful}.
\end{tabular}
\end{adjustbox}
\end{table}

\setlength{\tabcolsep}{5pt} 
\renewcommand{\arraystretch}{1.5} 
\begin{table}[t]
\centering
\caption{Case: Unsuccessful Expert Collaboration Example 2}
\label{case-bad2}
\begin{adjustbox}{max width=\linewidth}
\begin{tabular}{@{}p{17cm}@{}}
\toprule
\textbf{Image}  \\ 
\midrule
\makebox[\linewidth]{\includegraphics[width=0.8\textwidth]{wrong2.jpg}}
\end{tabular}
\end{adjustbox}
\begin{adjustbox}{max width=\linewidth}
\begin{tabular}{@{}p{17cm}@{}}
\toprule
\textbf{Question}  \\ 
\midrule
Are the two animals in the picture the same color?
\end{tabular}
\end{adjustbox}
\begin{adjustbox}{max width=\linewidth}
\begin{tabular}{@{}p{5cm}p{3.5cm}p{8cm}@{}}
\toprule
\textbf{Expert} & \textbf{Uncertainty} & \textbf{Output} \\ 
\midrule
Object Recognition & 0.1213 & 
1. \colorul[blue]{Two rhinoceroses standing side by side}, facing slightly left.

2. Both have a \colorul[blue]{grayish-brown skin tone}, prominent horns, and large ears.

3. A \colorul[blue]{yellow bird with red markings} is perched near the horn of each rhinoceros. \\
\addlinespace[3pt]
Scene Description & 0.9481 & 
1.Environment:
The \colorul[orange]{two rhinoceroses} in the image are in a \colorul[orange]{natural environment}, possibly a forest or savanna, with green vegetation in the background.
The lighting is natural, suggesting it might be early morning or late afternoon, with soft sunlight. The atmosphere is calm and serene.

2.Position and Pose:
The \colorul[orange]{two rhinoceroses} are standing \colorul[orange]{side by side}, facing forward, with similar poses, indicating they may belong to the same group. \\
\addlinespace[3pt]
Text/OCR Analysis & 0.2315 & 
1.There is no visible text or written content in the image. Both animals appear to be the \colorul[green]{same color}, which is a typical \colorul[green]{grayish-brown hue} for rhinos. The visual evidence supporting this conclusion includes the \colorul[green]{consistent coloration} and \colorul[green]{texture visible on both animals' skin}. \\
\addlinespace[3pt]
\midrule
Fact Checker (Image) & No & 
1.Presence of Birds: The small birds on the rhinos’ heads are \colorul[yellow]{clearly separate} and do not affect the color evaluation of the rhinos.

2.Color Consistency: Both rhinos show a \colorul[yellow]{similar grayish-brown color}, as confirmed by the visual evidence.

3.Impact of Birds: The birds \colorul[yellow]{do not influence the perceived color} of the rhinos. \\
\addlinespace[3pt]
Completeness Checker & No & 
1.Texture and Lighting: Mention how \colorul[brown]{texture and lighting conditions affect color perception} and visual impact.

2.Background Elements: Briefly describe \colorul[brown]{background features like green foliage} to add context. \\
\addlinespace[3pt]
Logic Checker & No & 
1.Object Recognition: Correctly identified the two rhinos, their color, and the \colorul[cyan]{birds on their horns}.

2.Scene Description: Described the natural setting, \colorul[cyan]{lighting, and rhinos’ positions}.

3.Text/OCR Analysis: Noted absence of text and \colorul[cyan]{consistent rhino coloration}. \\
\end{tabular}
\end{adjustbox}
\begin{adjustbox}{max width=\linewidth}
\begin{tabular}{@{}p{17cm}@{}}
\toprule
\textbf{Answer}  \\ 
\midrule
The \colorul[red]{two rhinoceroses} in the picture are \colorul[red]{indeed the same color.}
\end{tabular}
\end{adjustbox}
\end{table}

\newpage

\clearpage

\end{document}